\documentclass[a4paper, 11pt]{article}

\usepackage{hyperref}
\usepackage{authblk}
\usepackage{hyperref}

\usepackage{amsmath,amssymb, amsfonts}
\usepackage{bm}
\usepackage{algorithm,algorithmic,bbm}

\usepackage{graphicx}
\usepackage{amsmath}
\usepackage{amsfonts}
\usepackage{color}
\usepackage{bm}
\usepackage{mathrsfs}

\usepackage{amssymb}
\usepackage{algorithm}
\usepackage{algorithmic}

\DeclareFontFamily{OT1}{pzc}{}
\DeclareFontShape{OT1}{pzc}{m}{it}{<-> s * [1.200] pzcmi7t}{}
\DeclareMathAlphabet{\mathpzc}{OT1}{pzc}{m}{it}

\newtheorem{assumption}{Assumption}

\newtheorem{lemma}{Lemma}
\newtheorem{theorem}{Theorem}

\newcommand{\BEAN}{\begin{eqnarray*}}
\newcommand{\EEAN}{\end{eqnarray*}}
\newcommand{\BEA}{\begin{eqnarray}}
\newcommand{\EEA}{\end{eqnarray}}
\newcommand{\BEQ}{\begin{equation}}
\newcommand{\EEQ}{\end{equation}}
\newcommand{\BIT}{\begin{itemize}}
\newcommand{\EIT}{\end{itemize}}
\newcommand{\eg}{{e.g.}}
\newcommand{\ie}{{i.e.}}
\newcommand{\nn}{{\nonumber}}
\newcommand{\ones}{\mathbf{1}}
\newcommand{\reals}{\mathbb{R}}
\newcommand{\expect}{\mathbb{E}}
\newcommand{\tp}{\mathsf{T}}

\newcommand{\bmmu}{\bm{\mu}}
\newcommand{\bmx}{\mathbf{x}}
\newcommand{\bmy}{\mathbf{y}}
\newcommand{\bmz}{\mathbf{z}}
\newcommand{\bmh}{\mathbf{h}}
\newcommand{\bmu}{\mathbf{u}}

\newcommand{\bmH}{\mathbf{H}}
\newcommand{\bmk}{\mathbf{k}}
\newcommand{\reg}{\mathrm{Regret}}
\newcommand{\cvs}{\mathrm{CV}}

\newcommand{\DOLR}{\mathrm{DOLR}}

\begin{document}

\title{\bf Distributed Online Linear Regression}

\author{Deming Yuan\thanks{D. Yuan is with the Research School of Engineering, The Australian National University, ACT 0200, Canberra, Australia; School of Automation, Nanjing University of Science and Technology, Nanjing 210094, China. (Email: dmyuan1012@gmail.com)}, Alexandre Proutiere\thanks{A. Proutiere is with Department of Automatic Control, KTH  Royal Institute of Technology, Stockholm 100-44, Sweden. (Email: alepro@kth.se)}, and Guodong Shi\thanks{G. Shi is with the Australian Center for Field Robotics, School of Aerospace, Mechanical and Mechatronic Engineering, The University of Sydney, NSW 2006, Sydney, Australia. (Email: guodong.shi@sydney.edu.au)}
}

\date{}

\maketitle

\begin{abstract}
We study online linear regression problems in a distributed setting, where the data is spread over a network. In each round, each network node proposes a linear predictor, with the objective of fitting the \emph{network-wide} data. It then updates its predictor for the next round according to the received local feedback and information received from neighboring nodes. The predictions made at a given node are assessed through the notion of regret, defined as the difference between their cumulative network-wide square errors and those of the best off-line network-wide linear predictor. Various scenarios are investigated, depending on the nature of the local feedback (full information or bandit feedback), on the set of available predictors (the decision set), and the way data is generated (by an oblivious or adaptive adversary). We propose simple and natural distributed regression algorithms, involving, at each node and in each round, a local gradient descent step and a communication and averaging step where nodes aim at aligning their predictors to those of their neighbors. We establish regret upper bounds typically in ${\cal O}(T^{3/4})$ when the decision set is unbounded and in ${\cal O}(\sqrt{T})$ in case of bounded decision set.
\end{abstract}


\section{Introduction}

Linear regression aims at identifying a predictor $\bmh\in \mathbb{R}^m\mapsto \bmh^\tp \bmy\in \mathbb{R}$ fitting some data $$\{ (\bmh(1),z(1)),\ldots ,(\bmh(T),z(T))\}\subseteq (\mathbb{R}^m\times\mathbb{R})^\tp$$ as accurately as possible, e.g., with small square loss $\sum_{t=1}^T {1\over 2}(\bmh(t)^\tp \bmy - z(t))^2$. The predictor is parametrized by $\bmy$ constrained to belong to the decision set ${\cal K}$, a convex subset of $\mathbb{R}^m$. In the already well-studied online version of this problem, data samples are observed sequentially, and each of them provides an opportunity to update the prediction of $\bmy$.

This paper investigates online regression problems in a distributed setting where the data is spread over a network. The network is modeled as a directed graph $\mathrm{G} = \left( \mathrm{V},\mathrm{E} \right)$ with $\mathrm{V}=\{1,\dots,n\}$. Each node $i\in \mathrm{V}$ is associated with the sequence of covariate vectors and corresponding outcomes $$\{ (\bmh_{i}(1),z_{i}(1)),\ldots ,(\bmh_{i}(T),z_i(T))\}\subseteq (\mathbb{R}^m\times\mathbb{R})^\tp.
$$ In each round, node $i$ first proposes a linear predictor parametrized by $\bmx_i(t)$ within the decision set ${\cal K}$, a convex subset of $\mathbb{R}^m$, and experiences a loss $\sum_{j=1}^n{1\over 2}(\bmh_j(t)^\tp \bmx_i(t)-z_j(t))^2$. Node $i$ can then update its predictor depending on the local feedback and on the information received from neighboring nodes, i.e., from nodes in $\mathrm{N}_i=\big\{j:(j,i)\in\mathrm{E}\big\}$. We consider two types of local feedback:
\begin{itemize}
\item[(i)] Full information feedback, where node $i$ has access to both $\bmh_i(t)$ and $z_i(t)$;
\item[(ii)] Bandit feedback, where node $i$ has only access to the local loss $\theta_{i,t}(\bmx_i(t))$ where for any $\bmy\in {\cal K}$,  $\theta_{i,t}(\bmy)\triangleq{1\over 2}(\bmh_i(t)^\tp \bmy-z_i(t))^2$.
\end{itemize}
At the end of round $t$, the information received at node $i$ from node $j\in \mathrm{N}_i$ depends on the predictor $\bmx_j(t)$ and the local feedback received at node $j$ in this round. In each round, each node is allowed to transmit a vector in $\mathbb{R}^m$ to its neighbors. The objective is to design distributed regression algorithms so that each node holds a predictor accurately fitting the {\it network-wide} data. The performance of such an algorithm is assessed, at node $i$, through its regret, defined by:
\begin{equation}
\begin{aligned}[b]
\reg_{\mathrm{LS}}(i,T)  \triangleq \sum_{t=1}^{T}  \sum_{j=1}^{n} \theta_{j,t}(\bmx_{i}(t)) - \sum_{t=1}^{T}  \sum_{j=1}^{n} \theta_{j,t}(\bmy^\star_{\mathrm{LS}}),
\label{regret-LS}
\end{aligned}
\end{equation}
where $\bmy^\star_{\mathrm{LS}}$ denotes the parameter of the offline {\it network-wide} optimal linear predictor:
$$
\bmy^\star_{\mathrm{LS}} \triangleq \arg\min_{\bmy\in{\cal K}} \sum_{t=1}^{T}  \sum_{j=1}^{n}{1\over 2} (\bmh_j(t)^\tp \bmy - z_j(t))^2.
$$
The data is arbitrary as if it was generated by an adversary. Most of our results concern {\it non-adaptive} adversaries where the data is generated before the first round, and unless otherwise specified, we consider this scenario. However, {\it adaptive} adversaries are also investigated, and in this case, the data $(\bmh_i(t),z_i(t))$ at node $i$ in round $t$ may depend on the predictions made so far at node $i$.

Distributed versions of online linear regression problems are motivated by at least two observations. First, many learning tasks involve very large datasets, and distributing the data and its treatment in a network of communicating computing units may be necessary (see e.g. \cite{submodular2016, scaman2018} and references therein). Then, when the data contains sensitive personal information (bio-medical  and social network data, among others), the whole data  might come naturally as physically separated subsets from different parties in a network, and merging those data subsets could lead to potential privacy risks, see e.g. \cite{nedic2018ieee}.

\subsection{Contributions}

In this paper, we investigate distributed online regression problems in various scenarios, depending on the decision set, and the type of local feedback. For each scenario, we devise distributed algorithms with sub-linear regret. Our algorithms are simple and naturally involve in each round a local gradient descent step and a communication and averaging step where nodes aim at aligning their predictors to those of their neighbors. The regret analysis of our algorithms is however challenging as it requires us to understand how these two steps interact. The key ingredients to establish sub-linear regret are an upper bound on the accumulative magnitude of the gradients used in the sequence of updates, and an upper bound on the cumulated disagreement of the predictors held at the various nodes of the network. Controlling the accumulative gradient is particularly technical especially in the case where the decision set is not restricted (${\cal K}=\mathbb{R}^m$). Note that existing convergence analyses of first-order convex optimization algorithms generally rely on assuming a bounded gradient. Without any restriction on the decision set, this assumption does not hold.\\
Here is a summary of our contributions:
\begin{itemize}
\item In the case of full information feedback and full decision set ${\cal K}=\mathbb{R}^m$, we show that our algorithm achieves a $\mathcal{O}(T^{3/4})$ regret. The scaling in the size of the network of our regret upper bound is specified for a few network examples, and provides preliminary insights into the communication complexity vs. regret trade-off. We also establish that a regret scaling as $\mathcal{O}(T^{3/4})$ is achieved under bandit feedback.
\item When the decision set ${\cal K}$ is bounded, a regret upper bound $\mathcal{O}(\sqrt{T})$ is established even in the case of bandit feedback and adaptive adversaries. The algorithm however involves in each round a projection onto ${\cal K}$. Such projections can be computationally expensive for complex ${\cal K}$. To circumvent this difficulty, we adopt the so-called {\it optimization with long-term constraints} framework \cite{mahdavi2012} where the decision constraints are relaxed and where one allows the use of projections onto a simpler set (typically a ball) containing the decision set ${\cal K}$. In this framework, we propose a distributed algorithm with regret scaling at most as $\mathcal{O}(\sqrt{T})$ and with cumulative constraints' violation no greater than $\mathcal{O}(T^{3/4})$, under both full information and bandit feedback.
\item Finally, we investigate the case where the linear regression admits exact solutions, i.e., there exists $\bmy^\star\in \mathbb{R}^m$ such that $\sum_{t=1}^{T}  \sum_{i=1}^{n} \theta_{i,t}(\bmy^\star)=0$. Under this assumption, we devise a distributed algorithm with $\mathcal{O}(\sqrt{T})$ regret.
\end{itemize}

%
%
%

\subsection{Related Work}

Various {\it centralized} online linear regression problems with full information feedback have been studied since the 1990's. \cite{foster91} considered online regression with binary labels. Later, \cite{bianchi} investigated the centralized version of our problem, and provided a ${\cal O}(\sqrt{T})$ regret upper bound for particular gradient descent algorithms. Regret upper bounds with a similar scaling were also shown for exponentiated gradient descent algorithms \cite{kivinen97}. \cite{vovk98} developed the so-called Aggregating Algorithm (AA) and established a ${\cal O}(\log(T))$ regret upper bound. Online linear regression remains an interesting and active area of research, see \cite{Bartlett15,Bartlett18} for recent developments. As far as we are aware, the present paper is the first providing a regret analysis for {\it distributed} online linear regression problems. The local gradient descent step involved in our algorithms is essentially similar to the gradient descent performed in \cite{bianchi} (at least in the full information feedback setting), and is combined with a simple local averaging step. This simplicity makes the regret tractable, and is sufficient to achieve sub-linear regret. Replacing the local gradient step by Vovk's algorithm in our algorithms could well lead to better regret, but the analysis of the resulting distributed algorithm seems out of reach for now.

Online linear regression with square loss is a particular instance of online convex optimization, widely studied in a centralized setting with full information and bandit feedback \cite{McMahan2004, sss2011,hazan2016now,flaxman2005soda,agarwal2010colt,hazan2014nips,bubeck2017stoc,shamir2017jmlr,saha2011aistat}. Under bandit feedback, centralized algorithms with ${\cal O}(\textrm{poly}(m)\sqrt{T})$ regret have been proposed, see \cite{bubeck2017stoc} for recent developments. There have also been a few attempts to investigate distributed online convex optimization problems, mainly in the engineering literature \cite{Raginsky2011,fan2013tkde,hosseini2016tac,zhang2017projection}. The results therein rely on assumptions we cannot afford in our settings: strong convexity, bounded loss function and its gradient, bounded decision set.

It is also worth mentioning work on off-line distributed convex optimization (the objective functions do not evolve over time in an arbitrary manner). The idea of distributed optimization for separable  functions can actually be traced back to \cite{tsitsiklis2986}, where a network structure was introduced to characterize the communication opportunities among the processors for the computation process. In recent years, this line of research was extended significantly in various directions  \cite{nedic2010tac,duchi2012tac,nedic2018ieee, scaman2018}, but results there do not apply to online optimization problems.

Finally, the case where the linear regression admits exact solutions has been treated in \cite{faber} for applications of learning classes of smooth functions.

\subsection{Notation and Terminology}
$\| \bmx \|$ denotes the Euclidean norm of a vector $\bmx\in\reals^m$. Let $\ones_{n} \in \reals^{n} $ and $\mathbf{0}_{s} \in \reals^{s} $ be the vectors with all entries equal to one and zero, respectively. Let $[\bmx]_i$ be the $i$th entry of a vector $\bmx$, and $[W]_{ij}$ the $(i,j)$th element of a matrix $W$. Let $\mathpzc{P}_{\mathcal{A}}(\bmx)$ be the Euclidean projection of a vector $\bmx$ onto a convex set $\mathcal{A}$, \ie, $\mathpzc{P}_{\mathcal{A}}(\bmx) = \arg\min_{\bmy\in \mathcal{A}} \| \bmx - \bmy \| $. The Euclidean ball centered in $\mathbf{0}_{m}$ of radius $R$ is $\mathbb{B}^{m}_{R} = \{ \bmx\in\reals^m  \,:\, \| \bmx \| \leq R  \}$.

$\bmH(t)$ denotes the $(n\times m)$ matrix whose $i$-th row is $\bmh_i(t)^\tp$. Associated with the graph $\mathrm{G}$, we introduce a weight matrix $W_{\mathrm{G}} \in\reals^{n\times n}$ that captures the information flow among the nodes. In the reminder of the paper without further mention, we impose the following assumption: (i) $\mathrm{G}$ is strongly connected; (ii)
$[W_{\mathrm{G}}]_{ij} \geq 0 $, $\forall i,j\in\mathrm{V}$, and $[W_{\mathrm{G}}]_{ij} > 0 $ if and only if $j\in\mathrm{N}_i$; (iii)
$W_{\mathrm{G}}$ is doubly stochastic, \ie, $\sum_{j=1}^{n} [W_{\mathrm{G}}]_{ij} = 1$ and $\sum_{i=1}^{n} [W_{\mathrm{G}}]_{ij} = 1$ for every $i,j\in\mathrm{V}$. These conditions imply that the second largest singular value of $W_{\mathrm{G}}$ satisfies
$\sigma_2\left(W_{\mathrm{G}} \right) <1$ (see, \eg, \cite{levin2008markov}).


\section{Full Information Feedback}\label{section-full}

In this section, we focus on the case of full information feedback, where at the end of each round $t$, each node $i$ has access to the local covariate vector $\bmh_i(t)$ and the corresponding outcome $z_i(t)$ to update the estimate $\bmx_i(t)$. We make no assumption on the magnitude of $\bmx_i(t)$, but assume the following.

\begin{assumption}\label{assumption-full-info} (i)
$\mathrm{rank(\bmH(t))} = m$ for at least one $t\in\{ 1,\ldots,T\}$; (ii)
$\left\| \bmh_{i}(t) \right\|^2 \leq \alpha_{\bmh}$ for all $i\in\mathrm{V}$ and $t=1,\ldots,T$ with $\alpha_{\bmh} > 0$;
(iii)
$\left|  \bmh_{i}(t)^\tp \bmy_{\mathrm{LS}}^{\star}  - z_{i}(t) \right| \leq \theta^{\star}$ holds for all $i\in\mathrm{V}$ and $t=1,\ldots,T$ with $\theta^{\star} > 0$.
\end{assumption}

\subsection{Algorithm and Regret Guarantees}
The proposed algorithm DOLR, whose pseudo-code is presented in Algorithm \ref{alg-odls}, is a distributed version of the classical Online Gradient Descent algorithm (see, \eg, \cite{sss2011, bianchi}) applied to the loss function $\theta_{i,t}(\bmy)  =  \frac{1}{2} (\bmh_{i}(t)^{\tp} \bmy - z_{i}(t) )^2$ with step size $\frac{1}{\alpha_{\bmh} T^\beta}$. However, in contrast to the literature on online convex optimization \cite{zinkevich2003icml,agarwal2010colt,hazan2014nips,bubeck2017stoc,sss2011}, we do not impose Lipschitz continuity or boundedness of the gradient of loss function $\theta_{i,t}(\bmy)$ (because we impose no constraints on the magnitude of $\bmy$). Examining the literature on (online) distributed multi-agent optimization, a key technical assumption needed to establish the convergence of (online) distributed gradient optimization is the boundedness of the gradient of the objective function. However, in our context this is equivalent to assuming the boundedness of $\left|  \bmh_{i}(t)^\tp \bmx_{i}(t)  - z_{i}(t) \right|$ for all $i\in\mathrm{V}$ and $t=1,\ldots,T$, which, in general, does not hold. A main challenge in the regret analysis of DOLR is hence to control the gradients accumulated over time.

\begin{algorithm}
\caption{$\DOLR$ with Full Information Feedback - DOLR-FIF}
\label{alg-odls}
\begin{algorithmic}[1]
\ENSURE Initial local estimates $\bmx_{i}(1) \in \reals^{m}$ for all $i\in\mathrm{V}$
\FOR{$t=1$ to $T$}
\STATE
Node $i$ locally computes $\mathpzc{l}_{i}(t) =  \bmx_{i}(t) - \frac{1}{\alpha_{\bmh} T^\beta} \cdot \bmh_{i}(t) \left(  \bmh_{i}(t)^\tp \bmx_{i}(t) - z_{i}(t)  \right)$

\STATE
Node $i$ receives $\mathpzc{l}_{j}(t)$ from $j\in\mathrm{N}_{i}$, and updates its estimated parameter vector as:
\begin{equation*}
\begin{aligned}
\bmx_{i}(t+1) =  [W_{\mathrm{G}}]_{ii} \cdot \mathpzc{l}_{i}(t) + \sum_{j\in\mathrm{N}_{i}} [W_{\mathrm{G}}]_{ij} \cdot \mathpzc{l}_{j}(t)
\end{aligned}
\end{equation*}
\ENDFOR
\end{algorithmic}
\end{algorithm}

\noindent
Denote $\bmx(1) = [\bmx_1(1)^\tp,\ldots,\bmx_n(1)^\tp]^\tp$. The following theorem provides an upper bound of the regret of DOLR-FIF.
\begin{theorem}\label{theorem-ls-regret}
Under Assumption \ref{assumption-full-info}, the regret of DOLR-FIF with $\beta = \frac{3}{4}$ satisfies for all $i\in\mathrm{V}$ and $T\ge 2$:
\begin{equation*}
\begin{aligned}
\reg_{\mathrm{LS}}(i,T)  \leq  C_{\mathrm{G}}(\bmx(1),\bmy^\star_{\mathrm{LS}})  T^{3/4},
\end{aligned}
\end{equation*}
where
$
C_{\mathrm{G}}(\bmx(1),\bmy^\star_{\mathrm{LS}}) = \mathcal{O}\left( \left( \frac{\sigma_2(W_{\mathrm{G}})}{1-\sigma_{2}(W_{\mathrm{G}})}  \right)^2 \left( n^4  + n^3 \| \bmx(1) - \ones_{n} \otimes  \bmy^\star_{\mathrm{LS}} \|^2 \right)    \right).
$
\end{theorem}

\subsection{Communication Complexity vs. Regret Trade-off}
The regret upper bound derived in Theorem \ref{theorem-ls-regret} depends on the network through $C_{\mathrm{G}}(\bmx(1),\bmy^\star_{\mathrm{LS}})$, whose leading term scales as  $\left( \frac{\sigma_2(W_{\mathrm{G}})}{1-\sigma_2(W_{\mathrm{G}})}  \right)^2n^4$. On the other hand, the number of vectors transmitted per round in the network (i.e., the communication complexity of DOLR) is simply equal to the number of edges in $\mathrm{G}$. By investigating the sensitivity of $C_{\mathrm{G}}(\bmx(1),\bmy^\star_{\mathrm{LS}})$ w.r.t. $\mathrm{G}$, we can get preliminary insights into the communication complexity vs. regret trade-off.

To this aim, we assume here that the weight matrix $W_{\mathrm{G}}$ is chosen as the \emph{maximum-degree weights}:
\begin{eqnarray}
[W_{\mathrm{G}}]_{ij}
&=&
\left\{
\begin{array}{ll}
\frac{1}{1 + d_{\max}},           &\qquad (j,i)\in\mathrm{E}  \\
1 - \frac{d_i}{1 + d_{\max}} ,    &\qquad i=j \\
0,                                &\qquad (j,i)\notin\mathrm{E}
\end{array}
\right.
\label{weight-max}
\end{eqnarray}
where $d_{\max} = \max_{i\in\mathrm{V}} \{d_i\}$ is the maximum degree of $\mathrm{G}$ ($d_i$ denotes the degree of node $i$). In \cite{duchi2012tac}, the authors established useful estimates of $\sigma_2 (W_{\mathrm{G}})$ for particular graphs $\mathrm{G}$, which in turn translate to explicit regret upper bounds for these graphs.

\BIT
\item
\emph{Complete graphs:} Every node is connected to every other node in the network. In this case we have $W_{\mathrm{G}}  = \frac{\ones_n \ones_n^{\tp}}{n}$ and $\sigma_2(W_{\mathrm{G}}) = 0$, which implies that the leading term in $C_{\mathrm{G}}(\bmx(1),\bmy^\star_{\mathrm{LS}})$ vanishes. Analyzing $C_{\mathrm{G}}(\bmx(1),\bmy^\star_{\mathrm{LS}})$, we may actually improve the regret upper bound of Theorem \ref{theorem-ls-regret} so that it scales as $\left( \| \bmx(1) - \ones_{n} \otimes  \bmy^\star_{\mathrm{LS}} \|^2  + n \right)T^{3/4}$. However, this requires maximum communication among the nodes in the network, i.e., $n(n-1)$ vectors are transmitted per round.
\item
\emph{Random geometric graph:} Nodes are generated independently and uniformly in a unit square and any two nodes are connected if the distance between them satisfies $r = \Omega \left( \sqrt{ \frac{ \log^{1+\epsilon} (n)}{n} } \right)$ for some $\epsilon > 0$. Then, with high probability $\sigma_2(W_{\mathrm{G}}) = 1 - \Omega\left(  \frac{ \log (n)}{n} \right)$ (see \cite{boyd2006}), which implies that the regret upper bound scales as  $\frac{n^6}{\log^2 (n)}T^{3/4}$. The maximum degree is here bounded by $\log^{1+\epsilon} (n) + \sqrt{2} c \log (n)$ w. p. at least $1-2/n^c$ (for any $c>0$). Hence at most approximately  $2\log^{1+\epsilon}(n)n$ vectors are transmitted every round.
\item
\emph{$k$-regular expander graph:} In this case $\sigma_2(W_{\mathrm{G}})$ is constant, and our regret upper bound scales as $n^4 T^{3/4}$. The communication complexity is $2kn$ vectors transmitted per round.
\item
\emph{Path graph:} In this case we have $\sigma_2(W_{\mathrm{G}}) = 1- \Theta \left( \frac{1}{n^2} \right)$ and the regret upper bound scales as $n^8 T^{3/4}$ for a number of vectors transmitted per round equal to $2n$.
\EIT

\section{Bandit Feedback}\label{section-bandit}
Next, we focus on the case of bandit feedback, where at the end of each round $t$, node $i$ has access to the bandit information of the local regression cost $\theta_{i,t}(\bmx_i(t))$, or more precisely, node $i$ can get the values of the function $\theta_{i,t}$ at two points around $\bmx_{i}(t)$. Indeed, the proposed algorithm, DOLR-BF,  whose pseudo-code is provided in Algorithm \ref{alg-odls-bandit}, is a distributed version of the Expected Gradient Descent algorithm \cite{agarwal2010colt, duchi2015, nesterov2017, shamir2017jmlr} for online convex optimization with multi-point bandit feedback. Algorithms with single-point bandit feedback such as those in \cite{flaxman2005soda} could be also studied similarly, but would exhibit worse regret guarantees (e.g. scaling as $T^{5/6}$).

\begin{algorithm}
\caption{$\DOLR$ with Bandit Feedback -- DOLR-BF}
\label{alg-odls-bandit}
\begin{algorithmic}[1]
\ENSURE Initial local estimates $\bmx_{i}(1) \in \reals^{m}$ for all $i\in\mathrm{V}$
\FOR{$t=1$ to $T$}
\STATE
Node $i$ queries $\theta_{i,t} ( \bmx_{i}(t)  +  \epsilon \bmu_{i}(t) )$ and $\theta_{i,t} ( \bmx_{i}(t)  -  \epsilon \bmu_{i}(t) )$ with $\bmu_{i}(t) \in \reals^m$ be a unit vector generated uniformly at random (\ie, $\| \bmu_{i}(t) \| = 1$), and computes
\begin{equation*}
\begin{aligned}
\mathpzc{g}_{i,t} ( \bmx_{i}(t) ) &=  \frac{m}{2 \epsilon} \big( \theta_{i,t} ( \bmx_{i}(t)  +  \epsilon \bmu_{i}(t) )  -   \theta_{i,t} ( \bmx_{i}(t)  -  \epsilon \bmu_{i}(t) ) \big) \cdot \bmu_{i}(t)
\end{aligned}
\end{equation*}
\STATE
Node $i$ locally computes $\mathpzc{l}_{i}(t) =  \bmx_{i}(t) - \frac{1}{\kappa T^\beta} \cdot \mathpzc{g}_{i,t} ( \bmx_{i}(t) )$
\STATE
Node $i$ receives $\mathpzc{l}_{j}(t)$ from $j\in\mathrm{N}_{i}$, and updates its estimated parameter vector as:
\begin{equation*}
\begin{aligned}
\bmx_{i}(t+1) =  [W_{\mathrm{G}}]_{ii} \cdot \mathpzc{l}_{i}(t) + \sum_{j\in\mathrm{N}_{i}} [W_{\mathrm{G}}]_{ij} \cdot \mathpzc{l}_{j}(t)
\end{aligned}
\end{equation*}
\ENDFOR
\end{algorithmic}
\end{algorithm}

To account for the randomness introduced in Algorithm \ref{alg-odls-bandit}, we slightly modify the definition of regret:
\begin{equation}
\begin{aligned}
\reg_{\mathrm{LS}}(i,T)   \triangleq \sum_{t=1}^{T}  \sum_{j=1}^{n} \expect\left[ \theta_{j,t}(\bmx_{i}(t)) \right]
- \sum_{t=1}^{T}  \sum_{j=1}^{n}  \theta_{j,t}(\bmy^\star_{\mathrm{LS}}) .
\label{regret-LS-bandit}
\end{aligned}
\end{equation}

\begin{theorem}\label{theorem-ls-bandit}
If Assumption \ref{assumption-full-info} holds and $\alpha_{\bmh} < n$, then the regret of DOLR-BF with parameters $\beta = \frac{3}{4}$, $\epsilon  =  \frac{1}{\sqrt{T}}$, and $\kappa >  \frac{2 n m^2 \alpha_{\bmh}}{n - \alpha_{\bmh}}$, satisfies for all $i\in\mathrm{V}$ and $T\ge 2$:
\begin{equation*}
\begin{aligned}
\reg_{\mathrm{LS}}(i,T)  &\leq&  B_{\mathrm{G}}(\bmx(1),\bmy^\star_{\mathrm{LS}},m)  T^{3/4},
\end{aligned}
\end{equation*}
where $ B_{\mathrm{G}}(\bmx(1),\bmy^\star_{\mathrm{LS}},m) = \mathcal{O}\left( \left( \frac{\sigma_2(W_{\mathrm{G}})}{1-\sigma_2(W_{\mathrm{G}})} \right)^2  \left(  n^5 m^2 + n^4 m^4 + n^3 m^2 \| \bmx(1) - \ones_{n} \otimes  \bmy^\star_{\mathrm{LS}} \|^2 \right) \right) $.
\end{theorem}

\subsection{Adaptive Adversaries and Bounded Decision Space}\label{subsection-ada-adv}

In this subsection, we consider the case where the data is generated by an {\it adaptive} adversary as in \cite{flaxman2005soda}: at each node $i$ and in any round $t$, the covariate vector and its outcome $(\bmh_i(t),z_i(t))$ may depend on the predictions made so far by node $i$, i.e., on $(\bmh_i(1),z_i(1),\bmx_i(1),\ldots, \bmh_i(t-1),z_i(t-1),\bmx_i(t-1))$. We assume that $(\bmh_i(t,z_i(t)) \in \mathcal{K}_{\mathbf{Ad}}$ where
\begin{equation}
\begin{aligned}
\mathcal{K}_{\mathbf{Ad}} =
\{ (\bmh,z ) \,:\,  \| \bmh \|^2 \leq \alpha_{\mathbf{h}},\ | z  | \leq \alpha_z  \}.
\label{space-k-ad}
\end{aligned}
\end{equation}

In addition, we consider scenarios where the decision space ${\cal K}$ is compact, and w.l.o.g. that 0 lies in the interior of ${\cal K}$ (see e.g. \cite{agarwal2010colt} for a discussion). Hence we make the following assumption (commonly adopted even in centralized online bandit optimization \cite{flaxman2005soda,agarwal2010colt,hazan2014nips,bubeck2017stoc,saha2011aistat}).
\begin{assumption}\label{assumption-adversary}
The decision space of all the nodes, denoted $\mathcal{K}$, satisfies that
\begin{equation*}
\begin{aligned}
\mathbb{B}^{m}_{r}  \subseteq \mathcal{K} \subseteq \mathbb{B}^{m}_{R},\quad\hbox{for some }   0 < r \leq R .
\end{aligned}
\end{equation*}
\end{assumption}

For this regression problem, we propose DOLR-BF-AA (AA stands for 'Adaptive Adversaries') obtained by simply replacing the update Step 4 of DOLR-BF by
\begin{equation*}
\begin{aligned}
\bmx_{i}(t+1) &=  \mathpzc{P}_{(1-\xi) \mathcal{K}} \Big(   [W_{\mathrm{G}}]_{ii} \cdot \mathpzc{l}_{i}(t) + \sum_{j\in\mathrm{N}_{i}} [W_{\mathrm{G}}]_{ij} \cdot \mathpzc{l}_{j}(t)  \Big).
\end{aligned}
\end{equation*}
When $\xi = \epsilon/r$, the projection onto $(1-\xi) \mathcal{K}$ ensures that $\bmx_i(t+1)$ remains in ${\cal K}$. Now since the adversaries may adapt to the previously randomly selected predictors, the covariate vectors and their outcomes may be random as well. To account for this additional randomness and for the restriction on the decision set, we modify the definition of regret once more:
$$
\reg_{\mathrm{LS}}(i,T) \triangleq \sum_{t=1}^{T}  \sum_{j=1}^{n} \expect \left[ \theta_{j,t}(\bmx_{i}(t)) \right] - \min_{\bmy\in\mathcal{K}}\sum_{t=1}^{T}  \sum_{j=1}^{n} \expect \left[ \theta_{j,t}(\bmy) \right].
$$
The following theorem provides an upper bound of the regret of DOLR-BF-AA. The scaling in $\sqrt{T}$ of this upper bound results from the compactness of the decision set ${\cal K}$ and that of the adversary strategy set ${\cal K}_{\mathbf{Ad}}$.

\begin{theorem}\label{theorem-ls-bandit-adversary}
Assume that $\mathrm{rank(\bmH(t))} = m$ for at least one $t\in\{ 1,\ldots,T\}$ and that Assumption \ref{assumption-adversary} holds. The regret of DOLR-BF-AA with parameters $\beta=1/2$, $\epsilon  =  \frac{1}{\sqrt{T}}$, $\kappa=1$, and $\xi=\epsilon/r$, satisfies for all $i\in\mathrm{V}$ and $T\ge \left\lceil  1/r^2 \right\rceil$:
\begin{equation*}
\begin{aligned}
\reg_{\mathrm{LS}}(i,T)  &\leq  \tilde{B}_{\mathrm{G}}(m,L)  \sqrt{T},
\end{aligned}
\end{equation*}
where $ \tilde{B}_{\mathrm{G}}(m,L) = \mathcal{O}\left( \frac{\sigma_2(W_{\mathrm{G}})}{1-\sigma_2(W_{\mathrm{G}})}   n^{3/2} m^2 L^3 \right) $ and $L = \alpha_{\bmh} R + \sqrt{\alpha_{\bmh}} \alpha_z$.
\end{theorem}

\section{Regression with Long-term Constraints}\label{section-constraints}

When the decision set ${\cal K}$ is a convex (strict) subset of $\mathbb{R}^m$, the previous algorithm (as any other typical algorithm for online optimization with constraints) performs a projection onto ${\cal K}$ in each round. This projection can be computationally expensive when ${\cal K}$ is complex. To circumvent this issue, \cite{mahdavi2012} proposes to allow the algorithm to violate the constraints by projecting onto a simpler set ${\cal B}$ containing ${\cal K}$. Under such simplification, the algorithm performance is quantified using both its regret and its cumulative constraints' violation. In this section, we present distributed versions of algorithms in \cite{mahdavi2012} and provide upper bounds on their regret and cumulative constraints' violation.

For illustrative purposes, we assume that ${\cal K}$ is a polyope, defined through a (potential large) set of linear inequalities:
\begin{equation*}
\begin{aligned}
\mathcal{K}  \triangleq  \Big\{ \bmy \in \reals^m \,:\,  \bmk_q^\tp \bmy \leq 0, q= 1,\ldots,s \Big\} \subset \mathbb{B}_R^m,
\end{aligned}
\end{equation*}
where the constraint vectors $\bmk_q\in \mathbb{R}^m$, $q=1,\ldots,s$, are known to all nodes. With such a decision set, we consider algorithms using projections on the (simpler) set ${\cal B}=\mathbb{B}_R^m$.
We make the following assumption.
\begin{assumption}\label{assumption-constraints}
The constraint vectors $\bmk_q$, $q=1,\ldots,s$, satisfy: $\| \bmk_q \| \leq K_{\mathrm{I}}$ for some $K_{\mathrm{I}}>0$.
\end{assumption}

The regret of a distributed regression algorithm with long-term constraints remains defined as:
\begin{equation}
\begin{aligned}[b]
\reg_{\mathrm{LS}}(i,T) & \triangleq  \sum_{t=1}^{T} \sum_{j=1}^{n} \theta_{j,t}(\bmx_{i}(t)) -  \sum_{t=1}^{T} \sum_{j=1}^{n} \theta_{j,t}(\bmy^\star_{\mathrm{LS}}),
\label{regret-dolr-constraints-regret}
\end{aligned}
\end{equation}
at a node $i\in\mathrm{V}$, where $\bmy_{\mathrm{LS}}^{\star} = \arg\min_{\bmy\in \mathcal{K}}\sum_{t=1}^{T}  \sum_{j=1}^{n} \theta_{j,t}(\bmy) $; and its {\em network-wide} cumulative constraints' violation is:
\begin{equation}
\begin{aligned}[b]
\cvs(s,T) &\triangleq \sum_{t=1}^{T} \sum_{i=1}^{n} \sum_{q=1}^{s} \left[ \bmk_q^{\tp}\bmx_{i}(t) \right]_+,
\label{regret-dolr-constraints-full-cvs}
\end{aligned}
\end{equation}
where $[a]_+=\max\{0,a\}$ for all $a\in \mathbb{R}$.

\subsection{Full Information Feedback}

In the proposed algorithm, node $i$ local computations are based on the following \emph{augmented Lagrangian}: for all $\bmy\in \mathbb{R}^m$ and $\bmmu\in\mathbb{R}^s$,
\begin{equation}
\begin{aligned}[b]
\mathscr{L}_{i,t}(\bmy,\bmmu) &\triangleq  \theta_{i,t}(\bmy)
+ \sum_{q=1}^{s} [\bmmu]_q \left[ \bmk_q^{\tp}\bmy \right]_+ - \frac{\pi}{2} \| \bmmu \|^2,\qquad t=1,\ldots,T,
\label{regularized-lagrangian}
\end{aligned}
\end{equation}
where $\bmmu \in\reals^s$ is the vector of Lagrange multipliers associated with the constraints at node $i$, $[\bmmu]_q$ denotes the $q$-th component of $\bmmu$, and $\pi > 0$ is the regularization parameter. The pseudo-code of our algorithm is presented in Algorithm \ref{alg-dolr-cons}. There, each node $i$ sequentially updates its estimate $\mathbf{x}_i(t)\in\mathbb{R}^m$ and a dual vector $\bmmu_i(t)\in \mathbb{R}^s$.

\begin{algorithm}
\caption{$\DOLR$ with Full Information Feedback and Constraints -- DOLR-FIFC}
\label{alg-dolr-cons}
\begin{algorithmic}[1]
\ENSURE Initial local estimates $\bmx_{i}(1) \in \mathcal{K}$ and dual vectors $\bmmu_i(1) = \mathbf{0}_s$ for all $i\in\mathrm{V}$\FOR{$t=1$ to $T$}
\STATE
Node $i$ locally computes
\begin{equation*}
\begin{aligned}
\mathpzc{l}_{i}(t) =  \bmx_{i}(t) - \eta \Big( \bmh_{i}(t) \left(  \bmh_{i}(t)^\tp \bmx_{i}(t) - z_{i}(t)  \right) + \sum_{q=1}^{s} [\bmmu_i(t)]_q \partial\left[ \bmk_q^{\tp}\bmx_i(t) \right]_+  \Big)
\end{aligned}
\end{equation*}
\STATE
Node $i$ receives $\mathpzc{l}_{j}(t)$ from $j\in\mathrm{N}_{i}$, and updates its estimate as
\begin{equation*}
\begin{aligned}
\bmx_{i}(t+1) =  \mathpzc{P}_{\mathbb{B}^m_R} \Big( [W_{\mathrm{G}}]_{ii} \cdot \mathpzc{l}_{i}(t) + \sum_{j\in\mathrm{N}_{i}} [W_{\mathrm{G}}]_{ij} \cdot \mathpzc{l}_{j}(t)  \Big)
\end{aligned}
\end{equation*}
\STATE
Node $i$ updates its dual vector as $\left[\bmmu_i(t+1)\right]_q =  \frac{\left[ \bmk_q^{\tp}\bmx_i(t+1) \right]_+}{\pi}$, for $q= 1,\ldots,s$
\ENDFOR
\end{algorithmic}
\end{algorithm}

Note that $\partial \left[ \bmk_q^{\tp}\bmx_i(t) \right]_+$ can be calculated explicitly  for $q=1,\ldots,s$ as
\begin{eqnarray*}
\partial\left[ \bmk_q^{\tp}\bmx_i(t) \right]_+
&=&
\left\{
\begin{array}{ll}
\bmk_q,                   &\qquad \mathrm{if}\ \bmk_q^{\tp}\bmx_i(t) > 0 \\
0 ,                            &\qquad  \mathrm{otherwise} .
\end{array}
\right.
\end{eqnarray*}
Moreover, the second step of the algorithm is a gradient descent of the local  augmented Lagrangian $\mathscr{L}_{i,t}(\bmy,\bmmu)$ since $
\mathpzc{l}_{i}(t) =  \bmx_{i}(t) - \eta   \nabla_{\bmy} \mathscr{L}_{i,t}(\bmx_{i}(t),\bmmu_i(t))$. Denote $L = \alpha_{\bmh} R + \sqrt{\alpha_{\bmh}} \alpha_z$.

\begin{theorem}\label{theorem-dolr-constraints}
Assume that $\mathrm{rank(\bmH(t))} = m$ for at least one $t\in\{ 1,\ldots,T\}$, that for all $t$, $(\bmh_{i}(t),z_i(t) ) \in \mathcal{K}_{\mathbf{Ad}}$ defined in (\ref{space-k-ad}), and that Assumption \ref{assumption-constraints} holds. The regret and the cumulative constraints' violation of DOLR-FIFC with parameters $\eta = \frac{1}{c s K_{\mathrm{I}}^2 T^\beta} $ and $\pi = \frac{1}{T^\beta}$ for some $\beta\in(0,1)$ and $c>1$ satisfy for all $i\in\mathrm{V}$ and $T\ge 2$:
\begin{equation*}
\begin{aligned}
\reg_{\mathrm{LSC}}(i,T)  &\leq E_{\mathrm{G}}(L,K_{\mathrm{I}},R) T^{\max\{\beta,1-\beta\}}  \\
\cvs(s,T) &\leq E_{\mathrm{G}}^{\dag}(L,K_{\mathrm{I}},R)  T^{1- \beta/2}
\end{aligned}
\end{equation*}
where $E_{\mathrm{G}}(L,K_{\mathrm{I}},R) = \mathcal{O}\left( \frac{1}{(1-\sigma_2(W_{\mathrm{G}}))^2} \left( n^{5/2} L R + n^4 L^2 \right) + n K_{\mathrm{I}}^2 R^2   \right)$ and $E_{\mathrm{G}}^{\dag}(L,K_{\mathrm{I}},R) = \mathcal{O}\left( n L K_{\mathrm{I}} R   \right)$.
\end{theorem}

The regret and cumulative constraints' violation of DOLR-FIFC upper bounds scale as $T^{\max\{\beta,1-\beta\}}$ and $T^{1- \beta/2}$, respectively. These scalings are identical to those of the centralized algorithms proposed in \cite{mahdavi2012,jenatton2016,yuan2018nips}. The parameter $\beta$ tunes the trade-off between regret and cumulative constraints' violation, and DOLR-FIFC achieves a regret scaling as $\sqrt{T}$ for the {\it balanced} case $\beta=1/2$. Another approach to alleviate the computational cost of projections relies on leveraging the conditional gradient algorithm. This has been investigated in \cite{zhang2017projection}, where the authors establish that such an approach would yield a ${\cal O}(T^{3/4})$ regret (instead of ${\cal O}(\sqrt{T})$ in our case). Further note that in the next subsection, we establish that these scalings also hold in case of bandit feedback (the three aforementioned papers deal with full information feedback only).

\subsection{Bandit Feedback}

To extend the previous algorithm to the case of bandit feedback, we replace the augmented Lagragian (\ref{regularized-lagrangian}) by the following smoothed version:
\begin{equation}
\begin{aligned}[b]
\mathscr{L}^{\mathrm{b}}_{i,t}(\bmy,\bmmu) &\triangleq  \hat{\theta}_{i,t}(\bmy)
+ \sum_{q=1}^{s} [\bmmu]_q \left[ \bmk_q^{\tp}\bmy \right]_+ - \frac{\pi}{2} \| \bmmu \|^2,\qquad  i\in\mathrm{V},t=1,\ldots,T
\label{regularized-lagrangian-smoothed}
\end{aligned}
\end{equation}
where $\hat{\theta}_{i,t}(\bmy)   \triangleq \expect_{\mathbf{v}\in\mathbb{B}_1^m} \big[ \theta_{i,t}(\bmy  +  \epsilon  \mathbf{v}) \big]$. The algorithm with bandit feedback, whose pseudo-code is presented below, uses an estimate of the gradient of $\mathscr{L}^{\mathrm{b}}_{i,t}(\bmy,\bmmu)$, as well as an additional shrinkage parameter $\xi$ when projecting onto $\mathbb{B}^m_R$ (to ensure that the query points $\bmx_{i}(t)  \pm  \epsilon \bmu_{i}(t)$ belong to the decision set $\mathcal{K}$).

\begin{algorithm}
\caption{$\DOLR$ with Bandit Feedback and Constraints -- DOLR-BFC}
\label{alg-dolr-cons-bandit}
\begin{algorithmic}[1]
\ENSURE Initial local estimates $\bmx_{i}(1) \in \mathcal{K}$ and dual vectors $\bmmu_i(1) = \mathbf{0}_s$ for all $i\in\mathrm{V}$
\FOR{$t=1$ to $T$}
\STATE
Node $i$ locally computes $\mathpzc{l}_{i}(t) =  \bmx_{i}(t) - \eta \left( \mathpzc{g}_{i,t} ( \bmx_{i}(t) ) + \sum_{q=1}^{s} [\bmmu_i(t)]_q \partial\left[ \bmk_q^{\tp}\bmx_i(t) \right]_+  \right)$, where $\mathpzc{g}_{i,t} ( \bmx_{i}(t) )$ is computed as in Step 2 of Algorithm \ref{alg-odls-bandit}
\STATE
Node $i$ receives $\mathpzc{l}_{j}(t)$ from $j\in\mathrm{N}_{i}$, and updates its estimate
\begin{equation*}
\begin{aligned}
\bmx_{i}(t+1) =  \mathpzc{P}_{(1-\xi)\mathbb{B}^m_R} \Big( [W_{\mathrm{G}}]_{ii} \cdot \mathpzc{l}_{i}(t) + \sum_{j\in\mathrm{N}_{i}} [W_{\mathrm{G}}]_{ij} \cdot \mathpzc{l}_{j}(t)  \Big)
\end{aligned}
\end{equation*}
\STATE
Node $i$ updates its dual vector as $\left[\bmmu_i(t+1)\right]_q =  \frac{\left[ \bmk_q^{\tp}\bmx_i(t+1) \right]_+}{\pi}$, for $q= 1,\ldots,s$
\ENDFOR
\end{algorithmic}
\end{algorithm}

\begin{theorem}\label{theorem-dolr-constraints-bandit}
Assume that $\mathrm{rank(\bmH(t))} = m$ for at least one $t\in\{ 1,\ldots,T\}$, that for all $t$, $(\bmh_{i}(t),z_i(t) ) \in \mathcal{K}_{\mathbf{Ad}}$ defined in (\ref{space-k-ad}), and that Assumption \ref{assumption-constraints} holds. The regret and the cumulative constraints' violation of DOLR-BFC with parameters   $\eta = \frac{1}{c s K_{\mathrm{I}}^2 T^\beta}$, $\pi = \frac{1}{T^\beta}$, $\epsilon = \frac{1}{T^\gamma}$ and $\xi = \frac{1}{R T^\gamma}$ for some $\beta\in(0,1)$, $\gamma \geq \beta$ and $c>1$ satisfy for all $i\in\mathrm{V}$ and $T\ge 2$:
\begin{equation*}
\begin{aligned}
\expect\left[ \reg_{\mathrm{LSC}}(i,T) \right]  &\leq F_{\mathrm{G}}(m,L,K_{\mathrm{I}},R)  T^{\max\{\beta,1-\beta\}}  \\
\expect\left[ \cvs(s,T) \right] &\leq F_{\mathrm{G}}^{\dag}(m,L,K_{\mathrm{I}},R)  T^{1- \beta/2}
\end{aligned}
\end{equation*}
where $F_{\mathrm{G}}  = \mathcal{O}\left( \frac{1}{(1-\sigma_2(W_{\mathrm{G}}))^2} \left( m^2 n^{5/2} L^2 + n^{5/2} L R + n^4 L^2 \right) + n K_{\mathrm{I}}^2 R^2  \right)$ and $F_{\mathrm{G}}^{\dag}= \mathcal{O}\left( m n L K_{\mathrm{I}} R \right)$.
\end{theorem}

\section{Distributed Online Exact Linear Regression}\label{section-exact}

We conclude the paper by investigating scenarios where the network-wide online linear regression has exact solutions, i.e., $\sum_{t=1}^{T}  \sum_{i=1}^{n} \theta_{i,t}(\bmy^\star)=0$. In other words, we make the following assumption.

\begin{assumption}\label{assumption-LE-solution}
The network-wide exact linear regression solution set $\mathcal{S}^{\star}(T) = \{ \bmy \,:\, \bmh_i(t)^\tp \bmy^\star = z_i(t) , i\in \mathrm{V}, t=1,\ldots, T \}$ is non-empty.
\end{assumption}

We consider the full information feedback, but do not make any assumption on the decision set, i.e., ${\cal K}=\mathbb{R}^m$. The centralized version of this problem was studied in \cite{faber}. The existence of a linear model exactly matching the data allows us to improve the regret upper bound of the regret (also referred to as $\ell_2$-regret) of DOLR-FIF, here equal to:
\begin{equation*}
\begin{aligned}
\reg_{\mathrm{LS}}(i,T)   = \sum_{t=1}^{T}  \sum_{j=1}^{n}    \frac{1}{2} (\bmh_{j}(t)^{\tp} \bmx_{i}(t) - z_{j}(t) )^2.
\end{aligned}
\end{equation*}
It further makes it possible to devise an algorithm, referred to as DOELR (Distributed Online Exact Linear Regression), with sub-linear $\ell_1$-regret defined at node $i\in\mathrm{V}$ as:
\begin{equation}
\begin{aligned}
\reg_{\ell_1}(i,T)   \triangleq  \sum_{t=1}^{T}  \sum_{j=1}^{n} \frac{1}{\| \bmh_{j}(t) \|} \left| \bmh_{j}(t)^\tp \bmx_{i}(t) - z_{j}(t)  \right|.
\label{regret}
\end{aligned}
\end{equation}

\subsection{$\ell_2$-Regret}

The algorithm DOLR-FIF can be applied here, and the following theorem provides a regret upper bound scaling as $\sqrt{T}$ (instead of $T^{3/4}$ in absence of an exact linear model). Denote $\bmx_{\mathrm{avg}}(1) = \frac{1}{n} \sum_{i=1}^{n} \bmx_i(1)$.

\begin{theorem}\label{theorem-convergence-v2}
Assume that $\left\| \bmh_{i}(t) \right\|^2 \leq \alpha_{\bmh}$ holds for all $i\in\mathrm{V}$ and $t=1,\ldots,T$ for some $\alpha_{\bmh} > 0$, and that Assumption \ref{assumption-LE-solution} holds. The regret of DOLR-FIF with $ \beta  = \frac{1}{2}$ satisfies for all $i\in\mathrm{V}$ and $T\ge 2$:
\begin{equation*}
\begin{aligned}
\reg_{\mathrm{LS}}(i,T) &\leq&  R_{\mathrm{G}}(\bmx(1))\sqrt{T},
\end{aligned}
\end{equation*}
where $ R_{\mathrm{G}}(\bmx(1)) = \mathcal{O}\left( \left(  \frac{\sigma_2(W_{\mathrm{G}})}{1-\sigma_2(W_{\mathrm{G}})}  \right)^2 n^3  \sum_{i=1}^{n} \mathrm{dist}^2(\bmx_{i}(1), \mathpzc{P}_{\mathcal{S}^{\star}(T)} \left( \bmx_{\mathrm{avg}}(1) \right) ) \right) $.
\end{theorem}

\subsection{$\ell_1$-Regret}

To get low $\ell_1$-regret, we propose the following algorithm.

\begin{algorithm}
\caption{Distributed Online Exact Linear Regression -- DOELR}
\label{alg-odle}
\begin{algorithmic}[1]
\ENSURE Initial local estimates $\bmx_{i}(1) \in \reals^{m}$ for all $i\in\mathrm{V}$
\FOR{$t=1$ to $T$}
\STATE
Node $i$ computes $\mathpzc{l}_{i}(t) =  \bmx_{i}(t) -  \bmh_{i}(t) \left(  \bmh_{i}(t)^\tp \bmx_{i}(t) - z_{i}(t)  \right)\Big/\| \bmh_{i}(t)  \|^2$
\STATE
Node $i$ receives $\mathpzc{l}_{j}(t)$ from $j\in\mathrm{N}_{i}$, and updates its estimate as
\begin{equation*}
\begin{aligned}
\bmx_{i}(t+1) &=  [W_{\mathrm{G}}]_{ii} \cdot \mathpzc{l}_{i}(t) + \sum_{j\in\mathrm{N}_{i}} [W_{\mathrm{G}}]_{ij} \cdot \mathpzc{l}_{j}(t)
\end{aligned}
\end{equation*}
\ENDFOR
\end{algorithmic}
\end{algorithm}

\begin{theorem}\label{theorem-convergence}
Under Assumption \ref{assumption-LE-solution}, set $\bmx_{i}(1) = \bmx_1 \in \reals^m$ for all $i\in\mathrm{V}$. The $\ell_1$-regret of DOELR satisfies for all $i\in\mathrm{V}$ and $T\ge 2$:
\begin{equation*}
\begin{aligned}
\reg_{\ell_1}(i,T) &\leq&  P_{\mathrm{G}}(\bmx(1)) \sqrt{T},
\end{aligned}
\end{equation*}
where $ P_{\mathrm{G}}(\bmx(1)) = \mathcal{O}\left( \frac{\sigma_2(W_{\mathrm{G}})}{1-\sigma_2(W_{\mathrm{G}})} n^{2} \sqrt{\sum_{i=1}^{n} \mathrm{dist}^2(\bmx_{i}(1), \mathpzc{P}_{\mathcal{S}^{\star}(T)} \left( \bmx_{\mathrm{avg}}(1) \right) )} \right) $ .
\end{theorem}


\section{Conclusions}

This paper introduces various distributed online linear regression problems, and shows that for these problems, sub-linear regret can be achieved by simple distributed algorithms, combining local gradient and local averaging steps in each round. The paper leaves open numerous and interesting questions. For example, one may wonder whether the regret upper bounds are order-optimal; can we derive informative regret lower bounds? How would these bounds scale with the network size? We may also think to extend our results to more generic or different distributed online convex problems. As we mentioned earlier, a critical technical ingredient in the regret analysis is an upper bound on the accumulated magnitude of the gradient of the loss function, and hence we certainly need to start with specific convex programs where this is possible.

\newpage

\appendix
\section{Proof of Theorem \ref{theorem-ls-regret}}

\subsection{Key Lemma}
The next lemma provides a bound on the term $\sum_{t=1}^{T} \sum_{i=1}^{n} \left\| \bmh_{i}(t) \left(  \bmh_{i}(t)^\tp \bmx_{i}(t) - z_{i}(t)  \right)  \right\|^2$, which is shown to be crucial in establishing the regret of Algorithm \ref{alg-odls}. Moreover, a bound on the accumulative disagreement among all the nodes in the network over the total number of $T$ rounds is established as well. The disagreement is measured by the overall distance of the states of nodes to the average state defined by:
\begin{equation*}
\begin{aligned}
\bmx_{\mathrm{avg}}(t) \triangleq  \frac{1}{n} \sum_{i=1}^{n}  \bmx_{i}(t)  ,\qquad\qquad   t=1,\ldots,T.
\end{aligned}
\end{equation*}

\begin{lemma}\label{lemma-ls-grad}
Let Assumption \ref{assumption-full-info} hold. If $0 < \beta \leq 1$, then along Algorithm \ref{alg-odls} there holds
\BIT
\item[(i)]
$
\sum_{t=1}^{T} \sum_{i=1}^{n} \left\| \bmh_{i}(t) \left(  \bmh_{i}(t)^\tp \bmx_{i}(t) - z_{i}(t)  \right)  \right\|^2
\leq C_1 \cdot T^{\beta} + C_2  \cdot T.
$
\item[(ii)]
$
\sum_{t=1}^{T} \sum_{i=1}^{n} \big\| \bmx_{i}(t) - \bmx_{\mathrm{avg}}(t)  \big\|
\leq  C_3 + C_4 \cdot T^{\frac{1-\beta}{2}} + C_5 \cdot  T^{1-\beta}
$
\EIT
where
\begin{equation*}
\begin{aligned}
C_1 &= \frac{2^\beta  }{2^\beta - 1} \alpha_{\bmh}^2     \sum_{i=1}^{n} \| \bmx_{i}(1) - \bmy^\star_{\mathrm{LS}} \|^2  \nn\\
C_2  &= \frac{2^{\beta} }{2^{\beta} - 1} \alpha_{\bmh}  n (\theta^{\star})^2  \\
C_3 &=   \sum_{i=1}^{n}  \left\| \bmx_{i}(1) - \bmx_{\mathrm{avg}}(1) \right\| +    \sqrt{n}  \frac{\sigma_2(W_{\mathrm{G}})}{1 - \sigma_2(W_{\mathrm{G}})}  \left( \sum_{i=1}^{n}  \| \bmx_{i}(1)  \| \right)  \\
C_4 &=  n  \frac{  \sigma_2(W_{\mathrm{G}})}{\alpha_{\bmh} ( 1 - \sigma_2(W_{\mathrm{G}}) ) }   \sqrt{C_1}   \\
C_5 &= n  \frac{  \sigma_2(W_{\mathrm{G}})}{\alpha_{\bmh} ( 1 - \sigma_2(W_{\mathrm{G}}) ) }   \sqrt{C_2}.
\end{aligned}
\end{equation*}
\end{lemma}

\noindent{\em Proof.}
(i) Denote $\eta = \frac{1}{\alpha_{\bmh} T^\beta}$. We establish the bound by deriving the general evolution of $\sum_{i=1}^{n} \| \bmx_{i}(t+1) - \bmy^\star_{\mathrm{LS}} \|^2$,
\begin{equation}
\begin{aligned}[b]
\sum_{i=1}^{n} \| \bmx_{i}(t+1) - \bmy^\star_{\mathrm{LS}} \|^2
&= \sum_{i=1}^{n} \Bigg\| \sum_{j=1}^{n} [W_{\mathrm{G}}]_{ij} \mathpzc{l}_{j}(t) - \bmy^\star_{\mathrm{LS}} \Bigg\|^2
\leq  \sum_{i=1}^{n} \sum_{j=1}^{n} [W_{\mathrm{G}}]_{ij} \big\|  \mathpzc{l}_{j}(t) - \bmy^\star_{\mathrm{LS}} \big\|^2  \\
&= \sum_{j=1}^{n} \left( \sum_{i=1}^{n} [W_{\mathrm{G}}]_{ij} \right)  \big\|  \mathpzc{l}_{j}(t) - \bmy^\star_{\mathrm{LS}} \big\|^2
= \sum_{j=1}^{n} \big\| \mathpzc{l}_{j}(t) - \bmy^\star_{\mathrm{LS}} \big\|^2
\label{lemma-ls-grad-3}
\end{aligned}
\end{equation}
where the first inequality follows from the convexity of norm square function and doubly stochasticity of $W_{\mathrm{G}}$, \ie, $\sum_{j=1}^{n} [W_{\mathrm{G}}]_{ij} = 1$ and the last equality from $\sum_{i=1}^{n} [W_{\mathrm{G}}]_{ij} = 1$. Using the update in Algorithm \ref{alg-odls}, we have
\begin{equation}
\begin{aligned}[b]
\sum_{i=1}^{n} \| \bmx_{i}(t+1) - \bmy^\star_{\mathrm{LS}} \|^2
&\leq \sum_{i=1}^{n} \big\| \bmx_{i}(t) - \eta \cdot \bmh_{i}(t) \left(  \bmh_{i}(t)^\tp \bmx_{i}(t) - z_{i}(t)  \right) - \bmy^\star_{\mathrm{LS}} \big\|^2  \\
&= \sum_{i=1}^{n} \| \bmx_{i}(t) - \bmy^\star_{\mathrm{LS}} \|^2 + \sum_{i=1}^{n}  \left\| \eta \cdot \bmh_{i}(t) \left(  \bmh_{i}(t)^\tp \bmx_{i}(t) - z_{i}(t)  \right) \right\|^2  \\
&\quad - 2 \eta \sum_{i=1}^{n} \left( \bmx_{i}(t) - \bmy^\star_{\mathrm{LS}} \right)^\tp \bmh_{i}(t) \left(  \bmh_{i}(t)^\tp \bmx_{i}(t) - z_{i}(t)  \right)  \\
&\leq \sum_{i=1}^{n} \| \bmx_{i}(t) - \bmy^\star_{\mathrm{LS}} \|^2 + \eta^2  \sum_{i=1}^{n}  \left\|  \bmh_{i}(t) \left(  \bmh_{i}(t)^\tp \bmx_{i}(t) - z_{i}(t)  \right) \right\|^2  \\
&\quad - 2 \eta \sum_{i=1}^{n} \left[ \theta_{i,t} \left( \bmx_{i}(t) \right) - \theta_{i,t} \left( \bmy^\star_{\mathrm{LS}} \right) \right]
\label{lemma-ls-grad-4}
\end{aligned}
\end{equation}
where the last inequality is based on the convexity of function $\theta_{i,t}(\cdot)$ (since $\nabla^2 \theta_{i,t}(\cdot) = \bmh_{i}(t) \bmh_{i}(t)^\tp$ is positive semi-definite) and the fact that $$
\bmh_{i}(t) \left(  \bmh_{i}(t)^\tp \bmx_{i}(t) - z_{i}(t)  \right) =  \nabla \theta_{i,t}(\bmx_{i}(t)).
$$ Rearranging the terms and dividing both sides by $(2 \eta)$, gives
\begin{equation}
\begin{aligned}[b]
&\sum_{i=1}^{n} \left[ \theta_{i,t} \left( \bmx_{i}(t) \right) - \theta_{i,t} \left( \bmy^\star_{\mathrm{LS}} \right)  \right] \leq \frac{1}{2 \eta} \left( \sum_{i=1}^{n} \| \bmx_{i}(t) - \bmy^\star_{\mathrm{LS}} \|^2 - \sum_{i=1}^{n} \| \bmx_{i}(t+1) - \bmy^\star_{\mathrm{LS}} \|^2 \right)\nonumber\\
&\quad   +  \frac{\eta}{2 } \sum_{i=1}^{n}  \left\|  \bmh_{i}(t) \left(  \bmh_{i}(t)^\tp \bmx_{i}(t) - z_{i}(t)  \right) \right\|^2
\label{lemma-ls-grad-5}
\end{aligned}
\end{equation}
summing the inequalities in (\ref{lemma-ls-grad-5}) over $t=1$ to $t=T$, we obtain
\begin{equation}
\begin{aligned}[b]
&\sum_{t=1}^{T} \sum_{i=1}^{n} \theta_{i,t} \left( \bmx_{i}(t) \right) - \sum_{t=1}^{T} \sum_{i=1}^{n} \theta_{i,t} \left( \bmy^\star_{\mathrm{LS}} \right)  \\
&\leq \frac{1}{2 \eta}  \sum_{t=1}^{T} \left( \sum_{i=1}^{n} \| \bmx_{i}(t) - \bmy^\star_{\mathrm{LS}} \|^2 - \sum_{i=1}^{n} \| \bmx_{i}(t+1) - \bmy^\star_{\mathrm{LS}} \|^2 \right)
 \\
& \quad + \frac{\eta}{2 } \sum_{t=1}^{T} \sum_{i=1}^{n}  \left\|  \bmh_{i}(t) \left(  \bmh_{i}(t)^\tp \bmx_{i}(t) - z_{i}(t)  \right) \right\|^2  \\
&\leq \frac{1}{2 \eta}    \sum_{i=1}^{n} \| \bmx_{i}(1) - \bmy^\star_{\mathrm{LS}} \|^2
+  \frac{\eta}{2 } \sum_{t=1}^{T} \sum_{i=1}^{n}  \left\|  \bmh_{i}(t) \left(  \bmh_{i}(t)^\tp \bmx_{i}(t) - z_{i}(t)  \right) \right\|^2.
\label{lemma-ls-grad-5a}
\end{aligned}
\end{equation}
We now use the self-boundedness property of function $\theta_{i,t}(\cdot)$, that is,
\begin{equation}
\begin{aligned}[b]
\left\| \bmh_{i}(t) \left(  \bmh_{i}(t)^\tp \bmy - z_{i}(t)  \right) \right\|^2
&\leq  2 \left\| \bmh_{i}(t) \right\|^2 \cdot  \theta_{i,t}(\bmy)
\leq 2 \alpha_{\bmh} \theta_{i,t}(\bmy) , \quad \forall \bmy\in\reals^m.
\label{lemma-ls-grad-5b-self-bound}
\end{aligned}
\end{equation}
This further leads to
\begin{equation}
\begin{aligned}[b]
&\sum_{t=1}^{T} \sum_{i=1}^{n}  \left\|  \bmh_{i}(t) \left(  \bmh_{i}(t)^\tp \bmx_{i}(t) - z_{i}(t)  \right) \right\|^2
\leq  \frac{1}{\eta \left(  \alpha_{\bmh}^{-1}  - \eta \right)}    \sum_{i=1}^{n} \| \bmx_{i}(1) - \bmy^\star_{\mathrm{LS}} \|^2 \nonumber\\
&\quad
+ \frac{2}{  \alpha_{\bmh}^{-1}  - \eta } \sum_{t=1}^{T} \sum_{i=1}^{n} \theta_{i,t} \left( \bmy^\star_{\mathrm{LS}} \right)
\label{lemma-ls-grad-6}
\end{aligned}
\end{equation}
substituting $\eta = \frac{1}{\alpha_{\bmh} T^\beta}$ into (\ref{lemma-ls-grad-6}), we have that for all $T\geq 2$,
\begin{equation}
\begin{aligned}[b]
&\sum_{t=1}^{T} \sum_{i=1}^{n}  \left\|  \bmh_{i}(t) \left(  \bmh_{i}(t)^\tp \bmx_{i}(t) - z_{i}(t)  \right) \right\|^2
\leq  \frac{2^\beta}{2^\beta - 1}   \alpha_{\bmh}^2  \sum_{i=1}^{n} \| \bmx_{i}(1) - \bmy^\star_{\mathrm{LS}} \|^2 \cdot T^{\beta} \\
& \quad  + \frac{2^{\beta} }{2^{\beta} - 1}  n (\theta^{\star})^2 \alpha_{\bmh}  \cdot T
\label{lemma-ls-grad-6a}
\end{aligned}
\end{equation}
where we used the assumption of $\left|  \bmh_{i}(t)^\tp \bmy_{\mathrm{LS}}^{\star}  - z_{i}(t) \right| \leq \theta^{\star}$ and the inequality $\frac{T^\beta}{T^\beta - 1}  \leq  \frac{2^\beta}{2^\beta - 1}$ for $T\geq 2$. This further implies that for all $T\geq 2$,
\begin{equation}
\begin{aligned}[b]
\sum_{t=1}^{T} \sum_{i=1}^{n}  \left\|  \bmh_{i}(t) \left(  \bmh_{i}(t)^\tp \bmx_{i}(t) - z_{i}(t)  \right) \right\|
&\leq  \sqrt{ n T \left( C_1 \cdot T^\beta + C_2 \cdot T \right) }  \\
& \leq  \sqrt{n C_1} \cdot T^{\frac{1 + \beta}{2}} + \sqrt{n C_2} \cdot T
\label{lemma-ls-grad-7}
\end{aligned}
\end{equation}
where we used the inequality that $\sqrt{a+b} \leq \sqrt{a} + \sqrt{b} $ for any two non-negative scalars $a$ and $b$.

(ii)
We first derive the general evolution of the average state $\bmx_{\mathrm{avg}}(t+1)$ ($t\geq 1$):
\begin{equation}
\begin{aligned}[b]
\bmx_{\mathrm{avg}}(t+1)= \frac{1}{n} \sum_{i=1}^{n} \sum_{j=1}^{n} [W_{\mathrm{G}}]_{ij} \mathpzc{l}_{j}(t)
= \frac{1}{n} \sum_{j=1}^{n} \Big( \sum_{i=1}^{n} [W_{\mathrm{G}}]_{ij} \Big) \mathpzc{l}_{j}(t)
= \frac{1}{n} \sum_{i=1}^{n} \mathpzc{l}_{i}(t)
\label{lemma-ls-grad-8}
\end{aligned}
\end{equation}
where the second-to-last equality follows from $\sum_{i=1}^{n} [W_{\mathrm{G}}]_{ij} = 1$. Substituting the expression for $\mathpzc{l}_{i,t}$ into (\ref{lemma-ls-grad-8}), we have
\begin{equation}
\begin{aligned}[b]
\bmx_{\mathrm{avg}}(t+1) &= \frac{1}{n} \sum_{i=1}^{n} \Big( \bmx_{i}(t) - \eta\cdot\underbrace{\bmh_{i}(t) \Big(  \bmh_{i}(t)^\tp \bmx_{i}(t) - z_{i}(t)  \Big)}_{ \triangleq \mathbf{s}_{i}(t)}  \Big)\nonumber\\
&= \bmx_{\mathrm{avg}}(t)  -  \eta \cdot \frac{1}{n} \sum_{i=1}^{n} \mathbf{s}_{i}(t)  \\
&= \bmx_{\mathrm{avg}}(1)  - \eta \sum_{\ell=1}^{t}  \frac{1}{n} \sum_{i=1}^{n} \mathbf{s}_{i}(\ell).
\label{lemma-ls-grad-9}
\end{aligned}
\end{equation}
On the other hand, we can write the general evolution of $ \bmx_{i}(t+1)$ as
\begin{equation}
\begin{aligned}[b]
\bmx_{i}(t+1)
=  \sum_{j=1}^{n} [W_{\mathrm{G}}]_{ij} \left(  \bmx_{j}(t) -  \eta\cdot\mathbf{s}_{j}(t) \right)
= \sum_{j=1}^{n} [W_{\mathrm{G}}^t]_{ij} \bmx_{j}(1)  - \eta \sum_{\ell = 1}^{t}  \sum_{j=1}^{n} [W_{\mathrm{G}}^{t+1-\ell}]_{ij} \mathbf{s}_{j}(\ell).
\label{lemma-ls-grad-10}
\end{aligned}
\end{equation}
Combining equations (\ref{lemma-ls-grad-9}) and (\ref{lemma-ls-grad-10}), we obtain
\begin{equation}
\begin{aligned}[b]
&\sum_{i=1}^{n} \big\| \bmx_{i}(t+1) - \bmx_{\mathrm{avg}}(t+1)  \big\|  \leq  \sum_{i=1}^{n} \sum_{j=1}^{n} \left|    [W_{\mathrm{G}}^t]_{ij}  - \frac{1}{n}  \right|  \cdot \|   \bmx_{j}(1)  \| \nonumber\\
&\quad + \eta \sum_{i=1}^{n} \sum_{\ell = 1}^{t}  \sum_{j=1}^{n} \left|    [W_{\mathrm{G}}^{t+1-\ell}]_{ij}  - \frac{1}{n}  \right|  \cdot \left\| \bmh_{j}(\ell) \left(  \bmh_{j}(\ell)^\tp \bmx_{j}(\ell) - z_{j}(\ell)  \right)  \right\| \\
&= \sum_{j=1}^{n} \left( \sum_{i=1}^{n} \left|    [W_{\mathrm{G}}^t]_{ij}  - \frac{1}{n}  \right|  \right) \|   \bmx_{j}(1)  \| \nonumber\\
&\quad
+ \eta \sum_{j=1}^{n} \sum_{\ell = 1}^{t}  \left( \sum_{i=1}^{n} \left|    [W_{\mathrm{G}}^{t+1-\ell}]_{ij}  - \frac{1}{n}  \right|  \right) \left\| \bmh_{j}(\ell) \left(  \bmh_{j}(\ell)^\tp \bmx_{j}(\ell) - z_{j}(\ell)  \right)  \right\|
\label{lemma-ls-grad-11}
\end{aligned}
\end{equation}
which, combined with the following inequality,
\begin{equation}
\begin{aligned}[b]
\sum_{i=1}^{n} \left|    [W_{\mathrm{G}}^t]_{ij}  - \frac{1}{n}  \right|
&\leq \left\|    W_{\mathrm{G}}^t  - \frac{\ones_n \ones_n^{\tp}}{n}  \right\|_1
\leq  \sqrt{n}  \left\|    W_{\mathrm{G}}^t  - \frac{\ones_n \ones_n^{\tp}}{n}  \right\|
\leq  \sqrt{n}    \sigma_2(W_{\mathrm{G}})^t
\label{lemma-ls-grad-12}
\end{aligned}
\end{equation}
gives that the following inequality holds for all $t\geq 2$,
\begin{equation}
\begin{aligned}[b]
&\sum_{i=1}^{n} \big\| \bmx_{i}(t) - \bmx_{\mathrm{avg}}(t)  \big\| \\
&\leq  \sqrt{n}    \left( \sum_{i=1}^{n}  \| \bmx_{i}(1)  \| \right) \sigma_2(W_{\mathrm{G}})^{t-1}
+  \sqrt{n} \eta  \sum_{\ell = 1}^{t-1}  \sigma_2(W_{\mathrm{G}})^{t-\ell}    \sum_{i=1}^{n} \left\| \bmh_{i}(\ell) \left(  \bmh_{i}(\ell)^\tp \bmx_{i}(\ell) - z_{i}(\ell)  \right)  \right\|.
\label{lemma-ls-grad-13}
\end{aligned}
\end{equation}
Summing the inequalities in (\ref{lemma-ls-grad-13}) over $t=1$ to $t=T$, gives
\begin{equation}
\begin{aligned}[b]
\sum_{t=1}^{T} \sum_{i=1}^{n} \big\| \bmx_{i}(t) - \bmx_{\mathrm{avg}}(t)  \big\|
&= \sum_{i=1}^{n} \big\| \bmx_{i}(1) - \bmx_{\mathrm{avg}}(1)  \big\| + \sum_{t=2}^{T} \sum_{i=1}^{n} \big\| \bmx_{i}(t) - \bmx_{\mathrm{avg}}(t)  \big\|  \\
&\leq  \sum_{i=1}^{n} \big\| \bmx_{i}(1) - \bmx_{\mathrm{avg}}(1)  \big\| +  \sqrt{n}    \left( \sum_{i=1}^{n}  \| \bmx_{i}(1)  \| \right)  \sum_{t=2}^{T}  \sigma_2(W_{\mathrm{G}})^{t-1} \\
&\quad+  \sqrt{n} \eta  \sum_{t=2}^{T} \sum_{\ell = 1}^{t-1}  \sigma_2(W_{\mathrm{G}})^{t-\ell}    \sum_{i=1}^{n} \left\| \bmh_{i}(\ell) \left(  \bmh_{i}(\ell)^\tp \bmx_{i}(\ell) - z_{i}(\ell)  \right)  \right\| .
\label{lemma-ls-grad-14}
\end{aligned}
\end{equation}
The second term on the right-hand side of (\ref{lemma-ls-grad-14}) can be bounded as follows:
\begin{equation}
\begin{aligned}[b]
\sqrt{n}    \left( \sum_{i=1}^{n}  \| \bmx_{i}(1)  \| \right)  \sum_{t=2}^{T}  \sigma_2(W_{\mathrm{G}})^{t-1}
&\leq \sqrt{n}  \frac{\sigma_2(W_{\mathrm{G}})}{1 - \sigma_2(W_{\mathrm{G}})}  \left( \sum_{i=1}^{n}  \| \bmx_{i}(1)  \| \right)
\label{lemma-ls-grad-15}
\end{aligned}
\end{equation}
where the last inequality follows from $\sigma_2(W_{\mathrm{G}})  < 1$. On the other hand, the last term can be bounded in the following way,
\begin{equation}
\begin{aligned}
&\sum_{t=2}^{T} \sum_{\ell = 1}^{t-1}  \sigma_2(W_{\mathrm{G}})^{t-\ell}    \sum_{i=1}^{n} \left\| \bmh_{i}(\ell) \left(  \bmh_{i}(\ell)^\tp \bmx_{i}(\ell) - \bmz_{i}(\ell)  \right) \right\| \nonumber\\
&\leq \sum_{t=1}^{T-1} \left(  \sum_{\ell = 1}^{T-1}  \sigma_2(W_{\mathrm{G}})^{\ell} \right)    \sum_{i=1}^{n} \left\| \bmh_{i}(t) \left(  \bmh_{i}(t)^\tp \bmx_{i}(t) - z_{i}(t)  \right) \right\|  \\
&\leq \frac{\sigma_2(W_{\mathrm{G}})}{1 - \sigma_2(W_{\mathrm{G}})}  \sum_{t=1}^{T-1} \sum_{i=1}^{n} \left\| \bmh_{i}(t) \left(  \bmh_{i}(t)^\tp \bmx_{i}(t) - z_{i}(t)  \right) \right\|   \\
&\leq  \frac{  \sigma_2(W_{\mathrm{G}})}{1 - \sigma_2(W_{\mathrm{G}})}  \left( \sqrt{n C_1} \cdot T^{\frac{1 + \beta}{2}} + \sqrt{n C_2} \cdot T  \right)
\label{lemma-ls-grad-16}
\end{aligned}
\end{equation}
where the last inequality is based on (\ref{lemma-ls-grad-7}). Hence, the desired estimate follows by combining the results in (\ref{lemma-ls-grad-14}), (\ref{lemma-ls-grad-15}) and (\ref{lemma-ls-grad-16}), and using $\eta = \frac{1}{\alpha_{\bmh}  T^\beta }$. The proof is complete.
\hfill$\square$

The accumulated disagreement over $T$ rounds grows as $\mathcal{O} (T^{1-\beta})$, which is a sublinear function of $T$. This means that when $\beta > 0 $, one has
\begin{equation*}
\begin{aligned}
\lim_{T\rightarrow\infty} \frac{1}{T} \sum_{t=1}^{T} \sum_{i=1}^{n} \big\| \bmx_{i}(t) - \bmx_{\mathrm{avg}}(t)  \big\|
=  \lim_{T\rightarrow\infty}    \mathcal{O} \left( T^{-\beta} \right) = 0.
\end{aligned}
\end{equation*}

\subsection{Proof of the theorem}
From (\ref{lemma-ls-grad-5a}), we have
\begin{equation}
\begin{aligned}[b]
&\sum_{t=1}^{T} \sum_{j=1}^{n} \theta_{j,t} \left( \bmx_{j}(t) \right) - \sum_{t=1}^{T} \sum_{j=1}^{n} \theta_{j,t} \left( \bmy^\star_{\mathrm{LS}} \right) \nonumber\\
&\leq \frac{1}{2 \eta}    \sum_{j=1}^{n} \| \bmx_{j}(1) - \bmy^\star_{\mathrm{LS}} \|^2
+  \frac{\eta}{2 } \sum_{t=1}^{T} \sum_{j=1}^{n}  \left\|  \bmh_{j}(t) \left(  \bmh_{j}(t)^\tp \bmx_{j}(t) - z_{j}(t)  \right) \right\|^2  \\
&\leq  \frac{\alpha_{\bmh}}{2 }   \sum_{j=1}^{n} \| \bmx_{j}(1) - \bmy^\star_{\mathrm{LS}} \|^2 \cdot T^\beta
+  \frac{1}{2 \alpha_{\bmh}} C_1  + \frac{1}{2 \alpha_{\bmh}} C_2 \cdot T^{1 - \beta}
\label{theorem-ls-regret-1}
\end{aligned}
\end{equation}
where the last inequality is based on Lemma \ref{lemma-ls-grad} and $\eta = \frac{1}{\alpha_{\bmh}  T^{\beta}}$. We now turn our attention to bounding the first term on the left-hand side, \ie, $\sum_{t=1}^{T} \sum_{j=1}^{n} \theta_{j,t} \left( \bmx_{j}(t) \right)$:
\begin{equation}
\begin{aligned}[b]
\sum_{t=1}^{T} \sum_{j=1}^{n} \theta_{j,t} \left( \bmx_{j}(t) \right)
&= \sum_{t=1}^{T} \sum_{j=1}^{n}  \frac{1}{2}  \Big|  \bmh_{j}(t)^\tp \bmx_{j}(t)  - z_{j}(t) \Big|^2 \nonumber\\
&= \sum_{t=1}^{T} \sum_{j=1}^{n}  \frac{1}{2}  \Big|  \bmh_{j}(t)^\tp \left( \bmx_{i}(t) + \bmx_{j}(t) - \bmx_{i}(t) \right) - z_{j}(t) \Big|^2  \\
&= \sum_{t=1}^{T} \sum_{j=1}^{n}  \frac{1}{2}  \Big|  \bmh_{j}(t)^\tp  \bmx_{i}(t)  - z_{j}(t) +  \bmh_{j}(t)^\tp \left( \bmx_{j}(t) - \bmx_{i}(t) \right)  \Big|^2
\label{theorem-ls-regret-2}
\end{aligned}
\end{equation}
now, by applying the following inequality to (\ref{theorem-ls-regret-2}),
\begin{equation}
\begin{aligned}
(a + b)^2 \geq \frac{1}{1+A} a^2 - \frac{1}{A} b^2, \qquad\qquad \forall A > 0 \ \mathrm{and}\ a,b\in\reals
\label{theorem-ls-regret-3}
\end{aligned}
\end{equation}
we further obtain (by setting $A = T^{-\gamma}$ with $\gamma > 0$)
\begin{equation}
\begin{aligned}[b]
&\sum_{t=1}^{T} \sum_{j=1}^{n} \theta_{j,t} \left( \bmx_{j}(t) \right)
\geq \frac{1}{1 + T^{-\gamma}}  \sum_{t=1}^{T} \sum_{j=1}^{n}  \frac{1}{2}  \Big|  \bmh_{j}(t)^\tp \bmx_{i}(t)  - z_{j}(t) \Big|^2 \nonumber\\
& \quad
-   T^{\gamma }  \sum_{t=1}^{T} \sum_{j=1}^{n}  \frac{1}{2}  \Big|  \bmh_{j}(t)^\tp \left( \bmx_{j}(t) - \bmx_{i}(t) \right)  \Big|^2  \\
&=  \frac{1}{1 + T^{-\gamma}} \sum_{t=1}^{T} \sum_{j=1}^{n}  \theta_{j,t} \left( \bmx_{i}(t) \right)
-   T^{\gamma }   \underbrace{\sum_{t=1}^{T} \sum_{j=1}^{n}  \frac{1}{2}  \Big|  \bmh_{j}(t)^\tp \left( \bmx_{j}(t) - \bmx_{i}(t) \right)  \Big|^2}_{\triangleq \mathcal{R}(T)}.
\label{theorem-ls-regret-4}
\end{aligned}
\end{equation}
We are left to bound term $\mathcal{R}(T)$:
\begin{equation}
\begin{aligned}[b]
\mathcal{R}(T) &\leq \sum_{t=1}^{T} \sum_{j=1}^{n}  \frac{1}{2} \big\| \bmh_{j}(t) \big\|^2 \cdot \big\| \bmx_{j}(t) - \bmx_{i}(t) \big\|^2  \\
&\leq   \alpha_{\bmh} \sum_{t=1}^{T} \sum_{j=1}^{n}  \big\| \bmx_{j}(t) - \bmx_{\mathrm{avg}}(t) \big\|^2 +  \alpha_{\bmh} \sum_{t=1}^{T} \sum_{j=1}^{n}  \big\| \bmx_{i}(t) - \bmx_{\mathrm{avg}}(t) \big\|^2           \\
&\leq  \alpha_{\bmh} (1+n) \sum_{t=1}^{T} \sum_{j=1}^{n}  \big\| \bmx_{j}(t) - \bmx_{\mathrm{avg}}(t) \big\|^2   \\
&\leq 2  \alpha_{\bmh} n \left(  \sum_{t=1}^{T} \sum_{i=1}^{n}  \big\| \bmx_{i}(t) - \bmx_{\mathrm{avg}}(t) \big\| \right)^2 \\
&\leq  6 \alpha_{\bmh} n \left(  C_3^2 +  C_4^2 \cdot T^{1-\beta} +   C_5^2 \cdot T^{2(1-\beta)}  \right)
\label{theorem-ls-regret-5}
\end{aligned}
\end{equation}
where the first inequality follows from the Cauchy-Schwarz inequality $\mathbf{a}^\tp \mathbf{b} \leq \| \mathbf{a} \| \cdot \| \mathbf{b} \|$ for any $\mathbf{a},\mathbf{b} \in \reals^m$, the third inequality from $\| \bmx_{\mathrm{avg}}(t) - \bmx_{i}(t) \|^2 \leq \sum_{j=1}^{n} \| \bmx_{j}(t) - \bmx_{\mathrm{avg}}(t) \|^2$, and the last inequality from Lemma \ref{lemma-ls-grad}(ii) and the inequality that $(\sum_{i=1}^{n} a_i) ^2  \leq  n \sum_{i=1}^{n} a_i^2  $ for any $a_i \in \reals$, $i=1,\ldots,n$. Combining the inequalities (\ref{theorem-ls-regret-1}), (\ref{theorem-ls-regret-4}) and (\ref{theorem-ls-regret-5}), we find that
\begin{equation}
\begin{aligned}[b]
\left( 1-   \frac{1}{1 + T^{\gamma}}  \right)  \sum_{t=1}^{T} \sum_{j=1}^{n}  \theta_{j,t} \left( \bmx_{i}(t) \right)
- \sum_{t=1}^{T} \sum_{j=1}^{n} \theta_{j,t} \left( \bmy^\star_{\mathrm{LS}} \right)
\leq \frac{1}{2 \alpha_{\bmh}} C_1  + \frac{\alpha_{\bmh}}{2 }   \sum_{j=1}^{n} \| \bmx_{j}(1) - \bmy^\star_{\mathrm{LS}} \|^2 \cdot T^\beta & \\
+   \frac{1}{2 \alpha_{\bmh}} C_2 \cdot T^{1 - \beta}
+ 6 \alpha_{\bmh} n C_3^2 \cdot T^{\gamma} + 6 \alpha_{\bmh} n C_4^2 \cdot T^{1-\beta+\gamma} +  6 \alpha_{\bmh} n C_5^2 \cdot T^{2(1-\beta)+\gamma}. &
\label{theorem-ls-regret-7}
\end{aligned}
\end{equation}
We need to balance between the twin goals of making $\gamma$ large and minimizing the overall bound as far as possible. It is easy to see that the bound in (\ref{theorem-ls-regret-7}) is dominated by the terms $T^\beta$ and $T^{2(1-\beta)+\gamma}$ (from $T^{2(1-\beta)+\gamma}$ one must have $\beta > \frac{1}{2}$ to get a sublinear function of $T$); hence, by setting $\beta = 2(1-\beta)+\gamma$ leads to the optimal choice of $\gamma = 3 \beta - 2$. We also have the condition on $\beta$: $\frac{2}{3} < \beta <1$, because $\gamma > 0$ ($\beta < 1$ is needed to guarantee that the bound is a sublinear function of $T$). Therefore, we conclude that by setting $\gamma = 3 \beta - 2$, the right-hand side on (\ref{theorem-ls-regret-7}) achieves the optimal bound, that is,
\begin{equation}
\begin{aligned}[b]
\left( 1-   \frac{1}{1 + T^{3 \beta - 2}}  \right)  \sum_{t=1}^{T} \sum_{j=1}^{n}  \theta_{j,t} \left( \bmx_{i}(t) \right)
- \sum_{t=1}^{T} \sum_{j=1}^{n} \theta_{j,t} \left( \bmy^\star_{\mathrm{LS}} \right)
&\leq  C_6 \cdot T^\beta
\label{theorem-ls-regret-8}
\end{aligned}
\end{equation}
where $C_6 =  \frac{\alpha_{\bmh}}{2 }   \sum_{i=1}^{n} \| \bmx_{i}(1) - \bmy^\star_{\mathrm{LS}} \|^2  + \frac{1}{2 \alpha_{\bmh}} ( C_1 +    C_2 )  + 6 \alpha_{\bmh} n \left(  C_3^2  +  C_4^2  +  C_5^2 \right)$. We further have the following inequality by doing some simple algebra,
\begin{equation}
\begin{aligned}[b]
&\sum_{t=1}^{T} \sum_{j=1}^{n}  \theta_{j,t} \left( \bmx_{i}(t) \right)
- \sum_{t=1}^{T} \sum_{j=1}^{n} \theta_{j,t} \left( \bmy^\star_{\mathrm{LS}} \right)\nonumber\\
&\leq  \frac{1}{ 1-   \frac{1}{1 + T^{3 \beta - 2}} }  C_6 \cdot T^\beta  + \frac{\frac{1}{1 + T^{3 \beta - 2}}}{ 1-   \frac{1}{1 + T^{3 \beta - 2}} }  \sum_{t=1}^{T} \sum_{j=1}^{n} \theta_{j,t} \left( \bmy^\star_{\mathrm{LS}} \right)  \\
&\leq  C_6  \left( 1 + T^{2-3\beta} \right)  \cdot  T^\beta + T^{2-3\beta} \cdot \sum_{t=1}^{T} \sum_{j=1}^{n} \theta_{j,t} \left( \bmy^\star_{\mathrm{LS}} \right)    \\
&\leq C_6 \cdot T^\beta  + C_6 \cdot T^{2(1-\beta)} + \frac{1}{2} n (\theta^{\star})^2 \cdot T^{3(1-\beta)}
\label{theorem-ls-regret-9}
\end{aligned}
\end{equation}
where in the last inequality we used the assumption of $\left|  \bmh_{j}(t)^\tp \bmy_{\mathrm{LS}}^{\star}  - z_{j}(t) \right| \leq \theta^{\star}$. It is easy to see that the optimal regret bound is achieved when $\beta  =  3(1-\beta)$, which gives the optimal choice of $\beta = \frac{3}{4}$. The left-hand side on (\ref{theorem-ls-regret-9}) is just $\reg_{\mathrm{LS}}(i,T)$, hence we arrive at the conclusion that
\begin{equation*}
\begin{aligned}
\reg_{\mathrm{LS}}(i,T)  &\leq  \left(  \frac{1}{2} n (\theta^{\star})^2 +  2 C_6  \right) \cdot T^{3/4} ,
\qquad\qquad T\geq 2.
\end{aligned}
\end{equation*}
This completes the proof of Theorem \ref{theorem-ls-regret}.
\hfill$\square$





\newpage

\section{Proof of Theorem \ref{theorem-ls-bandit}}

\subsection{Key Lemmas}
To facilitate our analysis, define a smooth function of the form:
\begin{equation}
\begin{aligned}
\hat{\theta}_{i,t}(\bmy)   \triangleq \expect_{\mathbf{v}\in\mathbb{B}_1^m} \big[ \theta_{i,t}(\bmy  +  \epsilon  \mathbf{v}) \big]
\label{smoothed-func}
\end{aligned}
\end{equation}
where $\mathbf{v}\in\reals^{m}$ is a vector selected uniformly at random from the unit ball $\mathbb{B}_1^m$. The next lemma establishes some nice connections between the functions $\theta_{i,t}(\bmy) $ and $\hat{\theta}_{i,t}(\bmy)$, as well as the connections between $\nabla \hat{\theta}_{i,t}(\bmx_{i}(t))$ and $\mathpzc{g}_{i,t} ( \bmx_{i}(t) )$. This lemma is of critical importance in our regret analysis.
\begin{lemma}\label{lemma-bandit-grad-est}
Let $\mathcal{F}_t$ be the be the $\sigma$-field generated by the entire history of the random variables to time $t$. Then for all $i\in\mathrm{V}$ and $t=1,\ldots,T$, we have the following.
\BIT
\item[(i)]
The gradient estimator $\mathpzc{g}_{i,t}$ satisfies:
\begin{equation*}
\begin{aligned}
\expect \left[    \mathpzc{g}_{i,t} ( \bmx_{i}(t) ) \mid \mathcal{F}_t \right]
&=   \nabla  \hat{\theta}_{i,t}( \bmx_{i}(t) )  \\
\left\| \mathpzc{g}_{i,t} ( \bmy )  \right\|
&\leq   m  \left\| \bmh_{i}(t) \left(  \bmh_{i}(t)^\tp \bmy - z_{i}(t)  \right) \right\| + m \| \bmh_{i}(t) \|^2 \epsilon,  \qquad  \forall\bmy\in\reals^m.
\end{aligned}
\end{equation*}
\item[(ii)]
The smoothed function $\hat{\theta}_{i,t} $ satisfies:
\begin{equation*}
\begin{aligned}
\big| \hat{\theta}_{i,t} ( \bmy) -  \theta_{i,t} ( \bmy ) \big|
&\leq   \left\| \bmh_{i}(t) \left(  \bmh_{i}(t)^\tp \bmy - z_{i}(t)  \right) \right\|  \epsilon  + \| \bmh_{i}(t) \|^2  \epsilon^2,  \qquad  \forall\bmy\in\reals^m.
\end{aligned}
\end{equation*}
\EIT
\end{lemma}

The proof of Lemma \ref{lemma-bandit-grad-est} frequently utilizes the following lemma that can be easily obtained for any convex function.
\begin{lemma}\label{lemma-bandit-a-simple-lemma}
For a convex function $f: \reals^m \rightarrow  \reals$, we have
\begin{equation*}
\begin{aligned}
\left| f(\bmx)  - f(\bmy)  \right|
&\leq   \max\{ \| \nabla f(\bmx) \| , \| \nabla f(\bmy) \| \}  \cdot \| \bmx  - \bmy\|,  \qquad  \bmx,\bmy\in\reals^m.
\end{aligned}
\end{equation*}
\end{lemma}
\noindent{\em Proof.}
From the convexity of $f$, we have
\begin{equation*}
\begin{aligned}
f(\bmx)  - f(\bmy)   &\geq  \nabla f(\bmy)^\tp ( \bmx - \bmy )
\geq - \max\{ \| \nabla f(\bmx) \| , \| \nabla f(\bmy) \| \}  \cdot \| \bmx  - \bmy\|
\end{aligned}
\end{equation*}
and similarly,
\begin{equation*}
\begin{aligned}
f(\bmx)  - f(\bmy)   &\leq  \nabla f(\bmx)^\tp ( \bmx - \bmy )
\leq \max\{ \| \nabla f(\bmx) \| , \| \nabla f(\bmy) \| \}  \cdot \| \bmx  - \bmy\|
\end{aligned}
\end{equation*}
hence the desired inequality follows by combining the preceding two inequalities.
\hfill$\square$

\noindent{\em Proof of Lemma \ref{lemma-bandit-grad-est}.}
(i) Using the expression for $\mathpzc{g}_{i,t} ( \bmy )$, we have
\begin{equation*}
\begin{aligned}
\left\| \mathpzc{g}_{i,t} ( \bmy )  \right\|
&=  \frac{m}{2 \epsilon} \left\|    \big( \theta_{i,t} ( \bmy  +  \epsilon \bmu_{i}(t) )  -   \theta_{i,t} ( \bmy  -  \epsilon \bmu_{i}(t) ) \big) \cdot \bmu_{i}(t)   \right\|    \\
&\leq  \frac{m}{2 \epsilon}  \left\|   \theta_{i,t} ( \bmy  +  \epsilon \bmu_{i}(t) )  -   \theta_{i,t} ( \bmy  -  \epsilon \bmu_{i}(t) )  \right\|  \cdot \|  \bmu_{i}(t)  \|  \\
&\leq  \frac{m}{2 \epsilon}  \max\{ \| \nabla \theta_{i,t} ( \bmy  +  \epsilon \bmu_{i}(t) ) \| , \| \nabla \theta_{i,t} ( \bmy  -  \epsilon \bmu_{i}(t) ) \| \} \cdot 2 \epsilon \|  \bmu_{i}(t) \|^2  \\
&= m \cdot \max\{ \| \nabla \theta_{i,t} ( \bmy  +  \epsilon \bmu_{i}(t) ) \| , \| \nabla \theta_{i,t} ( \bmy  -  \epsilon \bmu_{i}(t) ) \| \}
\end{aligned}
\end{equation*}
where the second inequality follows from Lemma \ref{lemma-bandit-a-simple-lemma} and the last equality from $\|   \bmu_{i}(t)  \| = 1$. We provide a bound on $\| \nabla \theta_{i,t} ( \bmy  \pm  \epsilon \bmu_{i}(t) ) \|$ as follows:
\begin{equation}
\begin{aligned}[b]
& \left\| \nabla \theta_{i,t} ( \bmy \pm \epsilon \bmu_{i}(t) ) -  \nabla \theta_{i,t} ( \bmy) \right\| \nonumber\\
&= \left\| \bmh_{i}(t) \left(  \bmh_{i}(t)^\tp \left(  \bmy  +  \epsilon \bmu_{i}(t) \right) - z_{i}(t)  \right) -  \bmh_{i}(t) \left(  \bmh_{i}(t)^\tp \bmy - z_{i}(t)  \right) \right\|  \\
&= \epsilon \left\| \bmh_{i}(t) \bmh_{i}(t)^\tp \bmu_{i}(t)  \right\|
\leq \epsilon \left\| \bmh_{i}(t) \bmh_{i}(t)^\tp \right\|   \cdot  \left\| \bmu_{i}(t)  \right\| \nonumber\\
&=   \sigma_{\max}\left(\bmh_{i}(t) \bmh_{i}(t)^\tp\right) \epsilon \nonumber\\
&= \| \bmh_{i}(t) \|^2  \epsilon
\label{lemma-bandit-grad-est-1}
\end{aligned}
\end{equation}
where $\sigma_{\max}(\cdot)$ denotes the largest singular value of a matrix, and the last equality follows from the fact that the largest singular value of $\frac{\bmh_{i}(t)\bmh_{i}(t)^\tp}{\| \bmh_{i}(t) \|^2} $ is one. Hence, the proof is complete by noting
\begin{equation*}
\begin{aligned}
\left\| \nabla \theta_{i,t} ( \bmy  \pm  \epsilon \bmu_{i}(t) ) -  \nabla \theta_{i,t} ( \bmy ) \right\|  \geq \left\| \nabla \theta_{i,t} ( \bmy  \pm  \epsilon \bmu_{i}(t) ) \right\| -  \left\| \nabla \theta_{i,t} ( \bmy ) \right\|
\end{aligned}
\end{equation*}
and $\nabla \theta_{i,t} ( \bmy )  =  \bmh_{i}(t) \left(  \bmh_{i}(t)^\tp \bmy - z_{i}(t)  \right)$.

(ii) Using the definition (\ref{smoothed-func}), we have
\begin{equation*}
\begin{aligned}
\big| \hat{\theta}_{i,t} ( \bmy) -  \theta_{i,t} ( \bmy ) \big|
&= \left| \expect_{\mathbf{v}\in\mathbb{B}_1^m} \big[ \theta_{i,t}(\bmy  +  \epsilon  \mathbf{v}) \big] -  \theta_{i,t} ( \bmy ) \right|\\
&\leq \expect_{\mathbf{v}\in\mathbb{B}_1^m}  \left[ \left| \theta_{i,t}(\bmy  +  \epsilon  \mathbf{v})  -  \theta_{i,t} ( \bmy ) \right| \right] \\
&\leq   \expect_{\mathbf{v}\in\mathbb{B}_1^m}  \left[ \max\{ \| \nabla \theta_{i,t} ( \bmy  +  \epsilon \mathbf{v} ) \| , \| \nabla \theta_{i,t} ( \bmy   ) \| \} \cdot \epsilon \| \mathbf{v} \| \right]
\end{aligned}
\end{equation*}
where in the first inequality we used Jensen's inequality and in the last inequality we recalled Lemma \ref{lemma-bandit-a-simple-lemma}. Then following an argument similar to that of part (i) and using $\|  \mathbf{v}  \| \leq 1$, we get the desired estimate.
\hfill$\square$

As in the full information feedback setting, the analyses of accumulated disagreement and regret rely on the boundedness of $\sum_{t=1}^{T} \sum_{i=1}^{n} \left\| \bmh_{i}(t) \left(  \bmh_{i}(t)^\tp \bmx_{i}(t) - z_{i}(t)  \right)  \right\|^2 $ in expectation.
This turns out to be true, provided that the step size $\eta = \frac{1}{\kappa T^\beta}$ and the parameter $\epsilon$ are chosen appropriately.
\begin{lemma}\label{lemma-ls-grad-bandit}
Let Assumptions \ref{assumption-full-info} hold. Suppose $\alpha_{\bmh} < n$. Let $\kappa >  \frac{2 n m^2 \alpha_{\bmh}}{n - \alpha_{\bmh}}$, $0 < \beta \leq 1$, and $\epsilon  =  \frac{1}{\sqrt{T}}$ for all $t=1,\ldots,T$. Then along Algorithm \ref{alg-odls-bandit} there holds
\BIT
\item[(i)]
$
\sum_{t=1}^{T} \sum_{i=1}^{n} \expect  \left[ \left\| \bmh_{i}(t) \left(  \bmh_{i}(t)^\tp \bmx_{i}(t) - z_{i}(t)  \right)  \right\|^2 \right]  \leq  B_1  + B_2 \cdot T.
$
\item[(ii)]
$
\sum_{t=1}^{T} \sum_{i=1}^{n} \expect \big[  \big\| \bmx_{i}(t) - \bmx_{\mathrm{avg}}(t)  \big\|  \big]
\leq B_3 + B_4 \cdot T^{\frac{1}{2} - \beta}  + B_5  \cdot  T^{1-\beta}
$

\EIT
where
\begin{equation*}
\begin{aligned}
B_1 &= \left(   \frac{1}{\kappa} n  m^2 \alpha_{\bmh}^2  +  2 n \alpha_{\bmh} + \frac{1}{2} n^2  \right) \hat{\kappa}    \\
B_2 &= \left(  \frac{1}{2} \kappa  \sum_{i=1}^{n} \expect  \left[ \| \bmx_{i}(1) - \bmy^\star_{\mathrm{LS}} \|^2 \right]  +  n \theta^{\star} \sqrt{\alpha_{\bmh}}    +  \frac{1}{2}  n (\theta^{\star})^2   \right)  \hat{\kappa}    \\
B_3 &=  \sum_{i=1}^{n} \expect \big[ \big\| \bmx_{i}(1) - \bmx_{\mathrm{avg}}(1)  \big\|  \big]
+  \sqrt{n}   \frac{\sigma_2(W_{\mathrm{G}})}{1 - \sigma_2(W_{\mathrm{G}})}  \left( \sum_{i=1}^{n} \expect [ \| \bmx_{i}(1) \| ] \right)  \\
B_4 &=  n m   \frac{ \sigma_2(W_{\mathrm{G}})}{\kappa (1 - \sigma_2(W_{\mathrm{G}})) }  \sqrt{2 \left( B_1  +  n \alpha_{\bmh}^2 \right)}  \\
B_5 &=  n  m  \frac{ \sigma_2(W_{\mathrm{G}})}{\kappa (1 - \sigma_2(W_{\mathrm{G}})) }  \sqrt{2 B_2}
\end{aligned}
\end{equation*}
with $\hat{\kappa} =   \frac{2 n \alpha_{\bmh}\kappa}{(n - \alpha_{\bmh})\kappa - 2 n m^2 \alpha_{\bmh}}$.
\end{lemma}

\noindent{\em Proof.}
(i) By an argument similar to that of (\ref{lemma-ls-grad-4}), it follows that
\begin{equation}
\begin{aligned}[b]
&\sum_{i=1}^{n} \| \bmx_{i}(t+1) - \bmy^\star_{\mathrm{LS}} \|^2 \\
&\leq \sum_{i=1}^{n} \big\| \bmx_{i}(t) - \eta \cdot \mathpzc{g}_{i,t} ( \bmx_{i}(t) ) - \bmy^\star_{\mathrm{LS}} \big\|^2  \\
&= \sum_{i=1}^{n} \| \bmx_{i}(t) - \bmy^\star_{\mathrm{LS}} \|^2 + \eta^2 \sum_{i=1}^{n}  \left\|  \mathpzc{g}_{i,t} ( \bmx_{i}(t) ) \right\|^2
- 2 \eta \sum_{i=1}^{n} \left( \bmx_{i}(t) - \bmy^\star_{\mathrm{LS}} \right)^\tp  \mathpzc{g}_{i,t} ( \bmx_{i}(t) )
\label{lemma-ls-grad-bandit-1}
\end{aligned}
\end{equation}
taking the conditional expectation on $\mathcal{F}_t$ and using Lemma \ref{lemma-bandit-grad-est}(i), yields
\begin{equation}
\begin{aligned}[b]
&\sum_{i=1}^{n} \expect  \left[   \| \bmx_{i}(t+1) - \bmy^\star_{\mathrm{LS}} \|^2 \mid \mathcal{F}_t \right] \\
&=  \sum_{i=1}^{n} \| \bmx_{i}(t) - \bmy^\star_{\mathrm{LS}} \|^2 + \eta^2 \sum_{i=1}^{n}  \expect  \big[ \left\| \mathpzc{g}_{i,t} ( \bmx_{i}(t) ) \right\|^2  \mid \mathcal{F}_t \big]
- 2 \eta \sum_{i=1}^{n}  \left( \bmx_{i}(t) - \bmy^\star_{\mathrm{LS}} \right)^\tp \nabla  \hat{\theta}_{i,t}( \bmx_{i}(t) )   \\
&\geq \sum_{i=1}^{n} \| \bmx_{i}(t) - \bmy^\star_{\mathrm{LS}} \|^2 + \eta^2 \sum_{i=1}^{n}  \expect  \big[ \left\| \mathpzc{g}_{i,t} ( \bmx_{i}(t) ) \right\|^2  \mid \mathcal{F}_t \big]
- 2 \eta \sum_{i=1}^{n}  \big[  \hat{\theta}_{i,t}( \bmx_{i}(t) )   -   \hat{\theta}_{i,t}( \bmy^\star_{\mathrm{LS}} )  \big]
\label{lemma-ls-grad-bandit-2}
\end{aligned}
\end{equation}
where in the inequality we used the convexity of function $\hat{\theta}_{i,t}(\cdot)$. Rearranging the terms, we have
\begin{equation}
\begin{aligned}[b]
\sum_{i=1}^{n}  \hat{\theta}_{i,t}( \bmx_{i}(t) )   -  \sum_{i=1}^{n} \hat{\theta}_{i,t}( \bmy^\star_{\mathrm{LS}} )
&\leq  \frac{1}{2 \eta} \left(  \sum_{i=1}^{n} \| \bmx_{i}(t) - \bmy^\star_{\mathrm{LS}} \|^2    -  \sum_{i=1}^{n} \expect  \left[   \| \bmx_{i}(t+1) - \bmy^\star_{\mathrm{LS}} \|^2 \mid \mathcal{F}_t \right]   \right)  \\
&\quad +  \frac{\eta}{2}   \sum_{i=1}^{n}  \expect  \big[ \left\| \mathpzc{g}_{i,t} ( \bmx_{i}(t) ) \right\|^2  \mid \mathcal{F}_t \big] .
\label{lemma-ls-grad-bandit-3}
\end{aligned}
\end{equation}
We provide a lower bound on the left-hand side of (\ref{lemma-ls-grad-bandit-3}) in terms of function $\theta_{i,t}$, by Lemma \ref{lemma-bandit-grad-est}(ii),
\begin{equation}
\begin{aligned}[b]
&\sum_{i=1}^{n}  \hat{\theta}_{i,t}( \bmx_{i}(t) )   -  \sum_{i=1}^{n} \hat{\theta}_{i,t}( \bmy^\star_{\mathrm{LS}} ) \nonumber\\
&\geq  \sum_{i=1}^{n}  \left(    \theta_{i,t}( \bmx_{i}(t) )  - \left\| \bmh_{i}(t) \left(  \bmh_{i}(t)^\tp \bmx_{i}(t) - z_{i}(t)  \right) \right\|   \epsilon  - \alpha_{\bmh}  \epsilon^2  \right)   \\
&\quad -  \sum_{i=1}^{n}  \left(    \theta_{i,t}( \bmy^\star_{\mathrm{LS}} )  + \left\| \bmh_{i}(t) \left(  \bmh_{i}(t)^\tp \bmy^\star_{\mathrm{LS}} - z_{i}(t)  \right) \right\|  \epsilon  + \alpha_{\bmh}  \epsilon^2  \right)  \\
&\geq  \sum_{i=1}^{n} \theta_{i,t}( \bmx_{i}(t) )  -  \sum_{i=1}^{n} \theta_{i,t}( \bmy^\star_{\mathrm{LS}} )  -  2 n \alpha_{\bmh}  \epsilon^2
- \epsilon  \sum_{i=1}^{n}  \left\| \bmh_{i}(t) \left(  \bmh_{i}(t)^\tp \bmy^\star_{\mathrm{LS}} - z_{i}(t)  \right) \right\| \\
&\quad-   \epsilon \sum_{i=1}^{n}  \left\| \bmh_{i}(t) \left(  \bmh_{i}(t)^\tp \bmx_{i}(t) - z_{i}(t)  \right)  \right\|
\label{lemma-ls-grad-bandit-4}
\end{aligned}
\end{equation}
the last two terms can be respectively bounded as follows:
\begin{equation}
\begin{aligned}[b]
\epsilon  \sum_{i=1}^{n}   \left\|   \bmh_{i}(t) \left(  \bmh_{i}(t)^\tp \bmy^\star_{\mathrm{LS}} - z_{i}(t)  \right) \right\|
&\leq  \epsilon  \sum_{i=1}^{n}  \big| \bmh_{i}(t)^\tp \bmy^\star_{\mathrm{LS}} - z_{i}(t)  \big|   \cdot  \big\| \bmh_{i}(t) \big\|
\leq  n \theta^{\star} \sqrt{\alpha_{\bmh}}   \epsilon
\label{lemma-ls-grad-bandit-5}
\end{aligned}
\end{equation}
because of $\left\| \bmh_{i}(t) \right\|^2 \leq \alpha_{\bmh} $ for all $i\in\mathrm{V}$ and $t=1,\ldots,T$ and the assumption that $\big| \bmh_{i}(t)^\tp \bmy^\star_{\mathrm{LS}} - z_{i}(t)  \big|  \leq \theta^{\star}$. On the other hand, we have
\begin{equation}
\begin{aligned}[b]
\epsilon \sum_{i=1}^{n}  \left\| \bmh_{i}(t) \left(  \bmh_{i}(t)^\tp \bmx_{i}(t) - z_{i}(t)  \right)  \right\|
&=  n \epsilon  \cdot  \frac{1}{n} \sum_{i=1}^{n}   \left\| \bmh_{i}(t) \left(  \bmh_{i}(t)^\tp \bmx_{i}(t) - z_{i}(t)  \right)  \right\|  \\
&\leq  \frac{1}{2}  \left( n \epsilon \right)^2  +  \frac{1}{2}  \left( \frac{1}{n} \sum_{i=1}^{n}  \left\| \bmh_{i}(t) \left(  \bmh_{i}(t)^\tp \bmx_{i}(t) - z_{i}(t)  \right)  \right\|  \right)^2  \\
&\leq  \frac{1}{2}   n^2 \epsilon^2  +  \frac{1}{2 n}  \sum_{i=1}^{n}  \left\| \bmh_{i}(t) \left(  \bmh_{i}(t)^\tp \bmx_{i}(t) - z_{i}(t)  \right)  \right\|^2
\label{lemma-ls-grad-bandit-6}
\end{aligned}
\end{equation}
because of the convexity of the norm square function. Combining the inequalities (\ref{lemma-ls-grad-bandit-4}), (\ref{lemma-ls-grad-bandit-5}) and (\ref{lemma-ls-grad-bandit-6}), we get
\begin{equation}
\begin{aligned}[b]
& \sum_{i=1}^{n}  \hat{\theta}_{i,t}( \bmx_{i}(t) )   -  \sum_{i=1}^{n} \hat{\theta}_{i,t}( \bmy^\star_{\mathrm{LS}} ) \geq  \sum_{i=1}^{n} \theta_{i,t}( \bmx_{i}(t) )  \!-\!  \sum_{i=1}^{n} \theta_{i,t}( \bmy^\star_{\mathrm{LS}} ) \\
&\quad  \!-\! \frac{1}{2 n}  \sum_{i=1}^{n}  \left\| \bmh_{i}(t) \left(  \bmh_{i}(t)^\tp \bmx_{i}(t) - z_{i}(t)  \right)  \right\|^2
\!-\!   n \theta^{\star} \sqrt{\alpha_{\bmh}}   \epsilon  \!-\! \left(  2 n \alpha_{\bmh} + \frac{1}{2} n^2  \right)  \epsilon^2.
\label{lemma-ls-grad-bandit-7}
\end{aligned}
\end{equation}
Substituting the preceding inequality into (\ref{lemma-ls-grad-bandit-3}) and using Lemma \ref{lemma-bandit-grad-est}(i), \ie,
\begin{equation}
\begin{aligned}[b]
&\sum_{i=1}^{n}  \expect  \big[ \left\| \mathpzc{g}_{i,t} ( \bmx_{i}(t) ) \right\|^2  \mid \mathcal{F}_t \big]\\
&\leq  \sum_{i=1}^{n} m^2  \left(   \left\| \bmh_{i}(t) \left(  \bmh_{i}(t)^\tp \bmx_{i}(t) - z_{i}(t)  \right)  \right\| +  \| \bmh_{i}(t) \|^2 \epsilon  \right)^2   \\
&\leq  2 m^2   \sum_{i=1}^{n} \left\| \bmh_{i}(t) \left(  \bmh_{i}(t)^\tp \bmx_{i}(t) - z_{i}(t)  \right)  \right\|^2 +  2 n  m^2 \alpha_{\bmh}^2 \epsilon^2
\label{lemma-ls-grad-bandit-8}
\end{aligned}
\end{equation}
we find the following estimate by taking the total expectation,
\begin{equation}
\begin{aligned}[b]
&\sum_{i=1}^{n}  \expect [  \theta_{i,t}( \bmx_{i}(t) ) ]  -  \sum_{i=1}^{n} \expect [ \theta_{i,t}( \bmy^\star_{\mathrm{LS}} )  ]\\
&\leq  \frac{1}{2 \eta} \left(  \sum_{i=1}^{n} \expect  \left[ \| \bmx_{i}(t) - \bmy^\star_{\mathrm{LS}} \|^2 \right]   -  \sum_{i=1}^{n} \expect  \left[ \| \bmx_{i}(t+1) - \bmy^\star_{\mathrm{LS}} \|^2 \right]   \right)  \\
&\quad  + \left( m^2 \eta + \frac{1}{2 n} \right)  \sum_{i=1}^{n} \expect  \left[ \left\| \bmh_{i}(t) \left(  \bmh_{i}(t)^\tp \bmx_{i}(t) - z_{i}(t)  \right)  \right\|^2 \right] \\
&\quad  + n \theta^{\star} \sqrt{\alpha_{\bmh}}   \epsilon +  \left( n  m^2 \alpha_{\bmh}^2 \eta + 2 n \alpha_{\bmh} + \frac{1}{2} n^2  \right)  \epsilon^2.
\label{lemma-ls-grad-bandit-9}
\end{aligned}
\end{equation}
Using the self-boundedness property of function $\theta_{i,t}(\cdot)$ (cf. (\ref{lemma-ls-grad-5b-self-bound})) and following similar lines as that of Lemma \ref{lemma-ls-grad}, we have
\begin{equation}
\begin{aligned}[b]
& \left( \frac{1}{2 \alpha_{\bmh}} - \frac{1}{2 n} -  m^2 \eta  \right) \sum_{t=1}^{T} \sum_{i=1}^{n} \expect  \left[ \left\| \bmh_{i}(t) \left(  \bmh_{i}(t)^\tp \bmx_{i}(t) - z_{i}(t)  \right)  \right\|^2 \right]   \\
&\leq  \frac{1}{2 \eta}  \sum_{i=1}^{n} \expect  \left[ \| \bmx_{i}(1) - \bmy^\star_{\mathrm{LS}} \|^2 \right]
+ \frac{1}{2}  n (\theta^{\star})^2 \cdot T  +  n  m^2 \alpha_{\bmh}^2  \cdot  \eta \epsilon^2 T
+  n \theta^{\star} \sqrt{\alpha_{\bmh}}  \cdot \epsilon T \\
& +  \Big(  2 n \alpha_{\bmh} + \frac{1}{2} n^2  \Big)  \cdot   \epsilon^2 T
\label{lemma-ls-grad-bandit-10}
\end{aligned}
\end{equation}
dividing both side by the positive term $\left( \frac{1}{2 \alpha_{\bmh}} - \frac{1}{2 n} -  m^2 \eta  \right)$ (due to $\kappa >  \frac{2 n m^2 \alpha_{\bmh}}{n - \alpha_{\bmh}}$ and $\alpha_{\bmh} < n$), yields
\begin{equation}
\begin{aligned}[b]
&\sum_{t=1}^{T} \sum_{i=1}^{n} \expect  \left[ \left\| \bmh_{i}(t) \left(  \bmh_{i}(t)^\tp \bmx_{i}(t) - z_{i}(t)  \right)  \right\|^2 \right]  \\
& \leq  \frac{\hat{\kappa}}{2 \eta}  \sum_{i=1}^{n} \expect  \left[ \| \bmx_{i}(1) - \bmy^\star_{\mathrm{LS}} \|^2 \right]
\!+\! \frac{1}{2}  n (\theta^{\star})^2 \hat{\kappa} \cdot T
\!+\!   n  m^2 \alpha_{\bmh}^2 \hat{\kappa} \cdot  \eta \epsilon^2  T\\
& \quad \!+\!  n \theta^{\star} \sqrt{\alpha_{\bmh}} \hat{\kappa} \cdot \epsilon T
\!+\!   \left(\!  2 n \alpha_{\bmh} + \frac{1}{2} n^2  \!\right) \hat{\kappa} \cdot   \epsilon^2 T  \\
& \leq \left(   \frac{1}{\kappa} n  m^2 \alpha_{\bmh}^2  +  2 n \alpha_{\bmh} + \frac{1}{2} n^2  \right) \hat{\kappa}
+ \left(  \frac{1}{2} \kappa  \sum_{i=1}^{n} \expect  \left[ \| \bmx_{i}(1) - \bmy^\star_{\mathrm{LS}} \|^2 \right]  +  n \theta^{\star} \sqrt{\alpha_{\bmh}}    +  \frac{1}{2}  n (\theta^{\star})^2   \right)  \hat{\kappa} \cdot T  \\
& =  B_1  +  B_2 \cdot T
\label{lemma-ls-grad-bandit-11}
\end{aligned}
\end{equation}
where in the last inequality we used $\eta = \frac{1}{\kappa T^\beta}$ and $\epsilon  = \frac{1}{\sqrt{T}}$ for all $t=1,\ldots,T$.

(ii) The analysis of the accumulated disagreement is similar to that of Lemma \ref{lemma-ls-grad}(ii), and it is easy to get
\begin{equation}
\begin{aligned}[b]
\sum_{t=1}^{T} \sum_{i=1}^{n}  \big\| \bmx_{i}(t) - \bmx_{\mathrm{avg}}(t)  \big\|
&\leq \sum_{i=1}^{n}  \big\| \bmx_{i}(1) - \bmx_{\mathrm{avg}}(1)  \big\|
+  n^{3/2}   \frac{\sigma_2(W_{\mathrm{G}})}{1 - \sigma_2(W_{\mathrm{G}})}  \max_{i\in \mathrm{V}}  \{   \| \bmx_{i}(1) \|  \}  \\
&\quad + n^{3/2}    \frac{ \sigma_2(W_{\mathrm{G}})}{1 - \sigma_2(W_{\mathrm{G}})}  \eta  \sum_{t=1}^{T-1} \sum_{i=1}^{n}  \left\| \mathpzc{g}_{i,t} ( \bmx_{i}(t) ) \right\|.
\label{lemma-ls-disagree-12}
\end{aligned}
\end{equation}
We are left to bound the last term on the right-hand side of (\ref{lemma-ls-disagree-12}). Summing the inequalities in (\ref{lemma-ls-grad-bandit-8}) over $t=1$ to $t=T$, taking the total expectation on both sides, and then combining with inequality (\ref{lemma-ls-grad-bandit-11}), we find that
\begin{equation}
\begin{aligned}[b]
\sum_{t=1}^{T} \sum_{i=1}^{n}  \expect  \big[ \left\| \mathpzc{g}_{i,t} ( \bmx_{i}(t) ) \right\|^2  \big]
&\leq  2 m^2  \sum_{t=1}^{T} \sum_{i=1}^{n} \expect  \left[ \left\| \bmh_{i}(t) \left(  \bmh_{i}(t)^\tp \bmx_{i}(t) - z_{i}(t)  \right)  \right\|^2 \right] +  2 n  m^2 \alpha_{\bmh}^2 \cdot \epsilon^2  T \\
&\leq  2 m^2  \left( B_1  +  n \alpha_{\bmh}^2 \right) + 2 m^2  B_2  \cdot T  .
\label{lemma-ls-grad-bandit-12}
\end{aligned}
\end{equation}
Using Jensen's inequality, we further have
\begin{equation}
\begin{aligned}[b]
&\sum_{t=1}^{T} \sum_{i=1}^{n}  \expect  \big[ \left\| \mathpzc{g}_{i,t} ( \bmx_{i}(t) ) \right\|  \big]\\
&\leq  \sqrt{   n T    \sum_{t=1}^{T} \sum_{i=1}^{n}  \expect  \big[ \left\| \mathpzc{g}_{i,t} ( \bmx_{i}(t) ) \right\|^2  \big] }\\
&\leq m \sqrt{ 2 n   \left( B_1  +  n \alpha_{\bmh}^2 \right) } \cdot \sqrt{T}  +  m \sqrt{ 2 n  B_2 } \cdot T.
\end{aligned}
\end{equation}
taking the total expectation on both sides of (\ref{lemma-ls-disagree-12}) and then using the preceding inequality and $\eta = \frac{1}{\kappa  T^{\beta}}$, we derive the desired bound.
\hfill$\square$

\subsection{Proof of the theorem}
Combining the inequalities (\ref{lemma-ls-grad-bandit-3}), (\ref{lemma-ls-grad-bandit-4}), (\ref{lemma-ls-grad-bandit-5}) and (\ref{lemma-ls-grad-bandit-8}), and taking the total expectation, we obtain
\begin{equation}
\begin{aligned}[b]
&\sum_{i=1}^{n}  \expect [  \theta_{i,t}( \bmx_{i}(t) ) ]  -  \sum_{i=1}^{n} \expect [ \theta_{i,t}( \bmy^\star_{\mathrm{LS}} )  ] \\
&\leq  \frac{1}{2 \eta} \left(  \sum_{i=1}^{n} \expect  \left[ \| \bmx_{i}(t) - \bmy^\star_{\mathrm{LS}} \|^2 \right]   -  \sum_{i=1}^{n} \expect  \left[ \| \bmx_{i}(t+1) - \bmy^\star_{\mathrm{LS}} \|^2 \right]   \right)\\
&\quad + m^2 \eta   \sum_{i=1}^{n} \expect  \left[ \left\| \bmh_{i}(t) \left(  \bmh_{i}(t)^\tp \bmx_{i}(t) - z_{i}(t)  \right) \right\|^2 \right] \\
&\quad + \epsilon \sum_{i=1}^{n} \expect  \left[ \left\| \bmh_{i}(t) \left(  \bmh_{i}(t)^\tp \bmx_{i}(t) - z_{i}(t)  \right) \right\|  \right]
+ n \theta^{\star} \sqrt{\alpha_{\bmh}}   \epsilon   +  2 n \alpha_{\bmh} \epsilon^2   +   n  m^2 \alpha_{\bmh}^2 \eta  \epsilon^2
\label{theorem-ls-bandit-1}
\end{aligned}
\end{equation}
in fact, this inequality is just (\ref{lemma-ls-grad-bandit-9}) by substituting (\ref{lemma-ls-grad-bandit-6}) into the preceding inequality. Then, summing the inequalities in (\ref{theorem-ls-bandit-1}) over $t=1$ to $t=T$, we have
\begin{equation}
\begin{aligned}[b]
&  \sum_{t=1}^{T} \sum_{i=1}^{n}  \expect [  \theta_{i,t}( \bmx_{i}(t) ) ]  -  \sum_{i=1}^{n} \expect [ \theta_{i,t}( \bmy^\star_{\mathrm{LS}} )  ]   \\
&\leq  \frac{1}{2 \eta} \sum_{t=1}^{T} \left(  \sum_{i=1}^{n} \expect  \left[ \| \bmx_{i}(t) - \bmy^\star_{\mathrm{LS}} \|^2 \right]   -  \sum_{i=1}^{n} \expect  \left[ \| \bmx_{i}(t+1) - \bmy^\star_{\mathrm{LS}} \|^2 \right]   \right) \\
&\quad + m^2 \eta  \sum_{t=1}^{T} \sum_{i=1}^{n} \expect  \left[ \left\| \bmh_{i}(t) \left(  \bmh_{i}(t)^\tp \bmx_{i}(t) - z_{i}(t)  \right) \right\|^2  \right] \\
&\quad + \sum_{t=1}^{T} \epsilon \sum_{i=1}^{n} \expect  \left[ \left\| \bmh_{i}(t) \left(  \bmh_{i}(t)^\tp \bmx_{i}(t) - z_{i}(t)  \right) \right\|  \right]
+ n \theta^{\star} \sqrt{\alpha_{\bmh}} \cdot  \epsilon T   +  2 n \alpha_{\bmh} \cdot \epsilon^2 T  + n  m^2 \alpha_{\bmh}^2   \cdot \eta \epsilon^2 T
\label{theorem-ls-bandit-2}
\end{aligned}
\end{equation}
combining this with (\ref{lemma-ls-grad-bandit-11}), using Jensen's inequality, and the fact that $\epsilon  =  \frac{1}{\sqrt{T}}$ for all $t=1,\ldots,T$, yields
\begin{equation}
\begin{aligned}[b]
\sum_{t=1}^{T} \sum_{i=1}^{n}  \expect [  \theta_{i,t}( \bmx_{i}(t) ) ]  -  \sum_{i=1}^{n} \expect [ \theta_{i,t}( \bmy^\star_{\mathrm{LS}} )  ]  &\leq  B_6 +  B_7 \cdot  T^{1-\beta}  + B_8 \cdot \sqrt{T} + B_9  \cdot  T^\beta
\label{theorem-ls-bandit-2a}
\end{aligned}
\end{equation}
where $B_6  = \frac{m^2}{\kappa} B_1 +   \sqrt{n  B_1}   +  \frac{n  m^2 \alpha_{\bmh}^2}{\kappa}  +  2 n \alpha_{\bmh}$, $B_7   =  \frac{m^2}{\kappa} B_2$, $B_8  = \sqrt{n  B_2}  + n \theta^{\star} \sqrt{\alpha_{\bmh}}$, and $B_9  = \frac{\kappa}{2} \sum_{i=1}^{n} \expect  \left[ \| \bmx_{i}(1) - \bmy^\star_{\mathrm{LS}} \|^2 \right]$.
From (\ref{theorem-ls-regret-4}) we know that
\begin{equation}
\begin{aligned}[b]
\sum_{t=1}^{T} \sum_{j=1}^{n} \theta_{j,t} \left( \bmx_{j}(t) \right)
&\geq   \frac{1}{1 + T^{-\gamma}} \sum_{t=1}^{T} \sum_{j=1}^{n}  \theta_{j,t} \left( \bmx_{i}(t) \right) -   T^{\gamma } \cdot  \mathcal{R}(T)
\label{theorem-ls-bandit-3}
\end{aligned}
\end{equation}
where $\gamma > 0$ and as in (\ref{theorem-ls-regret-5}), $\mathcal{R}(T)$ can be bounded as follows:
\begin{equation}
\begin{aligned}[b]
\mathcal{R}(T)
&\leq  2 \alpha_{\bmh} n \left(  \sum_{t=1}^{T} \sum_{i=1}^{n}  \big\| \bmx_{i}(t) - \bmx_{\mathrm{avg}}(t) \big\|  \right)^2\\
&\leq  6 \alpha_{\bmh} n B_3^2 + 6 \alpha_{\bmh} n B_4^2 \cdot T^{1- 2 \beta} +  6 \alpha_{\bmh} n B_5^2 \cdot T^{2(1-\beta)}  \\
&\leq  B_{10} \cdot T^{2(1-\beta)}
\label{theorem-ls-bandit-4}
\end{aligned}
\end{equation}
where $B_{10} = 6 \alpha_{\bmh} n \left(  B_3^2 +  B_4^2  +   B_5^2 \right) $ and the second inequality is based on Lemma \ref{lemma-ls-grad-bandit}(ii). Combining the results in (\ref{theorem-ls-bandit-2a}), (\ref{theorem-ls-bandit-3}) and (\ref{theorem-ls-bandit-4}), we find that
\begin{equation}
\begin{aligned}[b]
& \sum_{t=1}^{T} \sum_{j=1}^{n}  \expect [  \theta_{j,t}( \bmx_{i}(t) ) ]  -  \sum_{i=1}^{n} \expect [ \theta_{j,t}( \bmy^\star_{\mathrm{LS}} )  ]   \\
&\leq \frac{1 + T^\gamma}{T^\gamma}  \left(   B_6 +  B_7 \cdot  T^{1-\beta}  + B_8 \cdot \sqrt{T} + B_9  \cdot  T^\beta + B_{10} \cdot T^{2(1-\beta) + \gamma} \right) + T^{-\gamma} \sum_{j=1}^{n} \expect [ \theta_{j,t}( \bmy^\star_{\mathrm{LS}} )  ]    \\
&\leq  \left(1 +  T^{-\gamma}\right) \left(   B_6 +  B_7 \cdot  T^{1-\beta}  + B_8 \cdot \sqrt{T} + B_9  \cdot  T^\beta + B_{10} \cdot T^{2(1-\beta) + \gamma} \right) + \frac{1}{2} n (\theta^{\star})^2 \cdot  T^{1-\gamma}  \\
&\leq  2 \left(   B_6 +  B_7 \cdot  T^{1-\beta}  + B_8 \cdot \sqrt{T} + B_9  \cdot  T^\beta + B_{10} \cdot T^{2(1-\beta) + \gamma} \right) + \frac{1}{2} n (\theta^{\star})^2 \cdot  T^{1-\gamma}
\label{theorem-ls-bandit-5}
\end{aligned}
\end{equation}
due to the assumption of $\left|  \bmh_{j}(t)^\tp \bmy_{\mathrm{LS}}^{\star}  - z_{j}(t) \right| \leq \theta^{\star}$. It can be derived from the term $T^{2(1-\beta) + \gamma}$ that $\beta >  \frac{1}{2}$ must hold. Based on this fact, we can see that there exist three dominant terms on the right-hand side of (\ref{theorem-ls-bandit-5}): $T^\beta $, $T^{2(1-\beta) + \gamma}$ and $T^{1- \gamma}$. Suppose that $\beta = \frac{1}{2} + \pi$ with $\pi \in(0,\frac{1}{2}) $, then the last two dominant terms become $T^{ 1 - 2 \pi + \gamma}$ and $T^{1- \gamma}$, which yields the optimal choice of $\gamma = \pi$ by setting $1 - 2 \pi + \gamma = 1- \gamma$. We now are left to balance between $T^{\frac{1}{2} + \pi}$ and $T^{1- \pi}$, by setting $\frac{1}{2} + \pi = 1 - \pi$ we get the optimal choice of $\pi = \frac{1}{4}$. Hence, we have the final regret bound:
\begin{equation*}
\begin{aligned}
\reg_{\mathrm{LS}}(i,T)  &\leq&  B \cdot T^{3/4},\qquad\qquad    T\geq 2
\end{aligned}
\end{equation*}
where
\begin{equation*}
\begin{aligned}
B   &=  \frac{1}{2} n (\theta^{\star})^2  +  \kappa \sum_{i=1}^{n} \expect  \left[ \| \bmx_{i}(1) - \bmy^\star_{\mathrm{LS}} \|^2 \right] + 12 \alpha_{\bmh} n \left(  B_3^2 +  B_4^2  +   B_5^2 \right)   \\
&\quad +  2 \left( \sqrt{n  B_1}   + \sqrt{n  B_2} + \frac{m^2}{\kappa} \left( B_1 +  B_2 + n \alpha_{\bmh}^2\right) +      n \theta^{\star} \sqrt{\alpha_{\bmh}} + 2 n \alpha_{\bmh}   \right).
\end{aligned}
\end{equation*}
This completes the proof of the desired theorem.
\hfill$\square$





\newpage

\section{Proof of Theorem \ref{theorem-ls-bandit-adversary}}

We first provide a bound on the gradient of loss function $\theta_{i,t}(\bmy)$ for any $\bmy\in\mathcal{K}$,
\begin{equation}
\begin{aligned}[b]
& \left\|  \bmh_{i}(t) \left(  \bmh_{i}(t)^\tp \bmy - z_{i}(t)  \right) \right\|
\leq \| \bmh_{i}(t) \| \left( \| \bmh_{i}(t) \| \cdot \| \bmy \|  + z_{i}(t) \right)
\leq \alpha_{\bmh} R + \sqrt{\alpha_{\bmh}} \alpha_z
= L.
\label{theorem-ls-bandit-adversary-0}
\end{aligned}
\end{equation}
It follows  that for any $\bmy \in (1-\xi)\mathcal{K}$,
\begin{equation}
\begin{aligned}[b]
\sum_{i=1}^{n} \| \bmx_{i}(t+1) - \bmy \|^2
&= \sum_{i=1}^{n} \bigg\| \mathpzc{P}_{(1-\xi) \mathcal{K}} \Big( \sum_{j=1}^{n}  [W_{\mathrm{G}}]_{ij} \cdot \mathpzc{l}_{j}(t) \Big) - \bmy \bigg\|^2\\
&\leq \sum_{i=1}^{n} \bigg\| \sum_{j=1}^{n} [W_{\mathrm{G}}]_{ij} \cdot \mathpzc{l}_{j}(t) - \bmy \bigg\|^2
\label{theorem-ls-bandit-adversary-1}
\end{aligned}
\end{equation}
where we used the nonexpansiveness of the Euclidean projection, \ie, $\|  \mathpzc{P}_{(1-\xi) \mathcal{K}} (\bmx) -  \mathpzc{P}_{(1-\xi) \mathcal{K}} (\bmy) \|  \leq  \|  \bmx  -  \bmy \|$ for any $\bmx,\bmy\in\reals^m$. Then, following an argument similar to that of (\ref{lemma-ls-grad-bandit-1})--(\ref{lemma-ls-grad-bandit-3}), we have that, for any $\bmy \in (1-\xi)\mathcal{K}$,
\begin{equation}
\begin{aligned}[b]
&\sum_{i=1}^{n}  \hat{\theta}_{i,t}( \bmx_{i}(t) )   -  \sum_{i=1}^{n} \hat{\theta}_{i,t}( \bmy ) \\
&\leq  \frac{1}{2 \eta} \left(  \sum_{i=1}^{n} \| \bmx_{i}(t) - \bmy \|^2    -  \sum_{i=1}^{n} \expect  \left[   \| \bmx_{i}(t+1) - \bmy \|^2 \mid \mathcal{F}_t \right]   \right)
+  \frac{\eta}{2}   \sum_{i=1}^{n}  \expect  \big[ \left\| \mathpzc{g}_{i,t} ( \bmx_{i}(t) ) \right\|^2  \mid \mathcal{F}_t \big] .
\label{theorem-ls-bandit-adversary-2}
\end{aligned}
\end{equation}
Now we have a new bound on the gradient estimator, by using Lemma \ref{lemma-bandit-grad-est} and (\ref{theorem-ls-bandit-adversary-0}),
\begin{equation}
\begin{aligned}[b]
\expect  \big[ \left\| \mathpzc{g}_{i,t} ( \bmx_{i}(t) ) \right\|^2  \mid \mathcal{F}_t \big]
&\leq m^2 \left(  \left\| \bmh_{i}(t) \left(  \bmh_{i}(t)^\tp \bmy - z_{i}(t)  \right) \right\| + m \| \bmh_{i}(t) \|^2 \epsilon \right)^2\\ &
\leq 2 m^2 L^2 + 2 m^2 \alpha_{\bmh}^2 \epsilon^2.
\label{theorem-ls-bandit-adversary-3}
\end{aligned}
\end{equation}
On the other hand, the left-hand side of (\ref{theorem-ls-bandit-adversary-2}) can be lower bounded as follows:
\begin{equation*}
\begin{aligned}
\sum_{i=1}^{n}  \hat{\theta}_{i,t}( \bmx_{i}(t) )   -  \sum_{i=1}^{n} \hat{\theta}_{i,t}( \bmy )
&\geq  \sum_{i=1}^{n}  \left(    \theta_{i,t}( \bmx_{i}(t) )  - \left\| \bmh_{i}(t) \left(  \bmh_{i}(t)^\tp \bmx_{i}(t) - z_{i}(t)  \right) \right\|   \epsilon  - \alpha_{\bmh}  \epsilon^2  \right)   \\
&\quad -  \sum_{i=1}^{n}  \left(    \theta_{i,t}( \bmy )  + \left\| \bmh_{i}(t) \left(  \bmh_{i}(t)^\tp \bmy - z_{i}(t)  \right) \right\|  \epsilon  + \alpha_{\bmh}  \epsilon^2  \right)  \\
&\geq  \sum_{i=1}^{n} \theta_{i,t}( \bmx_{i}(t) )  -  \sum_{i=1}^{n} \theta_{i,t}( \bmy )  - 2 n L \epsilon - 2 n \alpha_{\bmh}  \epsilon^2
\end{aligned}
\end{equation*}
this, combined with (\ref{theorem-ls-bandit-adversary-2}) and (\ref{theorem-ls-bandit-adversary-3}), gives that for any $\bmy\in\mathcal{K}$,
\begin{equation}
\begin{aligned}[b]
&\sum_{t=1}^{T} \sum_{j=1}^{n} \expect\left[ \theta_{j,t}( \bmx_{i}(t) ) \right]  -  \sum_{t=1}^{T} \sum_{j=1}^{n} \expect\left[ \theta_{j,t}\left( (1-\xi)\bmy \right) \right] \\
& \leq  \frac{1}{2 \eta}  \sum_{j=1}^{n} \expect\left[ \| \bmx_{j}(1) - \bmy \|^2  \right]
+ 2 n L \cdot \epsilon T + 2 n \alpha_{\bmh}  \cdot \epsilon^2 T
+  n m^2 L^2 \cdot \eta T + n m^2 \alpha_{\bmh}^2 \cdot \eta \epsilon^2 T
\label{theorem-ls-bandit-adversary-4}
\end{aligned}
\end{equation}
where we denote $\eta = \frac{1}{\sqrt{T}}$ and used Lemma \ref{lemma-bandit-grad-est}. Using Assumption \ref{assumption-adversary}, we have that, for any $\bmy\in\mathcal{K}$,
\begin{equation}
\begin{aligned}[b]
\theta_{j,t}\left( (1-\xi)\bmy \right) - \theta_{j,t}\left( \bmy \right)
&\leq \nabla \theta_{j,t}\left( (1-\xi)\bmy \right)^\tp \left( (1-\xi)\bmy - \bmy  \right)\\&
\leq \bmh_{j}(t) \left(  \bmh_{j}(t)^\tp (1-\xi)\bmy - z_{j}(t)  \right) ^\tp \left( \xi\bmy \right) \\
&\leq \left\| \bmh_{j}(t) \left(  \bmh_{j}(t)^\tp (1-\xi)\bmy - z_{j}(t)  \right) \right\| \cdot \| \xi\bmy \|
\leq  L R \xi
\label{theorem-ls-bandit-adversary-5}
\end{aligned}
\end{equation}
and
\begin{equation}
\begin{aligned}
\theta_{j,t}\left( \bmx_{j}(t) \right) - \theta_{j,t}\left( \bmx_{i}(t) \right)
&\geq L \| \bmx_{j}(t) - \bmx_{i}(t) \|.
\label{theorem-ls-bandit-adversary-6}
\end{aligned}
\end{equation}
On the other hand, following the disagreement analysis of Lemma \ref{lemma-ls-grad-bandit} and using the new bound in (\ref{theorem-ls-bandit-adversary-0}), we have
\begin{equation}
\begin{aligned}[b]
&\sum_{t=1}^{T} \sum_{i=1}^{n} \expect \big[  \big\| \bmx_{i}(t) - \bmx_{\mathrm{avg}}(t)  \big\|  \big]\\
&\leq B_3 + 2 n^{3/2} m^2 \frac{\sigma_2(W_{\mathrm{G}})}{1-\sigma_2(W_{\mathrm{G}})} L^2 \cdot \eta T
+ 2 n^{3/2} m^2 \frac{\sigma_2(W_{\mathrm{G}})}{1-\sigma_2(W_{\mathrm{G}})} \alpha_{\bmh}^2 \cdot \eta \epsilon^2 T.
\label{theorem-ls-bandit-adversary-7}
\end{aligned}
\end{equation}
Combining the results in (\ref{theorem-ls-bandit-adversary-4})--(\ref{theorem-ls-bandit-adversary-7}), we find that
\begin{equation*}
\begin{aligned}
& \sum_{t=1}^{T} \sum_{j=1}^{n} \expect\left[ \theta_{j,t}( \bmx_{i}(t) ) \right]  -  \sum_{t=1}^{T} \sum_{j=1}^{n} \expect\left[ \theta_{j,t}\left( \bmy \right) \right]  \\
&\leq 2 L B_3 +  \frac{1}{2 \eta}  \sum_{j=1}^{n} \expect\left[ \| \bmx_{j}(1) - \bmy \|^2  \right]
+ n L R \cdot \xi T + 2 n L \cdot \epsilon T + 2 n \alpha_{\bmh}  \cdot \epsilon^2 T  \\
&\quad +  \left( n m^2 L^2 + 4 n^{3/2} m^2 \frac{\sigma_2(W_{\mathrm{G}})}{1-\sigma_2(W_{\mathrm{G}})} L^3 \right)\cdot \eta T\\&
\quad \quad + \left( n m^2 \alpha_{\bmh}^2 + 4 n^{3/2} m^2 \frac{\sigma_2(W_{\mathrm{G}})}{1-\sigma_2(W_{\mathrm{G}})} L \alpha_{\bmh}^2 \right) \cdot \eta \epsilon^2 T
\end{aligned}
\end{equation*}
substituting $\eta = \frac{1}{\sqrt{T}}$, $\epsilon = \frac{1}{\sqrt{T}}$ and $\xi = \frac{\epsilon}{r}$ into the preceding inequality, we obtain
\begin{equation}
\begin{aligned}
\sum_{t=1}^{T} \sum_{j=1}^{n} \expect\left[ \theta_{j,t}( \bmx_{i}(t) ) \right]  -  \sum_{t=1}^{T} \sum_{j=1}^{n} \expect\left[ \theta_{j,t}\left( \bmy \right) \right]
&\leq   \frac{1}{2}  \sum_{j=1}^{n} \expect\left[ \| \bmx_{j}(1) - \bmy \|^2  \right] \cdot \sqrt{T} + \mathcal{O}(\sqrt{T})
\label{theorem-ls-bandit-adversary-8}
\end{aligned}
\end{equation}
then, maximizing both sides of (\ref{theorem-ls-bandit-adversary-8}) with respect to $\bmy\in\mathcal{K}$, we finally have
\begin{equation*}
\begin{aligned}
&\sum_{t=1}^{T} \sum_{j=1}^{n} \expect\left[ \theta_{j,t}( \bmx_{i}(t) ) \right]  -  \min_{\bmy\in\mathcal{K}} \sum_{t=1}^{T} \sum_{j=1}^{n} \expect\left[ \theta_{j,t}\left( \bmy \right) \right]\\
&\leq   \frac{1}{2}  \max_{\bmy\in\mathcal{K}} \left\{ \sum_{j=1}^{n} \expect\left[ \| \bmx_{j}(1) - \bmy \|^2  \right] \right\} \cdot \sqrt{T} + \mathcal{O}(\sqrt{T})  \\
&\leq \frac{1}{2} n R^2 \cdot \sqrt{T} + \mathcal{O}(\sqrt{T})
= \mathcal{O}(\sqrt{T}).
\end{aligned}
\end{equation*}
The proof is complete.
\hfill$\square$





\newpage

\section{Proofs of Theorems \ref{theorem-dolr-constraints} and \ref{theorem-dolr-constraints-bandit}}

\subsection{Key Lemma}
We have the following lemma that characterizes the disagreement among all the nodes.
\begin{lemma}\label{lemma-dolr-constraints-disagree}
Let Assumption \ref{assumption-constraints} hold. Assume that $\mathrm{rank(\bmH(t))} = m$ for at least one $t\in\{ 1,\ldots,T\}$ and $(\bmh_{i}(t),z_i(t) ) \in \mathcal{K}_{\mathbf{Ad}}$ in (\ref{space-k-ad}).
\BIT
\item[(i)] (Full information feedback) Then along Algorithm \ref{alg-dolr-cons} there holds
\begin{equation*}
\begin{aligned}[b]
\sum_{t=1}^{T} \sum_{i=1}^{n} \big\| \bmx_{i}(t) - \bmx_{\mathrm{avg}}(t)  \big\|
\leq C_3 + n L E  \cdot \eta T + K_{\mathrm{I}} E \sum_{t=1}^{T-1} \sum_{i=1}^{n} \sum_{q=1}^{s} \eta \left[ \bmmu_i(t) \right]_q
\end{aligned}
\end{equation*}
where $E = \frac{3 \sqrt{n}}{1-\sigma_2(W_{\mathrm{G}})} +4 $ and $C_3$ is given in Lemma \ref{lemma-ls-grad}.
\item[(ii)] (Bandit feedback) Then along Algorithm \ref{alg-dolr-cons-bandit} there holds
\begin{equation*}
\begin{aligned}[b]
\sum_{t=1}^{T} \sum_{i=1}^{n} \expect \big[  \big\| \bmx_{i}(t) - \bmx_{\mathrm{avg}}(t)  \big\|  \big]
\leq B_3 + m n L E  \cdot \eta T + K_{\mathrm{I}} E  \sum_{t=1}^{T-1} \sum_{i=1}^{n} \sum_{q=1}^{s} \eta \left[ \bmmu_i(t) \right]_q
\end{aligned}
\end{equation*}
where $B_3$ is given in Lemma \ref{lemma-ls-grad-bandit}.
\EIT
\end{lemma}
\noindent{\em Proof.}
First, we have the following bound on the gradient of $\theta_{i,t}(\bmy)$ for all $\bmy\in \mathcal{K}$, according to Assumption \ref{assumption-constraints} and the fact that $(\bmh_{i}(t),z_i(t) ) \in \mathcal{K}_{\mathbf{Ad}}$:
\begin{equation}
\begin{aligned}[b]
& \left\| \nabla \theta_{i,t}(\bmy)  \right\| = \left\|  \bmh_{i}(t) \left(  \bmh_{i}(t)^\tp \bmy - z_{i}(t)  \right) \right\|
\leq \alpha_{\bmh} R + \sqrt{\alpha_{\bmh}} \alpha_z = L.
\label{bound-on-gradient-constraints}
\end{aligned}
\end{equation}
(i)
The general evolution of the average state can be derived as follows:
\begin{equation*}
\begin{aligned}[b]
\bmx_{\mathrm{avg}}(t+1)
&= \frac{1}{n} \sum_{i=1}^{n}  \mathpzc{P}_{\mathbb{B}_{R}^{m}}\left( \sum_{j=1}^{n} [W_{\mathrm{G}}]_{ij} \mathpzc{l}_{j}(t) \right)
= \frac{1}{n} \sum_{i=1}^{n}  \sum_{j=1}^{n} [W_{\mathrm{G}}]_{ij} \mathpzc{l}_{j}(t)  + \frac{1}{n} \sum_{i=1}^{n} \mathbf{p}_i(t)
\label{lemma-dolr-constraints-disagree-1}
\end{aligned}
\end{equation*}
where $\mathbf{p}_i(t) = \mathpzc{P}_{\mathbb{B}_{R}^{m}}\left( \sum_{j=1}^{n} [W_{\mathrm{G}}]_{ij} \mathpzc{l}_{j}(t) \right) - \sum_{j=1}^{n} [W_{\mathrm{G}}]_{ij} \mathpzc{l}_{j}(t) $. Hence, we can derive the following general expression,
\begin{equation}
\begin{aligned}[b]
\bmx_{\mathrm{avg}}(t+1) &= \bmx_{\mathrm{avg}}(1)  - \eta \sum_{\ell=1}^{t}  \frac{1}{n} \sum_{i=1}^{n} \nabla_{\bmy} \mathscr{L}_{i,t}(\bmx_{i}(\ell),\bmmu_i(\ell)) + \sum_{\ell=1}^{t}  \frac{1}{n} \sum_{i=1}^{n} \mathbf{p}_i(\ell)
\label{lemma-dolr-constraints-disagree-2}
\end{aligned}
\end{equation}
and similarly,
\begin{equation}
\begin{aligned}[b]
\bmx_{i}(t+1)  &= \sum_{j=1}^{n} [W_{\mathrm{G}}^t]_{ij} \bmx_{j}(1)  - \eta \sum_{\ell = 1}^{t}  \sum_{j=1}^{n} [W_{\mathrm{G}}^{t+1-\ell}]_{ij} \nabla_{\bmy} \mathscr{L}_{j,t}(\bmx_{j}(\ell),\bmmu_j(\ell)) \\
&\quad + \sum_{\ell = 1}^{t-1}  \sum_{j=1}^{n} [W_{\mathrm{G}}^{t-\ell}]_{ij} \mathbf{p}_j(\ell) + \mathbf{p}_i(t) .
\label{lemma-dolr-constraints-disagree-3}
\end{aligned}
\end{equation}
On the other hand, we have the following bound on $\mathbf{p}_i(t)$,
\begin{equation}
\begin{aligned}[b]
\sum_{i=1}^{n} \left\| \mathbf{p}_i(t) \right\| & = \left\| \mathpzc{P}_{\mathbb{B}_{R}^{m}}\left( \sum_{j=1}^{n} [W_{\mathrm{G}}]_{ij} \mathpzc{l}_{j}(t) \right) - \sum_{j=1}^{n} [W_{\mathrm{G}}]_{ij} \left( \bmx_{j}(t) - \eta \nabla_{\bmy} \mathscr{L}_{j,t}(\bmx_{j}(t),\bmmu_j(t)) \right) \right\| \\
&\leq 2  \eta  \sum_{i=1}^{n} \sum_{j=1}^{n} [W_{\mathrm{G}}]_{ij} \left\| \nabla_{\bmy} \mathscr{L}_{j,t}(\bmx_{j}(t),\bmmu_j(t)) \right\| =  2  \eta \sum_{i=1}^{n} \left\| \nabla_{\bmy} \mathscr{L}_{i,t}(\bmx_{i}(t),\bmmu_i(t)) \right\|
\label{lemma-dolr-constraints-disagree-4}
\end{aligned}
\end{equation}
where the inequality follows from the nonexpansiveness of the Euclidean projection and last equality from the doubly stochasticity of $W_{\mathrm{G}}$. Combining the equations in (\ref{lemma-dolr-constraints-disagree-2}), (\ref{lemma-dolr-constraints-disagree-3}) and (\ref{lemma-dolr-constraints-disagree-4}) and following similar lines as that of Lemma \ref{lemma-ls-grad}(ii), we obtain
\begin{equation}
\begin{aligned}[b]
\sum_{t=1}^{T} \sum_{i=1}^{n} \big\| \bmx_{i}(t) - \bmx_{\mathrm{avg}}(t)  \big\|
\leq \sum_{i=1}^{n}  \left\| \bmx_{i}(1) - \bmx_{\mathrm{avg}}(1) \right\| +    \sqrt{n}  \frac{\sigma_2(W_{\mathrm{G}})}{1 - \sigma_2(W_{\mathrm{G}})}  \left( \sum_{i=1}^{n}  \| \bmx_{i}(1)  \| \right) & \\
+ n \left( \frac{3 \sqrt{n}}{1-\sigma_2(W_{\mathrm{G}})} +4 \right) L \cdot \eta T + \left( \frac{3 \sqrt{n}}{1-\sigma_2(W_{\mathrm{G}})} +4 \right) K_{\mathrm{I}} \sum_{t=1}^{T-1} \sum_{i=1}^{n} \sum_{q=1}^{s} \eta \left[ \bmmu_i(t) \right]_q &
\label{lemma-dolr-constraints-disagree-5}
\end{aligned}
\end{equation}
which completes the proof in part (i).

(ii)
Following similar analysis as that of part (i), we have
\begin{equation}
\begin{aligned}[b]
\bmx_{\mathrm{avg}}(t+1) &= \bmx_{\mathrm{avg}}(1)  - \eta \sum_{\ell=1}^{t}  \frac{1}{n} \sum_{i=1}^{n} \nabla_{\bmy} \mathscr{L}^{\mathrm{b}}_{i,t}(\bmx_{i}(\ell),\bmmu_i(\ell)) + \sum_{\ell=1}^{t}  \frac{1}{n} \sum_{i=1}^{n} \mathbf{p}_i^{\mathrm{b}}(\ell) \\
\bmx_{i}(t+1)  &= \sum_{j=1}^{n} [W_{\mathrm{G}}^t]_{ij} \bmx_{j}(1)  - \eta \sum_{\ell = 1}^{t}  \sum_{j=1}^{n} [W_{\mathrm{G}}^{t+1-\ell}]_{ij} \nabla_{\bmy} \mathscr{L}^{\mathrm{b}}_{j,t}(\bmx_{j}(\ell),\bmmu_j(\ell)) \\
&\quad+ \sum_{\ell = 1}^{t-1}  \sum_{j=1}^{n} [W_{\mathrm{G}}^{t-\ell}]_{ij} \mathbf{p}_j^{\mathrm{b}}(\ell) + \mathbf{p}_i^{\mathrm{b}}(t)
\label{lemma-dolr-constraints-disagree-6}
\end{aligned}
\end{equation}
where $\mathbf{p}_i^{\mathrm{b}}(t) = \mathpzc{P}_{(1-\xi)\mathbb{B}_{R}^{m}}\left( \sum_{j=1}^{n} [W_{\mathrm{G}}]_{ij} \mathpzc{l}_{j}(t) \right) - \sum_{j=1}^{n} [W_{\mathrm{G}}]_{ij} \mathpzc{l}_{j}(t) $, which satisfies
\begin{equation}
\begin{aligned}[b]
\sum_{i=1}^{n} \big\| \mathbf{p}_i^{\mathrm{b}}(t) \big\| & \leq  2  \eta \sum_{i=1}^{n} \big\| \nabla_{\bmy} \mathscr{L}^{\mathrm{b}}_{i,t}(\bmx_{i}(t),\bmmu_i(t)) \big\|.
\label{lemma-dolr-constraints-disagree-7}
\end{aligned}
\end{equation}
We are left to bound the term $\big\| \nabla_{\bmy} \mathscr{L}^{\mathrm{b}}_{i,t}(\bmx_{i}(t),\bmmu_i(t)) \big\|$,
\begin{equation}
\begin{aligned}[b]
\big\| \nabla_{\bmy} \mathscr{L}^{\mathrm{b}}_{i,t}(\bmx_{i}(t),\bmmu_i(t)) \big\|
&= \bigg\| \mathpzc{g}_{i,t} ( \bmx_{i}(t) ) + \sum_{q=1}^{s} [\bmmu_i(t)]_q \partial\left[ \bmk_q^{\tp}\bmx_i(t) \right]_+ \bigg\|\\ &\leq \| \mathpzc{g}_{i,t} ( \bmx_{i}(t) ) \| + K_{\mathrm{I}} \sum_{q=1}^{s} [\bmmu_i(t)]_q  \\
&\leq m L + K_{\mathrm{I}} \sum_{q=1}^{s} [\bmmu_i(t)]_q
\label{lemma-dolr-constraints-disagree-8}
\end{aligned}
\end{equation}
where in the last inequality we have combined Lemma \ref{lemma-bandit-grad-est} with the boundedness of the gradinet of $\theta_{i,t}(\bmy)$ (\ref{bound-on-gradient-constraints}) to get a new bound $\| \mathpzc{g}_{i,t} ( \bmx_{i}(t) ) \| \leq m L$. Hence, the desired result follows by combining (\ref{lemma-dolr-constraints-disagree-6}), (\ref{lemma-dolr-constraints-disagree-7}) and (\ref{lemma-dolr-constraints-disagree-8}). The proof is complete.
\hfill$\square$

\subsection{Proof of Theorem \ref{theorem-dolr-constraints}}
It follows from Algorithm \ref{alg-dolr-cons} that
\begin{equation}
\begin{aligned}[b]
\sum_{i=1}^{n} \| \bmx_{i}(t+1) - \bmy^\star_{\mathrm{LSC}} \|^2
&= \sum_{i=1}^{n} \bigg\| \mathpzc{P}_{\mathbb{B}_{R}^{m}} \Big(  \sum_{j=1}^{n} [W_{\mathrm{G}}]_{ij} \mathpzc{l}_{j}(t) \Big) - \bmy^\star_{\mathrm{LSC}} \bigg\|^2\\
&\leq  \sum_{i=1}^{n} \sum_{j=1}^{n} [W_{\mathrm{G}}]_{ij} \left\|   \mathpzc{l}_{j}(t) - \bmy^\star_{\mathrm{LSC}} \right\|^2  \\
&= \sum_{i=1}^{n} \left\| \mathpzc{l}_{i}(t) - \bmy^\star_{\mathrm{LSC}} \right\|^2
\label{theorem-dolr-constraints-disagree-1}
\end{aligned}
\end{equation}
where in the first equality we used the fact that $\mathcal{K}\subseteq \mathbb{B}_{R}^{m}$. Expanding the last term on the right side further gives
\begin{equation}
\begin{aligned}[b]
\sum_{j=1}^{n} \| \bmx_{j}(t+1) - \bmy^\star_{\mathrm{LSC}} \|^2
&\leq \sum_{j=1}^{n} \left\| \bmx_{j}(t) - \eta   \nabla_{\bmy} \mathscr{L}_{j,t}(\bmx_{j}(t),\bmmu_j(t)) - \bmy^\star_{\mathrm{LSC}} \right\|^2 \\
&\leq  \sum_{j=1}^{n} \| \bmx_{j}(t) - \bmy^\star_{\mathrm{LSC}} \|^2 + \eta^2  \sum_{j=1}^{n} \left\| \nabla_{\bmy} \mathscr{L}_{j,t}(\bmx_{j}(t),\bmmu_j(t)) \right\|^2 \\
&\quad- 2 \eta \sum_{j=1}^{n} \left[ \mathscr{L}_{j,t}(\bmx_{j}(t),\bmmu_j(t)) - \mathscr{L}_{j,t}(\bmy^\star_{\mathrm{LSC}},\bmmu_j(t)) \right]
\label{theorem-dolr-constraints-disagree-2}
\end{aligned}
\end{equation}
where the last inequality is based on the convexity of $\mathscr{L}_{j,t}$ with respect to the first argument. The term $\sum_{j=1}^{n} \left[ \mathscr{L}_{j,t}(\bmx_{j}(t),\bmmu_j(t)) - \mathscr{L}_{j,t}(\bmy^\star_{\mathrm{LSC}},\bmmu_j(t)) \right]$ can be expanded as follows, by using the definition of $\mathscr{L}_{j,t}$ (cf. (\ref{regularized-lagrangian})):
\begin{equation}
\begin{aligned}[b]
&\sum_{j=1}^{n} \left[ \mathscr{L}_{j,t}(\bmx_{j}(t),\bmmu_j(t)) - \mathscr{L}_{j,t}(\bmy^\star_{\mathrm{LSC}},\bmmu_j(t)) \right] \\
&= \sum_{j=1}^{n} \left[ \theta_{j,t}(\bmx_{j}(t)) - \theta_{j,t}(\bmy^\star_{\mathrm{LSC}}) \right] + \sum_{j=1}^{n} \sum_{q=1}^{s} [\bmmu_j(t)]_q \left[ \bmk_q^{\tp}\bmx_{j}(t) \right]_+ - \sum_{j=1}^{n} \sum_{q=1}^{s} [\bmmu_j(t)]_q \left[ \bmk_q^{\tp}\bmy^\star_{\mathrm{LSC}} \right]_+ \\
&\geq \sum_{j=1}^{n} \left[ \theta_{j,t}(\bmx_{i}(t)) - \theta_{j,t}(\bmy^\star_{\mathrm{LSC}}) \right] -  L \sum_{j=1}^{n} \left\| \bmx_{j}(t) - \bmx_{i}(t) \right\|   + \frac{1}{\pi} \sum_{j=1}^{n} \sum_{q=1}^{s} \left[ \bmk_q^{\tp}\bmx_{j}(t) \right]_+^2
\label{theorem-dolr-constraints-disagree-3}
\end{aligned}
\end{equation}
where the last inequality follows from (\ref{bound-on-gradient-constraints}), Step 4 in Algorithm \ref{alg-dolr-cons}, and the fact that $\bmk_q^{\tp}\bmy^\star_{\mathrm{LSC}} \leq 0$ for all $q=1,\ldots,s$. Summing the inequalities in (\ref{theorem-dolr-constraints-disagree-2}) over $t=1$ to $T$ and using (\ref{theorem-dolr-constraints-disagree-3}), yields
\begin{equation}
\begin{aligned}[b]
\reg_{\mathrm{LSC}}(i,T)
&\leq \frac{1}{2 \eta}  \sum_{j=1}^{n} \| \bmx_{j}(1) - \bmy^\star_{\mathrm{LSC}} \|^2
+ \frac{\eta}{2}  \sum_{t=1}^{T} \sum_{j=1}^{n} \left\| \nabla_{\bmy} \mathscr{L}_{j,t}(\bmx_{j}(t),\bmmu_j(t)) \right\|^2 \\
&\quad+ L \sum_{t=1}^{T}  \sum_{j=1}^{n} \left\| \bmx_{j}(t) - \bmx_{i}(t) \right\|
- \frac{1}{\pi} \sum_{t=1}^{T} \sum_{j=1}^{n} \sum_{q=1}^{s} \left[ \bmk_q^{\tp}\bmx_{j}(t) \right]_+^2.
\label{theorem-dolr-constraints-disagree-4}
\end{aligned}
\end{equation}
We now bound the second and third terms on the right-hand side of (\ref{theorem-dolr-constraints-disagree-4}). Note that
\begin{equation}
\begin{aligned}[b]
\left\| \nabla_{\bmy} \mathscr{L}_{j,t}(\bmx_{j}(t),\bmmu_j(t)) \right\|^2
&= \bigg\| \bmh_{j}(t) \left(  \bmh_{j}(t)^\tp \bmx_{j}(t) - z_{j}(t)  \right)
+ \sum_{q=1}^{s} [\bmmu_j(t)]_q \partial\left[ \bmk_q^{\tp}\bmx_j(t) \right]_+ \bigg\|^2 \\
&\leq  2 L^2 + 2 s K_{\mathrm{I}}^2 \sum_{q=1}^{s} \left\|  [\bmmu_j(t)]_q \right\|^2
\leq 2 L^2 +  \frac{2 s K_{\mathrm{I}}^2}{\pi^2} \sum_{q=1}^{s} \left[ \bmk_q^{\tp}\bmx_j(t) \right]_+^2
\label{theorem-dolr-constraints-disagree-5}
\end{aligned}
\end{equation}
where the inequality is based on Assumption \ref{assumption-constraints} and Step 4 in Algorithm \ref{alg-dolr-cons}. The third term on the right-hand side of (\ref{theorem-dolr-constraints-disagree-4}) can be bounded by Lemma \ref{lemma-dolr-constraints-disagree}(i), that is,
\begin{equation}
\begin{aligned}[b]
\sum_{t=1}^{T} \sum_{j=1}^{n} \left\| \bmx_{j}(t) - \bmx_{i}(t) \right\|
&  \leq 2 n \sum_{t=1}^{T} \sum_{j=1}^{n} \big\| \bmx_{j}(t) - \bmx_{\mathrm{avg}}(t)  \big\| \\
&  \leq 2 n C_3 + 2 n^2 L E  \cdot \eta T +    \frac{2 n K_{\mathrm{I}} E \eta }{\pi} \sum_{t=1}^{T} \sum_{i=1}^{n} \sum_{q=1}^{s} \left[ \bmk_q^{\tp}\bmx_i(t) \right]_+.
\label{theorem-dolr-constraints-disagree-6}
\end{aligned}
\end{equation}
Combining the inequalities in (\ref{theorem-dolr-constraints-disagree-4}), (\ref{theorem-dolr-constraints-disagree-5}) and ((\ref{theorem-dolr-constraints-disagree-6})), we further obtain
\begin{equation}
\begin{aligned}[b]
\reg_{\mathrm{LSC}}(i,T)
&\leq 2 n L C_3 + \frac{1}{2 \eta}  \sum_{j=1}^{n} \| \bmx_{j}(1) - \bmy^\star_{\mathrm{LSC}} \|^2
+ n L^2 \left( 1 + 2 n E\right)\cdot \eta T \\
&\quad+ \frac{2 n L K_{\mathrm{I}} E  \eta }{\pi} \sum_{t=1}^{T} \sum_{j=1}^{n} \sum_{q=1}^{s} \left[ \bmk_q^{\tp}\bmx_j(t) \right]_+
- \frac{\pi - s K_{\mathrm{I}}^2  \eta}{\pi^2}  \sum_{t=1}^{T} \sum_{j=1}^{n} \sum_{q=1}^{s} \left[ \bmk_q^{\tp}\bmx_{j}(t) \right]_+^2   \\
&\leq 2 n L C_3 + \frac{1}{2 \eta}  \sum_{j=1}^{n} \| \bmx_{j}(1) - \bmy^\star_{\mathrm{LSC}} \|^2
+ n L^2 \left( 1 + 2 n E\right)\cdot \eta T \\
&\quad+  \underbrace{\frac{2 n L K_{\mathrm{I}} E  \eta }{\pi} \cdot \cvs(s,T) - \frac{\pi - s K_{\mathrm{I}}^2  \eta}{s n T \pi^2} \cdot \left( \cvs(s,T) \right)^2}_{\triangleq  f (\cvs(s,T))}
\label{theorem-dolr-constraints-disagree-7}
\end{aligned}
\end{equation}
where in the last inequality we used the following fact,
\begin{equation*}
\begin{aligned}[b]
\sum_{t=1}^{T} \sum_{j=1}^{n} \sum_{q=1}^{s} \left[ \bmk_q^{\tp}\bmx_{j}(t) \right]_+^2
&\geq \frac{1}{s n T} \left( \sum_{t=1}^{T} \sum_{j=1}^{n} \sum_{q=1}^{s} \left[ \bmk_q^{\tp}\bmx_{j}(t) \right]_+ \right)^2
= \frac{1}{s n T} \left( \cvs(s,T) \right)^2.
\end{aligned}
\end{equation*}
We are left to bound the term $f (\cvs(s,T))$, which is a quadratic function of $\cvs(s,T)$. By substituting the expressions for $\eta = \frac{1}{c s K_{\mathrm{I}}^2 T^\beta}$ and $\pi = \frac{1}{T^\beta}$ into $f (\cvs(s,T))$ it follows that
\begin{equation}
\begin{aligned}[b]
f (\cvs(s,T)) =  \frac{2 n L E }{c s K_{\mathrm{I}}} \cdot \cvs(s,T) - \frac{c - 1}{c s n T^{1-\beta}} \cdot \left( \cvs(s,T) \right)^2
\leq \frac{n^3 L^2 E^2}{c (c-1) s K_{\mathrm{I}}^2} \cdot T^{1-\beta}
\label{theorem-dolr-constraints-disagree-8}
\end{aligned}
\end{equation}
this, combined with (\ref{theorem-dolr-constraints-disagree-7}), gives
\begin{equation}
\begin{aligned}[b]
\reg_{\mathrm{LSC}}(i,T)
&\leq 2 n L C_3 + 2 c s n K_{\mathrm{I}}^2 R^2 \cdot T^\beta
+ \left( \frac{ n L^2 \left( 1 + 2 n E\right)}{c s K_{\mathrm{I}}^2}  +  \frac{n^3 L^2 E^2}{c (c-1) s K_{\mathrm{I}}^2} \right) \cdot T^{1-\beta}.
\label{theorem-dolr-constraints-disagree-9}
\end{aligned}
\end{equation}
Hence, the proof of the regret bound is complete. We now turn our attention to bound the constraint violations $\cvs(s,T)$. It follows from inequalities (\ref{theorem-dolr-constraints-disagree-2}), (\ref{theorem-dolr-constraints-disagree-3}), (\ref{theorem-dolr-constraints-disagree-5}) and (\ref{theorem-dolr-constraints-disagree-6}) that
\begin{equation}
\begin{aligned}[b]
&\sum_{t=1}^{T} \sum_{i=1}^{n} \left[ \theta_{i,t}(\bmx_{i}(t)) - \theta_{i,t}(\bmy^\star_{\mathrm{LSC}}) \right]\\
&\leq  \frac{1}{2 \eta}  \sum_{i=1}^{n} \| \bmx_{i}(1) - \bmy^\star_{\mathrm{LSC}} \|^2
+ n L^2 \cdot \eta T - \frac{\pi - s K_{\mathrm{I}}^2  \eta}{s n T \pi^2} \cdot \cvs(s,T)^2 \\
&\leq 2 c s n K_{\mathrm{I}}^2 R^2 \cdot T^\beta + \frac{ n L^2 }{c s K_{\mathrm{I}}^2} \cdot T^{1-\beta} - \frac{c - 1}{c s n T^{1-\beta}} \cdot \cvs(s,T)^2
\label{theorem-dolr-constraints-disagree-10}
\end{aligned}
\end{equation}
the left-hand side on (\ref{theorem-dolr-constraints-disagree-10}) can be lower bounded by using the boundedness of the gradient of $\theta_{i,t}$ (cf. (\ref{bound-on-gradient-constraints})), that is,
\begin{equation}
\begin{aligned}[b]
\sum_{t=1}^{T} \sum_{i=1}^{n} \left[ \theta_{i,t}(\bmx_{i}(t)) - \theta_{i,t}(\bmy^\star_{\mathrm{LSC}}) \right]
&\geq   - n L T \left\| \bmx_{i}(t) - \bmy^\star_{\mathrm{LSC}} \right\|
\geq - 2 n L R \cdot T.
\label{theorem-dolr-constraints-disagree-11}
\end{aligned}
\end{equation}
Combining (\ref{theorem-dolr-constraints-disagree-10}) and (\ref{theorem-dolr-constraints-disagree-11}), we have the following bound on the constraint violations,
\begin{equation*}
\begin{aligned}[b]
\cvs(s,T)
&\leq  \sqrt{\frac{2}{c-1}} c s n K_{\mathrm{I}} R \cdot T^{1/2} +  \sqrt{\frac{1}{c-1}} \frac{n L}{K_{\mathrm{I}}} T^{1-\beta} + \sqrt{\frac{2}{c-1}} n \sqrt{c s L R} \cdot T^{1 - \beta/2} \\
&\leq \sqrt{\frac{1}{c-1}} n \left( 2 c s K_{\mathrm{I}} R + \frac{L}{K_{\mathrm{I}}} + 2 \sqrt{c s L R} \right) \cdot T^{1 - \beta/2}
\end{aligned}
\end{equation*}
where the last inequality is based on the fact that $\beta \in (0,1)$. This complete the proof.

\subsection{Proof of Theorem \ref{theorem-dolr-constraints-bandit}}
We first provide a new bound on $\big| \hat{\theta}_{i,t} ( \bmy) -  \theta_{i,t} ( \bmy ) \big|$, instead of Lemma \ref{lemma-bandit-grad-est}(ii),
\begin{equation}
\begin{aligned}[b]
& \big| \hat{\theta}_{i,t} ( \bmy) -  \theta_{i,t} ( \bmy ) \big|
= \left| \expect_{\mathbf{v}\in\mathbb{B}_1^m} \big[ \theta_{i,t}(\bmy  +  \epsilon  \mathbf{v}) \big] -  \theta_{i,t} ( \bmy ) \right|
\leq \expect_{\mathbf{v}\in\mathbb{B}_1^m}  \left[ \left| \theta_{i,t}(\bmy  +  \epsilon  \mathbf{v})  -  \theta_{i,t} ( \bmy ) \right| \right] \leq L \epsilon
\label{theorem-dolr-constraints-disagree-12}
\end{aligned}
\end{equation}
where we have used (\ref{bound-on-gradient-constraints}). Following an argument similar to that of part (i) and using Lemma \ref{lemma-bandit-grad-est}(i), we have
\begin{equation}
\begin{aligned}[b]
&\sum_{j=1}^{n} \expect\left[  \| \bmx_{j}(t+1) - (1- \xi)\bmy^\star_{\mathrm{LSC}} \|^2 \right] \\
&\leq  \sum_{j=1}^{n} \expect\left[ \| \bmx_{j}(t) - (1- \xi) \bmy^\star_{\mathrm{LSC}} \|^2 \right] + \eta^2  \sum_{j=1}^{n} \expect\left[ \big\| \nabla_{\bmy} \mathscr{L}^{\mathrm{b}}_{j,t}(\bmx_{j}(t),\bmmu_j(t)) \big\|^2 \right]  \\
&\quad- 2 \eta \sum_{j=1}^{n} \expect \left[ \mathscr{L}^{\mathrm{b}}_{j,t}(\bmx_{j}(t),\bmmu_j(t)) - \hat{\mathscr{L}}_{j,t}((1- \xi) \bmy^\star_{\mathrm{LSC}},\bmmu_j(t)) \right]
\label{theorem-dolr-constraints-disagree-13}
\end{aligned}
\end{equation}
we can expand the term $\sum_{i=1}^{n} \left[ \mathscr{L}^{\mathrm{b}}_{i,t}(\bmx_{i}(t),\bmmu_i(t)) - \mathscr{L}^{\mathrm{b}}_{i,t}((1-\xi)\bmy^\star_{\mathrm{LSC}},\bmmu_i(t)) \right]$ as follows, by using (\ref{regularized-lagrangian-smoothed}):
\begin{equation}
\begin{aligned}[b]
&\sum_{j=1}^{n} \left[ \mathscr{L}^{\mathrm{b}}_{j,t}(\bmx_{j}(t),\bmmu_j(t)) - \mathscr{L}^{\mathrm{b}}_{j,t}((1-\xi)\bmy^\star_{\mathrm{LSC}},\bmmu_j(t)) \right] \\
&= \sum_{j=1}^{n} \left[ \hat{\theta}_{j,t}(\bmx_{j}(t)) - \hat{\theta}_{j,t}((1-\xi)\bmy^\star_{\mathrm{LSC}}) \right] + \sum_{j=1}^{n} \sum_{q=1}^{s} [\bmmu_j(t)]_q \left[ \bmk_q^{\tp}\bmx_{j}(t) \right]_+  \\
&\geq \sum_{j=1}^{n} \left[ \theta_{j,t}(\bmx_{j}(t)) - \theta_{j,t}(\bmy^\star_{\mathrm{LSC}}) \right]   - 2 n L \epsilon - n L R \xi + \frac{1}{\pi} \sum_{j=1}^{n} \sum_{q=1}^{s} \left[ \bmk_q^{\tp}\bmx_{j}(t) \right]_+^2  \\
&\geq \sum_{j=1}^{n} \left[ \theta_{j,t}(\bmx_{i}(t)) - \theta_{j,t}(\bmy^\star_{\mathrm{LSC}}) \right] -  L \sum_{j=1}^{n} \left\| \bmx_{j}(t) - \bmx_{i}(t) \right\| - 2 n L \epsilon - n L R \xi + \frac{1}{\pi} \sum_{j=1}^{n} \sum_{q=1}^{s} \left[ \bmk_q^{\tp}\bmx_{j}(t) \right]_+^2
\label{theorem-dolr-constraints-disagree-14}
\end{aligned}
\end{equation}
where the equality follows from $\left[ \bmk_q^{\tp} (1-\xi) \bmy^\star_{\mathrm{LSC}} \right]_+ \leq 0$ for all $q=1,\ldots,s$ and the first inequality from (\ref{theorem-dolr-constraints-disagree-12}) and the bound $\theta_{j,t}((1-\xi)\bmy^\star_{\mathrm{LSC}}) \leq \theta_{j,t}(\bmy^\star_{\mathrm{LSC}}) + L \| \bmy^\star_{\mathrm{LSC}}\| \xi \leq \theta_{j,t}(\bmy^\star_{\mathrm{LSC}}) + L R \xi$. Hence, it follows from (\ref{theorem-dolr-constraints-disagree-13}) and (\ref{theorem-dolr-constraints-disagree-14}) that
\begin{equation}
\begin{aligned}[b]
&\expect\left[ \reg_{\mathrm{LSC}}(i,T) \right]  \\
&\leq \frac{1}{2 \eta}  \sum_{j=1}^{n} \| \bmx_{j}(1) - (1-\xi)\bmy^\star_{\mathrm{LSC}} \|^2
+ \frac{\eta}{2}  \sum_{t=1}^{T} \sum_{j=1}^{n} \expect\left[ \big\| \nabla_{\bmy} \mathscr{L}^{\mathrm{b}}_{j,t}(\bmx_{j}(t),\bmmu_j(t)) \big\|^2 \right] \\
&\quad+ L \sum_{t=1}^{T}  \sum_{j=1}^{n} \expect\left[ \left\| \bmx_{j}(t) - \bmx_{i}(t) \right\| \right]
+ 2 n L \cdot \epsilon T + n L R \cdot \xi T  - \frac{1}{\pi} \sum_{t=1}^{T} \sum_{j=1}^{n} \sum_{q=1}^{s} \expect\left[  \left[ \bmk_q^{\tp}\bmx_{j}(t) \right]_+^2  \right].
\label{theorem-dolr-constraints-disagree-15}
\end{aligned}
\end{equation}
Combining (\ref{theorem-dolr-constraints-disagree-15}) with the estimate (\ref{lemma-dolr-constraints-disagree-8}) and Lemma \ref{lemma-dolr-constraints-disagree}(ii), and then following similar analysis as that of part (i), we arrive at
\begin{equation}
\begin{aligned}[b]
\expect\left[ \reg_{\mathrm{LSC}}(i,T) \right]
\leq 2 n L B_3 + \frac{1}{2 \eta}  \sum_{j=1}^{n} \expect\left[ \| \bmx_{j}(1) - (1-\xi)\bmy^\star_{\mathrm{LSC}} \|^2 \right]
+  m n L^2 \left( m + 2 n E\right)\cdot \eta T  & \\
+ 2 n L \cdot \epsilon T + n L R \cdot \xi T + \underbrace{\frac{2 n L K_{\mathrm{I}} E  \eta }{\pi} \expect\left[ \cvs(s,T) \right]
- \frac{\pi - s K_{\mathrm{I}}^2  \eta}{s n T \pi^2}  \left( \expect\left[ \cvs(s,T) \right]\right)^2}_{=  f (\expect\left[ \cvs(s,T) \right])}. &
\label{theorem-dolr-constraints-disagree-16}
\end{aligned}
\end{equation}
By substituting the expressions for $\eta = \frac{1}{c s K_{\mathrm{I}}^2 T^\beta}$, $\pi = \frac{1}{T^\beta}$, $\epsilon = \frac{1}{T^\gamma}$ and $\xi = \frac{1}{R T^\gamma}$ into (\ref{theorem-dolr-constraints-disagree-16}) and using the estimate (\ref{theorem-dolr-constraints-disagree-8}), it follows that
\begin{equation}
\begin{aligned}[b]
&\expect\left[ \reg_{\mathrm{LSC}}(i,T) \right]  \\
&\leq 2 n L B_3 + 2 c s n K_{\mathrm{I}}^2 R^2 \cdot T^\beta  + \left( \frac{ m n L^2 \left( m + 2 n E\right)}{c s K_{\mathrm{I}}^2} + \frac{n^3 L^2 E^2}{c (c-1) s K_{\mathrm{I}}^2} \right) \cdot T^{1-\beta}  + 3 n L \cdot T^{1-\gamma} .
\label{theorem-dolr-constraints-disagree-17}
\end{aligned}
\end{equation}
The regret analysis is complete by nothing that $\gamma \geq \beta$. We now turn to bound the constraint violations. It follows from (\ref{theorem-dolr-constraints-disagree-13}), (\ref{theorem-dolr-constraints-disagree-14}), (\ref{theorem-dolr-constraints-disagree-15}) and (\ref{theorem-dolr-constraints-disagree-16}) that
\begin{equation*}
\begin{aligned}[b]
& \sum_{t=1}^{T} \sum_{i=1}^{n} \expect\left[ \theta_{i,t}(\bmx_{i}(t)) - \theta_{i,t}(\bmy^\star_{\mathrm{LSC}}) \right] \\
&\leq  \frac{1}{2 \eta}  \sum_{i=1}^{n} \| \bmx_{i}(1) - (1-\xi)\bmy^\star_{\mathrm{LSC}} \|^2
+ m^2 n L^2 \cdot \eta T  + 2 n L \cdot \epsilon T + n L R \cdot \xi T \\
&\quad - \frac{\pi - s K_{\mathrm{I}}^2  \eta}{s n T \pi^2} \cdot \left( \expect\left[ \cvs(s,T) \right]\right)^2 \\
&\leq 2 c s n K_{\mathrm{I}}^2 R^2 \cdot T^\beta + \frac{ m^2 n L^2 }{c s K_{\mathrm{I}}^2} \cdot T^{1-\beta} + 3 n L \cdot T^{1-\gamma} - \frac{c - 1}{c s n T^{1-\beta}} \cdot  \left( \expect\left[ \cvs(s,T) \right]\right)^2
\end{aligned}
\end{equation*}
which, combined with (\ref{theorem-dolr-constraints-disagree-11}), yields
\begin{equation*}
\begin{aligned}[b]
& \expect\left[ \cvs(s,T) \right] \\
&\leq  \sqrt{\frac{2}{c-1}} c s n K_{\mathrm{I}} R \cdot T^{1/2} \!+\!  \sqrt{\frac{1}{c-1}} \frac{m n L}{K_{\mathrm{I}}} T^{1-\beta} \!+\! \sqrt{\frac{2}{c-1}} n \sqrt{c s L R} \cdot T^{1 - \beta/2}  \\
&\quad \!+\! \sqrt{\frac{3}{c-1}} n \sqrt{c s L} \cdot  T^{1 - (\beta+\gamma)/2 } \\
&\leq \sqrt{\frac{1}{c-1}} n \left( 2 c s K_{\mathrm{I}} R + \frac{m L}{K_{\mathrm{I}}} + 2 \sqrt{c s L R} + 3 \sqrt{c s L} \right) \cdot T^{1 - \beta/2}.
\end{aligned}
\end{equation*}
The proof is complete.
\hfill$\square$





\newpage

\section{Proof of Theorem \ref{theorem-convergence-v2}}

From Lemma \ref{lemma-ls-grad}, we immediately have that, for any $\bmy^{\star} \in \mathcal{S}^{\star}(T)$,
\begin{equation*}
\begin{aligned}
\sum_{t=1}^{T} \sum_{i=1}^{n} \left\| \bmh_{i}(t) \left(  \bmh_{i}(t)^\tp \bmx_{i}(t) - z_{i}(t)  \right)  \right\|^2
&\leq \bar{C}_1 \cdot T^{\beta}  \\
\sum_{t=1}^{T} \sum_{i=1}^{n} \big\| \bmx_{i}(t) - \bmx_{\mathrm{avg}}(t)  \big\|
&\leq  C_3 + C_4 \cdot T^{\frac{1-\beta}{2}}
\label{theorem-convergence-v2-1}
\end{aligned}
\end{equation*}
with $ \bar{C}_1 = \frac{2^\beta  }{2^\beta - 1} \alpha_{\bmh}^2     \sum_{i=1}^{n} \mathrm{dist}^2(\bmx_{i}(1), \mathpzc{P}_{\mathcal{S}^{\star}(T)} \left( \bmx_{\mathrm{avg}}(1) \right) )  $, which is obtained by taking the minimization over $\bmy^{\star} \in \mathcal{S}^{\star}(T)$ in both sides of (\ref{lemma-ls-grad-6a}). Then following similar lines as that of the proof of Theorem \ref{theorem-ls-regret}, we find that ($\gamma \in \reals$)
\begin{equation}
\begin{aligned}[b]
&\left( 1-   \frac{1}{1 + T^{\gamma}}  \right)  \sum_{t=1}^{T} \sum_{j=1}^{n}  \theta_{j,t} \left( \bmx_{i}(t) \right)
- \sum_{t=1}^{T} \sum_{j=1}^{n} \theta_{j,t} \left( \bmy^\star \right)
\leq \frac{1}{2 \alpha_{\bmh}} \bar{C}_1   \\
&\qquad+ \frac{\alpha_{\bmh}}{2 }   \sum_{j=1}^{n} \mathrm{dist}^2(\bmx_{j}(1), \mathpzc{P}_{\mathcal{S}^{\star}(T)} \left( \bmx_{\mathrm{avg}}(1) \right) ) \cdot T^\beta + 4 \alpha_{\bmh} n C_3^2 \cdot T^{\gamma} + 4 \alpha_{\bmh} n C_4^2 \cdot T^{1-\beta+\gamma}
\label{theorem-convergence-v2-2}
\end{aligned}
\end{equation}
which implies
\begin{equation}
\begin{aligned}[b]
\sum_{t=1}^{T} \sum_{j=1}^{n}  \theta_{j,t} \left( \bmx_{i}(t) \right)
&\leq (1+T^{-\gamma}) \bigg(  \frac{1}{2 \alpha_{\bmh}} \bar{C}_1  + \frac{\alpha_{\bmh}}{2 }   \sum_{j=1}^{n} \mathrm{dist}^2(\bmx_{j}(1), \mathpzc{P}_{\mathcal{S}^{\star}(T)} \left( \bmx_{\mathrm{avg}}(1) \right) ) \cdot T^\beta \\
&\qquad\qquad\qquad\qquad\qquad+ 4 \alpha_{\bmh} n C_3^2 \cdot T^{\gamma} + 4 \alpha_{\bmh} n C_4^2 \cdot T^{1-\beta+\gamma} \bigg)
\label{theorem-convergence-v2-3}
\end{aligned}
\end{equation}
where used the fact that $\theta_{j,t}(\bmy^\star) =  0$ for all $j\in\mathrm{V}$ and $t=1,\ldots,T$. We now distinguish three cases: $\gamma > 0 $, $\gamma < 0$ and $\gamma = 0$. When $\gamma > 0 $, the dominant terms on the right-hand side of (\ref{theorem-convergence-v2-3}) are $T^\beta$ and $T^{1-\beta+\gamma}$, under which case the optimal bound is $\mathcal{O} \left(T^{\frac{1}{2}+\gamma}\right)$. Similarly, the optimal bound is $\mathcal{O} \left(T^{\frac{1}{2}-\gamma}\right)$ when $\gamma < 0 $. Therefore, it is easy to see that the optimal bound is achieved when $\gamma = 0$ and $\beta = \frac{1}{2}$.
\hfill$\square$





\newpage

\section{Proof of Theorem \ref{theorem-convergence}}

\subsection{Key Lemma}
\begin{lemma}\label{theorem-disagree}
Let Assumption \ref{assumption-LE-solution} hold.
Then along Algorithm \ref{alg-odle} there holds
\begin{equation*}
\begin{aligned}
\sum_{t=1}^{T}  \sum_{i=1}^{n}  \big\| \bmx_{i}(t) - \bmx_{\mathrm{avg}}(t) \big\|
&\leq   P_1  +  P_2  \cdot \sqrt{T}  ,\qquad\qquad T\geq 1
\end{aligned}
\end{equation*}
where
$
P_1 = C_3
$
and
$
P_2 = n \frac{  \sigma_2(W_{\mathrm{G}})}{ 1 - \sigma_2(W_{\mathrm{G}}) }     \sqrt{\sum_{i=1}^{n} \mathrm{dist}^2(\bmx_{i}(1), \mathpzc{P}_{\mathcal{S}^{\star}(T)} \left( \bmx_{\mathrm{avg}}(1) \right) )}.
$
\end{lemma}

\noindent{\em Proof.}
First, by an argument similar to that of Lemma \ref{lemma-ls-grad}(ii) it follows that
\begin{equation*}
\begin{aligned}
& \sum_{i=1}^{n} \big\| \bmx_{i}(t) - \bmx_{\mathrm{avg}}(t)  \big\|  \leq \sqrt{n}  \left( \sum_{i=1}^{n} \| \bmx_{i}(1)  \|  \right)   \sigma_2(W_{\mathrm{G}})^{t-1}
\\
&+  \sqrt{n}   \sum_{\ell = 1}^{t-1}  \sigma_2(W_{\mathrm{G}})^{t-\ell}  \sum_{i=1}^{n} \left\| \bmh_{i}(\ell) \left(  \bmh_{i}(\ell)^\tp \bmx_{i}(\ell) - \bmz_{i}(\ell)  \right) \right\|  \Big/  \| \bmh_{i}(\ell) \|^2.
\end{aligned}
\end{equation*}
Summing the preceding inequalities over $t=1$ to $t=T$, we have
\begin{equation}
\begin{aligned}[b]
&\sum_{t=1}^{T} \sum_{i=1}^{n} \big\| \bmx_{i}(t) - \bmx_{\mathrm{avg}}(t)  \big\|\\
&=  \sum_{i=1}^{n} \big\| \bmx_{i}(1) - \bmx_{\mathrm{avg}}(1)  \big\|  +  \sum_{t=2}^{T} \sum_{i=1}^{n} \big\| \bmx_{i}(t) - \bmx_{\mathrm{avg}}(t)  \big\| \\
&\leq  \sum_{i=1}^{n} \big\| \bmx_{i}(1) - \bmx_{\mathrm{avg}}(1)  \big\|
+  \sqrt{n}  \left( \sum_{i=1}^{n} \| \bmx_{i}(1)  \|  \right)    \sum_{t=2}^{T} \sigma_2(W_{\mathrm{G}})^{t-1} \\
&\quad +  \sqrt{n}   \sum_{t=2}^{T} \sum_{\ell = 1}^{t-1}  \sigma_2(W_{\mathrm{G}})^{t-\ell}    \sum_{i=1}^{n} \left\| \bmh_{i}(\ell) \left(  \bmh_{i}(\ell)^\tp \bmx_{i}(\ell) - \bmz_{i}(\ell)  \right) \right\|\Big/  \| \bmh_{i}(\ell) \|^2 .
\label{theorem-disagree-1}
\end{aligned}
\end{equation}
We are left to bound the term $\sum_{\ell=1}^{t} \sum_{i=1}^{n} \left\|  \bmh_{i}(\ell) \left(  \bmh_{i}(\ell)^\tp \bmx_{i}(\ell) - \bmz_{i}(\ell)  \right)  \right\|$. First, note that $\mathpzc{l}_{i,t}$ is the Euclidean projection of $\bmx_{i,t}$ on the hyperplane $\mathcal{A}_{i}(t) = \{ \bmy : z_{i}(t) = \bmh_{i}(t)^\tp \bmy \}$, we have that, for any $\bmy^{\star} \in \mathcal{S}^{\star}(T)$,
\begin{equation}
\begin{aligned}[b]
\sum_{i=1}^{n} \big\| \bmx_{i}(t+1) - \bmy^\star  \big\|^2
&=  \sum_{i=1}^{n} \Bigg\|   \sum_{j=1}^{n} [W_{\mathrm{G}}]_{ij} \mathpzc{P}_{\mathcal{A}_{j}(t) } \left( \bmx_{j}(t) \right)  - \bmy^\star  \Bigg\|^2\\&
\leq \sum_{i=1}^{n} \Bigg\|   \mathpzc{P}_{\mathcal{A}_{i}(t) } \left( \bmx_{i}(t) \right)  - \bmy^\star  \Bigg\|^2  \\
&\leq \sum_{i=1}^{n} \big\| \bmx_{i}(t) - \bmy^\star  \big\|^2
- \sum_{i=1}^{n} \big\| \bmx_{i}(t) - \mathpzc{P}_{\mathcal{A}_{i}(t)} \left( \bmx_{i}(t) \right)  \big\|^2
\label{lemma-disagree-pB2}
\end{aligned}
\end{equation}
where the last inequality is based on the fact that $\bmy^\star \in \mathcal{A}_{i}(t) $ and the following inequality,
\begin{equation}
\begin{aligned}[b]
\|  \mathpzc{P}_{\mathcal{X}} (\bmx)  -  \bmy \|^2
&\leq  \|  \bmx  -  \bmy \|^2  - \|  \mathpzc{P}_{\mathcal{X}} (\bmx)  -  \bmx \|^2,\qquad
\mathrm{for}\ \mathrm{any}\ \bmy\in\mathcal{X}\ \mathrm{and}\ \bmx \in\reals^{m}  .
\label{lemma-disagree-pB3}
\end{aligned}
\end{equation}
Applying inequality (\ref{lemma-disagree-pB2}) recursively and using $$
\bmx_{i}(t) - \mathpzc{P}_{\mathcal{A}_{i}(t)} \left( \bmx_{i}(t) \right) = \bmh_{i}(t) \left(  \bmh_{i}(t)^\tp \bmx_{i}(t) - \bmz_{i}(t)  \right) \big/  \| \bmh_{i}(t) \|^2,
$$ we find that
\begin{equation}
\begin{aligned}[b]
&\sum_{i=1}^{n} \big\| \bmx_{i}(t+1) - \bmy^\star  \big\|^2
\leq \sum_{i=1}^{n} \big\| \bmx_{i}(1) - \bmy^\star  \big\|^2\\&\quad
- \sum_{\ell=1}^{t} \sum_{i=1}^{n} \left\| \bmh_{i}(\ell) \left(  \bmh_{i}(\ell)^\tp \bmx_{i}(\ell) - \bmz_{i}(\ell)  \right)  \right\|^2  \Big/ \| \bmh_{i}(\ell) \|^4
\label{lemma-disagree-pB4}
\end{aligned}
\end{equation}
which implies that for all $t\geq 1$,
\begin{equation}
\begin{aligned}[b]
\sum_{\ell=1}^{t} \sum_{i=1}^{n} \left\| \bmh_{i}(\ell) \left(  \bmh_{i}(\ell)^\tp \bmx_{i}(\ell) - \bmz_{i}(\ell)  \right)  \right\|^2 \Big/   \| \bmh_{i}(\ell) \|^4
&\leq  \sum_{i=1}^{n} \big\| \bmx_{i}(1) - \bmy^\star  \big\|^2 .
\label{lemma-disagree-pB5}
\end{aligned}
\end{equation}
Taking the minimization over $\bmy^{\star} \in \mathcal{S}^{\star}(T)$ in both sides of the preceding relation, we further have
\begin{equation}
\begin{aligned}[b]
\sum_{\ell=1}^{t} \sum_{i=1}^{n} \left\| \bmh_{i}(\ell) \left(  \bmh_{i}(\ell)^\tp \bmx_{i}(\ell) - \bmz_{i}(\ell)  \right)  \right\|^2 \Big/   \| \bmh_{i}(\ell) \|^4
&\leq  \min_{\bmy^{\star} \in \mathcal{S}^{\star}(T)} \sum_{i=1}^{n}  \big\| \bmx_{i}(1) - \bmy^\star  \big\|^2  \\
&= \sum_{i=1}^{n} \mathrm{dist}^2(\bmx_{i}(1), \mathpzc{P}_{\mathcal{S}^{\star}(T)} \left( \bmx_{\mathrm{avg}}(1) \right) ) .
\label{lemma-disagree-pB6}
\end{aligned}
\end{equation}
This further leads to
\begin{equation*}
\begin{aligned}
&\sum_{\ell=1}^{T} \sum_{i=1}^{n} \left\| \bmh_{i}(\ell) \left(  \bmh_{i}(\ell)^\tp \bmx_{i}(\ell) - \bmz_{i}(\ell)  \right)  \right\| \Big/   \| \bmh_{i}(\ell) \|^2\\
&\leq  \sqrt{T n  \sum_{\ell=1}^{T} \sum_{i=1}^{n} \left\| \bmh_{i}(\ell) \left(  \bmh_{i}(\ell)^\tp \bmx_{i}(\ell) - \bmz_{i}(\ell)  \right)  \right\|^2  \Big/   \| \bmh_{i}(\ell) \|^4 }  \\
&\leq \sqrt{n}  \sqrt{\sum_{i=1}^{n} \mathrm{dist}^2(\bmx_{i}(1), \mathpzc{P}_{\mathcal{S}^{\star}(T)} \left( \bmx_{\mathrm{avg}}(1) \right) )} \cdot \sqrt{T}
\end{aligned}
\end{equation*}
then, combining this with (\ref{theorem-disagree-1}) and following similar lines as that of Lemma \ref{lemma-ls-grad}(ii), we arrive at the desired conclusion.
\hfill$\square$


\subsection{Proof of the theorem}
Using the updates in Algorithm \ref{alg-odle}, one has that, for any $\bmy^{\star} \in \mathcal{S}^{\star}(T)$,
\begin{equation}
\begin{aligned}[b]
\sum_{i=1}^{n} \big\| \bmx_{i}(t+1) - \bmy^\star  \big\|^2
&\leq \sum_{i=1}^{n} \Bigg\| \sum_{j=1}^{n} [W_{\mathrm{G}}]_{ij} \mathpzc{l}_{j}(t) - \bmy^\star  \Bigg\|^2
\leq \sum_{i=1}^{n} \big\| \mathpzc{l}_{i}(t) - \bmy^\star  \big\|^2   \\
&= \sum_{i=1}^{n} \Big\| \bmx_{i}(t) - \bmh_{i}(t) \left(  \bmh_{i}(t)^\tp \bmx_{i}(t) - z_{i}(t)  \right)\Big/ \|\bmh_{i}(t)\|^2  - \bmy^\star  \Big\|^2  \\
&=  \sum_{i=1}^{n} \big\| \bmx_{i}(t) - \bmy^\star  \big\|^2 + \sum_{i=1}^{n} \left\| \bmh_{i}(t) \left(  \bmh_{i}(t)^\tp \bmx_{i}(t) - z_{i}(t)  \right)  \right\|^2 \Big/ \|\bmh_{i}(t)\|^4   \\
&\quad - 2 \underbrace{\sum_{i=1}^{n} \left( \bmx_{i}(t) - \bmy^\star \right)^\tp  \bmh_{i}(t) \left(  \bmh_{i}(t)^\tp \bmx_{i}(t) - z_{i}(t)  \right) \Big/ \|\bmh_{i}(t)\|^2 }_{\triangleq  \mathcal{P}(t)  }
\label{theorem-convergence-1}
\end{aligned}
\end{equation}
where the first inequality follows from the same argument as that of (\ref{lemma-ls-grad-3}). We now provide a lower bound on $\sum_{t=1}^{T} \mathcal{P}(t)$,
\begin{equation}
\begin{aligned}[b]
\sum_{t=1}^{T} \mathcal{P}(t)
&= \sum_{t=1}^{T} \sum_{j=1}^{n} \left| \bmh_{j}(t)^\tp \bmx_{j}(t) - z_{j}(t)  \right|^2 \Big/ \|\bmh_{j}(t)\|^2\\
&= \sum_{t=1}^{T} \sum_{j=1}^{n}\left| \bmh_{j}(t)^\tp \left(  \bmx_{j}(t) - \bmy^\star \right)  \right|^2 \Big/ \|\bmh_{j}(t)\|^2 \\
&\geq \frac{1}{n T}  \left( \sum_{t=1}^{T} \sum_{j=1}^{n} \left| \bmh_{j}(t)^\tp \left(  \bmx_{j}(t) - \bmy^\star \right)  \right| \Big/ \| \bmh_{j}(t) \| \right)^2
\label{theorem-convergence-2}
\end{aligned}
\end{equation}
where in the second equality we used $z_{i}(t) = \bmh_{i}(t)^\tp \bmy^\star $ and in the inequality we used the relation $(\sum_{i=1}^{n} a_i) ^2  \leq  n \sum_{i=1}^{n} a_i^2  $ for any $a_i \in \reals$, $i=1,\ldots,n$. By adding and subtracting $\bmx_{i}(t)$, we further have
\begin{equation}
\begin{aligned}[b]
&\sqrt{ n T \sum_{t=1}^{T} \mathcal{P}(t) }
\geq  \sum_{t=1}^{T} \sum_{j=1}^{n} \left| \bmh_{j}(t)^\tp \left(  \bmx_{i}(t) - \bmy^\star + \bmx_{j}(t) - \bmx_{i}(t) \right)  \right| \Big/ \| \bmh_{j}(t) \|  \\
&\geq  \sum_{t=1}^{T} \sum_{j=1}^{n} \left| \bmh_{j}(t)^\tp \left(  \bmx_{i}(t) - \bmy^\star \right) \right| \Big/ \| \bmh_{j}(t) \|
-  \sum_{t=1}^{T} \sum_{j=1}^{n} \left|  \bmh_{j}(t)^\tp \left( \bmx_{j}(t) - \bmx_{i}(t) \right) \right| \Big/ \| \bmh_{j}(t) \| \\
&\geq  \reg_{\ell_1}(i,T)  -  \sum_{t=1}^{T} \sum_{j=1}^{n} \left| \bmh_{j}(t)^\tp \left( \bmx_{j}(t) - \bmx_{i}(t) \right)  \right| \Big/ \| \bmh_{j}(t) \|
\label{theorem-convergence-3}
\end{aligned}
\end{equation}
where in the second inequality we used the relation $|  a - b | \geq | a | - | b |$ for any $a,b\in\reals$. Now we turn our attention to the last term on the right-hand side of (\ref{theorem-convergence-3}),
\begin{equation}
\begin{aligned}[b]
&  \sum_{t=1}^{T} \sum_{j=1}^{n} \left| \bmh_{j}(t)^\tp \left( \bmx_{j}(t) - \bmx_{i}(t) \right)  \right| \Big/ \| \bmh_{j}(t) \|
\leq  \sum_{t=1}^{T} \sum_{j=1}^{n}  \big\| \bmx_{j}(t) - \bmx_{i}(t) \big\|  \\
&\leq  \sum_{t=1}^{T} \sum_{j=1}^{n}   \big\| \bmx_{j}(t) - \bmx_{\mathrm{avg}}(t) \big\|  +   \sum_{t=1}^{T} \sum_{j=1}^{n} \big\| \bmx_{\mathrm{avg}}(t) - \bmx_{i}(t) \big\| \\
&\leq 2 n  \sum_{t=1}^{T} \sum_{j=1}^{n}   \big\| \bmx_{j}(t) - \bmx_{\mathrm{avg}}(t) \big\|
\leq 2 n P_1 + 2 n P_2 \sqrt{T}
\label{theorem-convergence-4}
\end{aligned}
\end{equation}
where the second-to-last inequality follows from $\| \bmx_{\mathrm{avg}}(t) - \bmx_{i}(t) \| \leq \sum_{j=1}^{n} \| \bmx_{j}(t) - \bmx_{\mathrm{avg}}(t) \|$. On the other hand, summing the inequalities in (\ref{theorem-convergence-1}) over $t=1$ to $t=T$ and using the results in (\ref{theorem-convergence-3}) and (\ref{theorem-convergence-4}), we obtain
\begin{equation}
\begin{aligned}[b]
\sum_{t=1}^{T} \mathcal{P}(t)
&\leq \frac{1}{2} \sum_{t=1}^{T} \left( \sum_{i=1}^{n} \big\| \bmx_{i}(t) - \bmy^\star  \big\|^2 - \sum_{i=1}^{n} \big\| \bmx_{i}(t+1) - \bmy^\star  \big\|^2 \right)  \\
&\quad + \frac{1}{2} \sum_{t=1}^{T} \sum_{i=1}^{n} \left\| \bmh_{i}(t) \left(  \bmh_{i}(t)^\tp \bmx_{i}(t) - z_{i}(t)  \right)  \right\|^2 \Big/ \|\bmh_{i}(t)\|^4    \\
&\leq \frac{1}{2}  \sum_{i=1}^{n} \big\| \bmx_{i}(1)  -  \bmy^\star \big\|^2 - \sum_{i=1}^{n} \big\| \bmx_{i}(T+1)  - \bmy^\star \big\|^2  + \frac{1}{2}  \sum_{i=1}^{n} \big\| \bmx_{i}(1)  -  \bmy^\star \big\|^2\\
&\leq  \sum_{i=1}^{n} \big\| \bmx_{i}(1) - \bmy^\star  \big\|^2
\label{theorem-convergence-5}
\end{aligned}
\end{equation}
where the second inequality is based on (\ref{lemma-disagree-pB5}). Following an argument similar to that of (\ref{lemma-disagree-pB6}), one has $\sum_{t=1}^{T} \mathcal{P}(t) \leq \sum_{i=1}^{n} \mathrm{dist}^2(\bmx_{i}(1), \mathpzc{P}_{\mathcal{S}^{\star}(T)} \left( \bmx_{\mathrm{avg}}(1) \right) )$.
This, combined with the inequalities (\ref{theorem-convergence-3}), (\ref{theorem-convergence-4}), and (\ref{theorem-convergence-5}), yields
\begin{equation}
\begin{aligned}[b]
\reg_{\ell_1}(i,T)
&\leq 2 n P_1 + \left( 2 n P_2 + \sqrt{n} \right)  \sqrt{\sum_{i=1}^{n} \mathrm{dist}^2(\bmx_{i}(1), \mathpzc{P}_{\mathcal{S}^{\star}(T)} \left( \bmx_{\mathrm{avg}}(1) \right) )} \cdot \sqrt{T}.
\label{theorem-convergence-6}
\end{aligned}
\end{equation}
The proof is complete.
\hfill$\square$


\begin{thebibliography}{10}

\bibitem{agarwal2010colt}
Alekh Agarwal, Ofer Dekel, and Lin Xiao.
\newblock Optimal algorithms for online convex optimization with multi-point
  bandit feedback.
\newblock In {\em COLT}, pages 28--40. Citeseer, 2010.

\bibitem{Bartlett15}
Peter Bartlett, Wouter Koolen, Alan Malek, Eiji Takimoto, and Manfred Warmuth.
\newblock Minimax fixed-design linear regression.
\newblock {\em Proceedings of Machine Learning Research}, 40:226--239, 2015.

\bibitem{boyd2006}
Stephen Boyd, Arpita Ghosh, Balaji Prabhakar, and Devavrat Shah.
\newblock Randomized gossip algorithms.
\newblock {\em IEEE Trans. on Information Theory}, 14:2508--2530, June 2006.

\bibitem{bubeck2017stoc}
S{\'e}bastien Bubeck, Yin~Tat Lee, and Ronen Eldan.
\newblock Kernel-based methods for bandit convex optimization.
\newblock In {\em Proceedings of the 49th Annual ACM SIGACT Symposium on Theory
  of Computing}, pages 72--85. ACM, 2017.

\bibitem{bianchi}
Nicolo Cesa-Bianchi, Philip~M Long, and Manfred~K Warmuth.
\newblock Worst-case quadratic loss bounds for prediction using linear
  functions and gradient descent.
\newblock {\em IEEE Transactions on Neural Networks}, 7(3):604--619, 1996.

\bibitem{duchi2012tac}
John~C Duchi, Alekh Agarwal, and Martin~J Wainwright.
\newblock Dual averaging for distributed optimization: Convergence analysis and
  network scaling.
\newblock {\em IEEE Transactions on Automatic control}, 57(3):592--606, 2012.

\bibitem{duchi2015}
John~C. Duchi, Michael~I. Jordan, Martin~J. Wainwright, and Andre Wibisono.
\newblock Optimal rates for zero-order convex optimization: The power of two
  function evaluations.
\newblock {\em {IEEE} Trans. Information Theory}, 61(5):2788--2806, 2015.

\bibitem{faber}
V.~Faber and J.~Mycielski.
\newblock Applications of learning theorems.
\newblock {\em Fundmenta Informaticue}, 15(2):145--167, 1991.

\bibitem{flaxman2005soda}
Abraham~D Flaxman, Adam~Tauman Kalai, and H~Brendan McMahan.
\newblock Online convex optimization in the bandit setting: gradient descent
  without a gradient.
\newblock In {\em Proceedings of the sixteenth annual ACM-SIAM symposium on
  Discrete algorithms}, pages 385--394. Society for Industrial and Applied
  Mathematics, 2005.

\bibitem{foster91}
Dean~P. Foster.
\newblock Prediction in the worst case.
\newblock {\em Ann. Statist.}, 19(2):1084--1090, 06 1991.

\bibitem{hazan2016now}
Elad Hazan et~al.
\newblock Introduction to online convex optimization.
\newblock {\em Foundations and Trends{\textregistered} in Optimization},
  2(3-4):157--325, 2016.

\bibitem{hazan2014nips}
Elad Hazan and Kfir Levy.
\newblock Bandit convex optimization: Towards tight bounds.
\newblock In {\em Advances in Neural Information Processing Systems}, pages
  784--792, 2014.

\bibitem{hosseini2016tac}
Saghar Hosseini, Airlie Chapman, and Mehran Mesbahi.
\newblock Online distributed convex optimization on dynamic networks.
\newblock {\em IEEE Trans. Automat. Contr.}, 61(11):3545--3550, 2016.

\bibitem{jenatton2016}
Rodolphe Jenatton, Jim Huang, and Cedric Archambeau.
\newblock Adaptive algorithms for online convex optimization with long-term
  constraints.
\newblock In Maria~Florina Balcan and Kilian~Q. Weinberger, editors, {\em
  Proceedings of The 33rd International Conference on Machine Learning},
  volume~48, pages 402--411, 2016.

\bibitem{kivinen97}
Jyrki Kivinen and Manfred~K. Warmuth.
\newblock Exponentiated gradient versus gradient descent for linear predictors.
\newblock {\em Inf. Comput.}, 132(1):1--63, January 1997.

\bibitem{levin2008markov}
David~A Levin, Yuval Peres, and Elizabeth~L Wilmer.
\newblock Markov chains and mixing times amer.
\newblock {\em American Mathematical Society}, 2008.

\bibitem{mahdavi2012}
Mehrdad Mahdavi, Rong Jin, and Tianbao Yang.
\newblock Trading regret for efficiency: online convex optimization with long
  term constraints.
\newblock {\em Journal of Machine Learning Research}, 13:2503--2528, 2012.

\bibitem{Bartlett18}
Alan Malek and Peter~L Bartlett.
\newblock Horizon-independent minimax linear regression.
\newblock In {\em Advances in Neural Information Processing Systems}, pages
  5264--5273, 2018.

\bibitem{McMahan2004}
H~Brendan McMahan and Avrim Blum.
\newblock Online geometric optimization in the bandit setting against an
  adaptive adversary.
\newblock In {\em International Conference on Computational Learning Theory},
  pages 109--123. Springer, 2004.

\bibitem{submodular2016}
Baharan Mirzasoleiman, Amin Karbasi, Rik Sarkar, and Andreas Krause.
\newblock Distributed submodular maximization.
\newblock {\em The Journal of Machine Learning Research}, 17(1):8330--8373,
  2016.

\bibitem{nedic2018ieee}
Angelia Nedi{\'c}, Alex Olshevsky, and Michael~G Rabbat.
\newblock Network topology and communication-computation tradeoffs in
  decentralized optimization.
\newblock {\em Proceedings of the IEEE}, 106(5):953--976, 2018.

\bibitem{nedic2010tac}
Angelia Nedic, Asuman Ozdaglar, and Pablo~A Parrilo.
\newblock Constrained consensus and optimization in multi-agent networks.
\newblock {\em IEEE Transactions on Automatic Control}, 55(4):922--938, 2010.

\bibitem{nesterov2017}
Yurii Nesterov and Vladimir Spokoiny.
\newblock Random gradient-free minimization of convex functions.
\newblock {\em Found. Comput. Math.}, 17(2):527--566, April 2017.

\bibitem{Raginsky2011}
Maxim Raginsky, Nooshin Kiarashi, and Rebecca Willett.
\newblock Decentralized online convex programming with local information.
\newblock In {\em American Control Conference (ACC), 2011}, pages 5363--5369.
  IEEE, 2011.

\bibitem{saha2011aistat}
Ankan Saha and Ambuj Tewari.
\newblock Improved regret guarantees for online smooth convex optimization with
  bandit feedback.
\newblock In {\em Proceedings of the Fourteenth International Conference on
  Artificial Intelligence and Statistics}, pages 636--642, 2011.

\bibitem{scaman2018}
Kevin Scaman, Francis Bach, Sebastien Bubeck, Laurent Massouli\'{e}, and
  Yin~Tat Lee.
\newblock Optimal algorithms for non-smooth distributed optimization in
  networks.
\newblock In {\em Advances in Neural Information Processing Systems 31}, pages
  2745--2754. 2018.

\bibitem{sss2011}
Shai Shalev-Shwartz et~al.
\newblock Online learning and online convex optimization.
\newblock {\em Foundations and Trends{\textregistered} in Machine Learning},
  4(2):107--194, 2012.

\bibitem{shamir2017jmlr}
Ohad Shamir.
\newblock An optimal algorithm for bandit and zero-order convex optimization
  with two-point feedback.
\newblock {\em Journal of Machine Learning Research}, 18(52):1--11, 2017.

\bibitem{tsitsiklis2986}
John Tsitsiklis, Dimitri Bertsekas, and Michael Athans.
\newblock Distributed asynchronous deterministic and stochastic gradient
  optimization algorithms.
\newblock {\em IEEE transactions on automatic control}, 31(9):803--812, 1986.

\bibitem{vovk98}
Volodya Vovk.
\newblock Competitive on-line linear regression.
\newblock In M.~I. Jordan, M.~J. Kearns, and S.~A. Solla, editors, {\em
  Advances in Neural Information Processing Systems 10}, pages 364--370. 1998.

\bibitem{fan2013tkde}
Feng Yan, Shreyas Sundaram, SVN Vishwanathan, and Yuan Qi.
\newblock Distributed autonomous online learning: Regrets and intrinsic
  privacy-preserving properties.
\newblock {\em IEEE Transactions on Knowledge and Data Engineering},
  25(11):2483--2493, 2013.

\bibitem{yuan2018nips}
Jianjun Yuan and Andrew Lamperski.
\newblock Online convex optimization for cumulative constraints.
\newblock In {\em Advances in Neural Information Processing Systems}, 2018.

\bibitem{zhang2017projection}
Wenpeng Zhang, Peilin Zhao, Wenwu Zhu, Steven~CH Hoi, and Tong Zhang.
\newblock Projection-free distributed online learning in networks.
\newblock In {\em International Conference on Machine Learning}, pages
  4054--4062, 2017.

\bibitem{zinkevich2003icml}
Martin Zinkevich.
\newblock Online convex programming and generalized infinitesimal gradient
  ascent.
\newblock In {\em Proceedings of the 20th International Conference on Machine
  Learning (ICML-03)}, pages 928--936, 2003.

\end{thebibliography}
      \end{document}